%% file: acl_latex.tex
\pdfoutput=1

\documentclass[11pt,table]{article}

\usepackage[final]{acl}
\PassOptionsToPackage{hyphens}{url}
\usepackage[frozencache,cachedir=.]{minted}
\definecolor{bg}{gray}{0.95}
\usepackage{times}
\usepackage{latexsym}
\usepackage{collcell}
\usepackage[T1]{fontenc}

\usepackage[utf8]{inputenc}

\usepackage{microtype}

\usepackage{inconsolata}

\usepackage{times,latexsym}
\usepackage{url}
\usepackage{pgf}
\usepackage{booktabs}

\usepackage{extdash} 

\usepackage{hhline}
\usepackage{amsmath}
\usepackage{amssymb}
\usepackage{multirow}
\usepackage{arydshln}
\usepackage{enumitem}
\usepackage{multicol}
\usepackage{xspace}
\usepackage{wasysym}
\usepackage{hyperref}
\hypersetup{
           breaklinks=true,   
           colorlinks=true,   
           pdfusetitle=true,  
           citecolor=darkblue, 
           linkcolor=darkblue, 
           urlcolor=darkblue
        }

\usepackage{rotating}
\usepackage[nameinlink,capitalize,noabbrev]{cleveref}
\usepackage{nicefrac}

\usepackage{adjustbox}
\usepackage{graphicx}
\usepackage{svg}
\usepackage{subcaption}

\definecolor{Gray}{gray}{0.9}
\definecolor{cb-blue-green} {RGB}{ 0,  073,  073}
\definecolor{cb-green-sea}  {RGB}{ 0, 146, 146}
\definecolor{cb-rose}       {RGB}{255, 109, 182}
\definecolor{cb-salmon-pink}{RGB}{255, 182, 119}
\definecolor{cb-purple}     {RGB}{ 73,   0, 146}
\definecolor{cb-blue}       {RGB}{ 0, 109, 219}
\definecolor{cb-lilac}      {RGB}{182, 109, 255}
\definecolor{cb-blue-sky}   {RGB}{109, 182, 255}
\definecolor{cb-blue-light} {RGB}{182, 219, 255}
\definecolor{cb-burgundy}   {RGB}{146,   0,   0}
\definecolor{cb-brown}      {RGB}{146,  73,   0}
\definecolor{cb-clay}       {RGB}{219, 209,   0}
\definecolor{cb-green-lime} {RGB}{ 36, 255,  36}
\definecolor{cb-yellow}     {RGB}{255, 255, 109}
\definecolor{cb-grey}       {RGB}{233, 233, 233}

\usepackage{expex}
\usepackage[linguistics]{forest}

\usepackage{tikz}
\usepackage{amssymb}
\usepackage{pifont}
\newcommand{\cmark}{\textcolor{cb-blue-green}{\ding{51}}}%
\newcommand{\xmark}{\textcolor{cb-burgundy}{\ding{55}}}%

\newcommand{\logo}[0]{\raisebox{-.3\height}{\includegraphics[width=.04\textwidth]{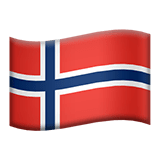}}} 

\newcommand{\robot}[0]{\includegraphics[width=.017\textwidth]{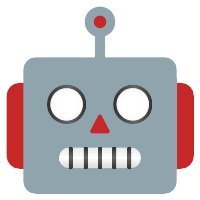}}

\newcommand{\human}[0]{\includegraphics[width=.017\textwidth]{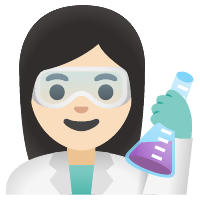}}

\newcommand{\openai}[0]{\includegraphics[width=.017\textwidth]{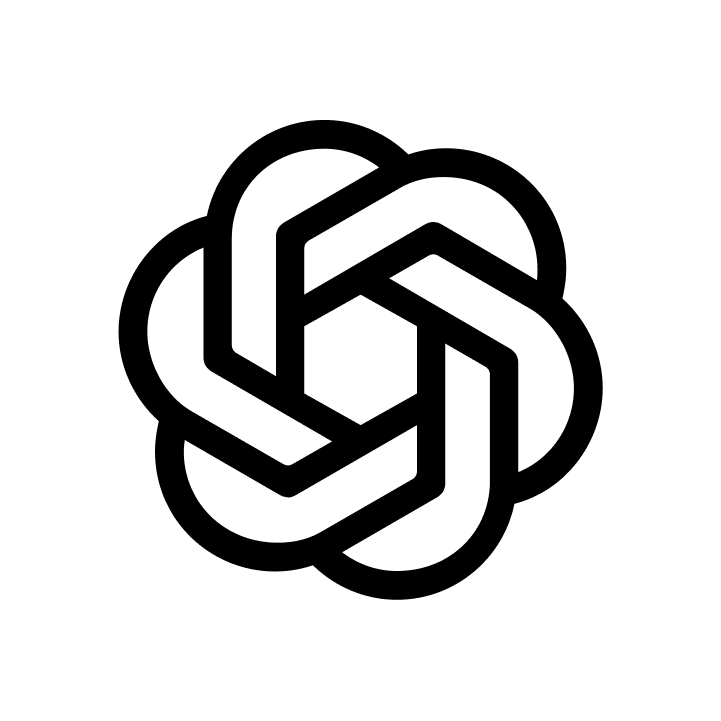}}

\newcommand{\cool}[0]{\includegraphics[width=.024\textwidth]{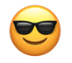}}

\newcommand{\cat}[0]{\includegraphics[width=.024\textwidth]{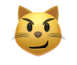}}

\newcommand{\rocket}[0]{\includegraphics[width=.024\textwidth]{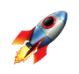}}

\title{\logo \hspace{0.01em} NorEval: A Norwegian Language Understanding and Generation Evaluation Benchmark}

\author{Vladislav Mikhailov$^1$ \hspace{0.7em} Tita Enstad$^2$ \hspace{0.7em} David Samuel$^1$ \\ \textbf{Hans Christian Farsethås}$^1$ \hspace{0.7em} \textbf{Andrey Kutuzov}$^1$ \hspace{0.7em} \textbf{Erik Velldal}$^1$ \hspace{0.7em} \textbf{Lilja Øvrelid}$^1$\\
$^1$University of Oslo \\
$^2$National Library of Norway \\
\small{
    \textbf{Correspondence:} \href{mailto:vladism@ifi.uio.no}{\texttt{vladism@ifi.uio.no}}
}}

\begin{document}
\maketitle

\begin{abstract}
This paper introduces NorEval, a new and comprehensive evaluation suite for large-scale standardized benchmarking of Norwegian generative language models (LMs). NorEval consists of 24 high-quality human-created datasets -- of which five are created from scratch. In contrast to existing benchmarks for Norwegian, NorEval covers a broad spectrum of task categories targeting Norwegian language understanding and generation, establishes human baselines, and focuses on  both of the official written standards of the Norwegian language: Bokmål and Nynorsk. All our datasets and a collection of over 100 human-written prompts are integrated into LM Evaluation Harness, ensuring flexible and reproducible evaluation. We describe the NorEval design and present the results of benchmarking 19 open-source pretrained and instruction-tuned LMs for Norwegian in various scenarios. Our benchmark, evaluation framework, and annotation materials are publicly available. 
\end{abstract}

\input{sections/intro}

\input{sections/related_work}

\input{sections/design}

\input{sections/setup}

\input{sections/results}

\input{sections/conclusion}

\input{sections/limitations}

\input{sections/ethical_statement}

\section*{Acknowledgments}
NorEval has developed from Mímir, a project on evaluating the impact of copyrighted data on pretraining Norwegian LMs \cite{rosa-etal-2025-impact}. We thank our student annotators for their annotation efforts. We also thank our volunteers for their time and contribution to establishing our human baselines: Helene Bøsei Olsen, Lilja Charlotte Storset, Sondre Wold, Petter Mæhlum, Victoria Ovedie Chruickshank Langø, Egil Rønningstad, Emil Poiesz, Thea Tollersrud, and Asbjørn Sæther.

\bibliography{custom,anthology}

\clearpage
\newpage

\appendix

\input{appendix/design}

\clearpage
\newpage

\input{appendix/datasets}

\clearpage
\newpage

\input{appendix/prompt_guidelines}

\clearpage
\newpage

\input{appendix/human_evaluation_guidelines}

\clearpage
\newpage

\input{appendix/results}

\clearpage
\newpage

\input{appendix/judge}

\end{document}

%% file: sections/intro.tex
\section{Introduction}
\label{sec:intro}
The advancement of language models (LMs) is inseparable from benchmarking -- the  systematic evaluation of their generalization abilities on standardized datasets across various criteria \cite{ruder2021challenges,srivastava2023beyond}. Despite its crucial role, benchmarking in resource-lean scenarios remains scarce due to the lack of diverse evaluation suites for low-resource languages, including Norwegian \cite{joshi-etal-2020-state,hedderich-etal-2021-survey}. 

Previous work focuses on Norwegian as part of medium-scale benchmarking efforts -- NorBench \cite{samuel-etal-2023-norbench} and NLEBench \cite{liu-etal-2024-nlebench} -- and broader Mainland Scandinavian evaluation initiatives -- ScandEval \cite{nielsen-2023-scandeval} and Scandinavian Embedding Benchmark (SEB; \citealp{enevoldsen2024scandinavian}). However, these benchmarks have several shortcomings that limit the scope of LM evaluation in Norwegian. 

\begin{itemize}[itemsep=-2pt,partopsep=0.5ex,parsep=1ex,leftmargin=1.5em]
    \item \textbf{Coverage and design.} These benchmarks exhibit a significant dataset overlap with a low variation in task formulations. NorBench and ScandEval cover traditional NLP tasks, SEB addresses text embedding evaluation, and NLEBench comprises a narrow spectrum of 
    Norwegian language generation tasks.
    \item \textbf{Data quality.} NLEBench and ScandEval include machine-translated English datasets, introducing potential evaluation biases that may conflict with Norwegian-specific values, culture, and knowledge.
    \item \textbf{Linguistic diversity.} Norwegian has two official written standards: Bokmål (BM) and Nynorsk (NN; the minority variant). The latter variant remains significantly underrepresented in previous work.
    \item \textbf{Human performance.} No existing benchmark establishes human baselines, which is a standard practice to approximate upper LM performance bounds.
\end{itemize}

\noindent
This paper introduces NorEval, a novel large-scale evaluation suite designed to benchmark Norwegian LMs on language understanding and generation tasks. NorEval comprises 24 human-created datasets across nine task categories, including sentiment analysis, Norwegian language knowledge, Norwegian-specific \& world knowledge, machine reading comprehension, commonsense reasoning, machine translation, text summarization, instruction following, and truthfulness. Our design enables various benchmarking scenarios, ranging from multi-prompt $k$-shot evaluation to side-by-side LM comparison on diverse user instructions.

\input{tables/benchmarks}

Our main contributions are: (i) we create NorEval, the largest multi-task benchmark for Norwegian Bokmål and Nynorsk that combines 19 existing peer-reviewed datasets with five datasets created from scratch; (ii) we curate a collection of over 100 dataset-specific prompts for robust evaluation; (iii) we establish five human baselines; (iv) we benchmark 19 pretrained and instruction-tuned Norwegian LMs against each other and humans; and (v) we release NorEval and our annotation materials.\footnote{\href{https://github.com/ltgoslo/noreval/tree/main}{\texttt{ltgoslo/noreval}}}

%% file: tables/benchmarks.tex
\begin{table*}[th!]
\centering
\scriptsize
\resizebox{\textwidth}{!}{
    \begin{tabular}{@{}llrcccccc@{}}
    \toprule
     & \multirow{2}{*}{\vspace{-0.9em}\textbf{Evaluation Scope}} & \multirow{2}{*}{\vspace{-0.9em}\textbf{Task Categories}} & \multicolumn{3}{c}{\textbf{\# Datasets}} & \multicolumn{3}{c}{\textbf{Method}} \\ \cmidrule{4-9}
    & & & \textbf{BM} & \textbf{NN} & \textbf{Total} & \human & \robot & \openai \human  \\
    \midrule
    NorBench & NLU \& NLG &  \begin{tabular}{@{}r@{}} POS-tagging, MT, \\ NER, sentiment analysis, \\ Acceptability classification, RC \end{tabular} & 8 & 2 & 10 &  \cmark & \xmark & \xmark \\
    \vspace{-0.8em}\\
    \hdashline
    \vspace{-1em}\\
 
    ScandEval & NLU \& NLG & \begin{tabular}{@{}r@{}} NER, sentiment analysis, \\ Acceptability classification, RC, \\ Commonsense reasoning, \\ Text summarization, multiple-choice QA \end{tabular}  & 8 & 2 & 10 & \cmark & \cmark & \xmark \\
    \vspace{-0.8em}\\
    \hdashline
    \vspace{-1em}\\
    SEB & \begin{tabular}{@{}l@{}} Text embedding \\ evaluation \end{tabular}  & \begin{tabular}{@{}r@{}} LID, sentiment analysis, \\ Acceptability classification, retrieval, \\ Dialect \& written form pairing, \\ Intent \& scenario classification, \\ Clustering, political speech classification \end{tabular}  & 11 & 3 & 14 & \cmark & \xmark & \xmark  \\ \vspace{-0.8em}\\
    \hdashline
    \vspace{-1em}\\
       NLEBench & NLU \& NLG & \begin{tabular}{@{}r@{}} NLI, RC, bias detection,  \\ Text summarization, yes/no QA, \\ Instruction following, \\ Paraphrase detection, open-ended conversation \end{tabular}  & 9 & \xmark & 9 & \xmark & \cmark & \cmark \\
    \vspace{-0.8em}\\
    \hdashline
    \vspace{-1em}\\
    NorEval & NLU \& NLG  & \begin{tabular}{@{}r@{}} Commonsense reasoning, \\ RC, sentiment analysis, \\ Norwegian language knowledge, MT, \\ Truthfulness, text summarization, \\ Instruction following, \\ Norwegian-specific \& world knowledge \end{tabular}  & 16 & 8 & 24 & \cmark & \xmark & \xmark \\
    \bottomrule
    \end{tabular}
}
\caption{\textbf{Comparison of multi-task benchmarks for Norwegian:} ScandEval \cite{nielsen-2023-scandeval}, Scandinavian Embedding Benchmark (SEB; \citealp{enevoldsen2024scandinavian}), NorBench \cite{samuel-etal-2023-norbench}, NLEBench \cite{liu-etal-2024-nlebench}, and NorEval (ours).  BM=Norwegian Bokmål; NN=Norwegian Nynorsk; \human=human-created; \robot=machine-translated; \openai \human= GPT-4o-created \& human-edited; NLU=Natural language understanding; NLG=Natural language generation; NER=named entity recognition; LID=language identification; RC=reading comprehension; NLI=natural language inference; QA=question answering; MT=machine translation.}
\label{tab:datasets}
\end{table*}

%% file: sections/related_work.tex
\section{Background}
\label{sec:background}
\paragraph{Norwegian Bokmål and Nynorsk} BM is the primary written standard, while an estimated 10--15\% of the Norwegian population uses NN -- especially in Western Norway. The national language legislation specifies that minimally 25\% of the written public service information should be in  NN to ensure representation of both varieties. While BM and NN are closely related, they exhibit lexical and grammatical differences, e.g., distinct pronouns, plural noun forms, definite noun forms, verb conjugation, and vocabulary units. Consider an example of such differences based on one of our text summarization prompts ``\texttt{\textcolor{cb-lilac}{Give} \textcolor{cb-rose}{a} \textcolor{cb-salmon-pink}{brief} \textcolor{cb-purple}{summary} \textcolor{cb-blue}{of} \textcolor{cb-burgundy}{the following} \textcolor{cb-blue}{text}: \{\{article\}\}}'' (see \S\ref{subsec:prompts}).

\begin{itemize}[itemsep=-2pt,partopsep=0.5ex,parsep=1ex,leftmargin=1.5em]
\item \textbf{BM.} ``\texttt{\textcolor{cb-lilac}{Gi} \textcolor{cb-rose}{et} \textcolor{cb-salmon-pink}{kortfattet} \textcolor{cb-purple}{sammendrag} \textcolor{cb-blue}{av} \textcolor{cb-burgundy}{følgende} \textcolor{cb-blue}{tekst}: \{\{article\}\}}''.
\item \textbf{NN.} ``\texttt{\textcolor{cb-lilac}{Gje} \textcolor{cb-rose}{eit} \textcolor{cb-salmon-pink}{kortfatta} \textcolor{cb-purple}{samandrag} \textcolor{cb-blue}{av} \textcolor{cb-burgundy}{følgande} \textcolor{cb-blue}{tekst}: \{\{article\}\}}''.
\end{itemize}

\noindent We make one of the first  attempts to increase the representation of NN in benchmarking LMs.

\begin{figure*}[th!]
    \centering
    \includegraphics[width=\textwidth]{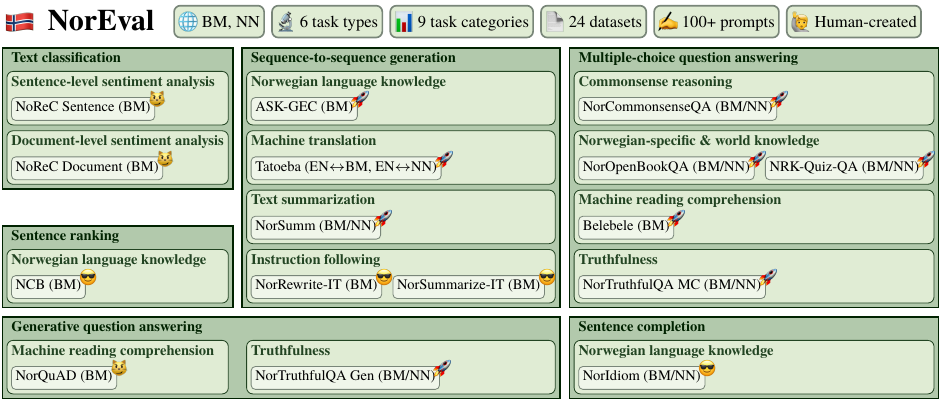}
    \caption{\textbf{Overview of the NorEval design.}  \cat \hspace{0.01em} denotes datasets used in the related studies (\S\ref{sec:background}), \rocket \hspace{0.01em} represents datasets not previously included in the existing Norwegian benchmarks, and \cool \hspace{0.01em} denotes our novel datasets introduced as part of NorEval. EN=English; BM=Norwegian Bokmål; NN=Norwegian Nynorsk.}
    \label{fig:noreval}
\end{figure*}

\paragraph{Norwegian Benchmarks}
\autoref{tab:datasets} provides an overview of existing Norwegian benchmarks w.r.t. the evaluation scope, task categories, the number of datasets, coverage of BM and NN, and dataset creation method. We describe them below.

\begin{enumerate}[itemsep=-2pt,partopsep=0.5ex,parsep=1ex,leftmargin=1.5em]
    \item \textbf{NorBench} is primarily designed to benchmark encoder-only LMs on a collection of ten traditional NLP tasks, such as PoS-tagging, NER (NorNE; \citealp{jorgensen-etal-2020-norne}), sentiment analysis at different levels of granularity (NoReC; \citealp{velldal-etal-2018-norec,ovrelid-etal-2020-fine}), acceptability classification (NoCoLA; \citealp{jentoft-samuel-2023-nocola}), machine translation, and extractive question answering (NorQuAD; \citealp{ivanova-etal-2023-norquad}). All datasets in NorBench are human-created; however, the support for NN is limited to  PoS-tagging and NER based on the Norwegian UD treebanks \cite{ovrelid-hohle-2016-universal,velldal-etal-2017-joint}.
    \item \textbf{ScandEval} is an evaluation suite coupled with a public leaderboard for Scandinavian languages: Danish, Faroese, Icelandic, Norwegian, and Swedish. The Norwegian datasets in ScandEval are based on existing resources, such as NoReC, NorNE, NorQuAD, and the SNL \& VG summarization dataset \cite{navjord2023beyond}. ScandEval introduces ScaLA, an acceptability classification dataset created through rule-based perturbation of sentences from the Norwegian UD treebanks. Moreover, its latest version contains machine-translated English datasets that are not curated or post-processed:\footnote{ScandEval has been extended to EuroEval, which supports existing and machine-translated evaluation resources for Norwegian: \href{https://euroeval.com/datasets/norwegian/}{\texttt{euroeval.com/datasets/norwegian}}.} MMLU \cite{hendrycks2021measuring}, ARC \cite{clark2018think}, XSum \cite{narayan-etal-2018-dont}, and HellaSwag \cite{zellers-etal-2019-hellaswag}. Similar to NorBench, the coverage of NN is limited to the datasets derived from the Norwegian UD treebanks.
    \item \textbf{SEB} is designed to evaluate text representations for Scandinavian languages across retrieval, bitext mining, text classification, and clustering tasks. With its distinct focus on text embedding models, SEB has little overlap with other Norwegian benchmarks (except for NorQuAD, ScaLA, and SNL \& VG) and primarily constructs its evaluation tasks by converting existing Norwegian resources and leveraging supported metadata and schemes.
    \item \textbf{NLEBench} is designed to evaluate the LM's Norwegian language generation capabilities. Although NLEBench covers various task categories, it does not address any NN evaluation scenario. Moreover, seven out of nine datasets are machine-translated without curation, raising concerns about the benchmark's reliability. The remaining two datasets comprise multi-turn conversation, closed question answering (QA), and abstractive summarization tasks; these are generated by GPT-4o and edited by Norwegian native speakers.
\end{enumerate}

\noindent NorEval expands the scope of benchmarking Norwegian LMs to task categories, datasets, and evaluation scenarios that have not been covered in the related studies, with the main focus on human-created resources. In particular, only three out of 24 NorEval datasets are included in NorBench, ScandEval, SEB, and NLEBench: NorQuAD and sentence- and document-level NoReC. 

%% file: sections/design.tex
\section{NorEval}
\label{sec:noreval}
Our main goal is to develop a high-quality standardized evaluation suite to benchmark Norwegian generative LMs across a broad spectrum of Norwegian language understanding and generation tasks. \autoref{fig:noreval} outlines the design of NorEval, which combines 19 existing peer-reviewed datasets with five novel datasets (\S\ref{subsec:tasks}), comprises a pool of over 100 prompts (\S\ref{subsec:prompts}), and offers a framework for systematic and reproducible LM evaluation (\S\ref{subsec:framework}).

\subsection{Tasks}
\label{subsec:tasks}
\autoref{app:design} presents an overview of our 24 datasets, including dataset descriptions and examples, task formulations, prompts, performance metrics, and general statistics. \autoref{app:datasets} details our novel datasets (NCB, NorIdiom, NorRewrite-instruct, and NorSummarize-Instruct). We describe NorEval based on nine high-level task categories:

\vspace{0.15cm} \noindent \textbf{Sentiment analysis} focuses on a binary polarity classification at the sentence- and document-level (NoReC Sentence \& Document).

\vspace{0.15cm} \noindent \textbf{Norwegian language knowledge} assesses an LM's capabilities to perform grammatical error correction (ASK-GEC; \citealp{jentoft2023grammatical}), adhere to language-specific punctuation rules (NCB; ours), and complete Norwegian idioms (NorIdiom; ours).

\vspace{0.15cm} \noindent \textbf{Norwegian-specific \& world knowledge} tests an LM's capabilities to answer multiple-choice questions based on real-world and Norwegian-specific cultural knowledge (NRK-Quiz-QA and NorOpenBookQA; \citealp{mikhailov-etal-2025-collection}).

\vspace{0.15cm} \noindent \textbf{Machine reading comprehension} evaluates the capabilities of LMs to answer questions related to an input text by selecting an answer from multiple choices (Belebele; \citealp{bandarkar-etal-2024-belebele}) or generating a text span (NorQuAD).

\vspace{0.15cm} \noindent \textbf{Commonsense reasoning} assesses an LM's capabilities to answer a multiple-choice question based on logical reasoning and world understanding (NorCommonsenseQA; \citealp{mikhailov-etal-2025-collection}).

\vspace{0.15cm} \noindent \textbf{Machine translation} tests an LM's translation capabilities among four language pairs from Tatoeba \cite{tiedemann-2020-tatoeba}: EN  $\leftrightarrow$ BM and EN $\leftrightarrow$ NN.

\vspace{0.15cm} \noindent \textbf{Text summarization} focuses on abstractive news summarization (NorSumm; \citealp{touileb-etal-2025-benchmarking}).

\vspace{0.15cm} \noindent \textbf{Instruction following} evaluates an LM's capabilities to follow  instructions on creative rewriting and summarization through, e.g., changing a text's tone and style, simplifying complex content, and adapting content for a specific audience (NorRewrite-Instruct and NorSummarize-Instruct; ours).

\vspace{0.15cm} \noindent \textbf{Truthfulness} tests whether an LM generates or selects answers that propagate false beliefs and misconceptions (NorTruthfulQA Multiple Choice \& Generation; \citealp{mikhailov-etal-2025-collection}).  

\subsection{Prompts}
\label{subsec:prompts}
We conduct a two-stage in-house annotation to create a collection of prompts that reflect diverse user formulations and answer formatting, with four-to-six prompts per dataset. The prompt examples are provided in \autoref{app:design}, and the annotation guidelines are documented in \autoref{sec:appendix_prompt_guidelines}.

\begin{itemize}[itemsep=-2pt,partopsep=0.5ex,parsep=1ex,leftmargin=1.5em]
    \item \textbf{Stage 1: Creating Prompts in Bokmål.} Three Norwegian native speakers create dataset-specific prompts in BM using two strategies: (i) manually translating English prompts from PromptSource \cite{bach-etal-2022-promptsource} and (ii) writing the prompts from scratch.
    \item \textbf{Stage 2: Adapting Prompts to Nynorsk.} We hire a BA student in linguistics to adapt the BM prompts to NN. The hourly pay rate is 227 NOK (approx. \$20).
\end{itemize}

\subsection{Evaluation Framework}
\label{subsec:framework}
All our datasets and prompts are integrated into LM Evaluation Harness \cite{eval-harness,biderman2024lessons},\footnote{\href{https://github.com/EleutherAI/lm-evaluation-harness/tree/main/lm_eval/tasks/noreval}{\texttt{github.com/EleutherAI/lm-evaluation-harness}}} a framework for flexible evaluation of generative LLMs in various scenarios. The framework provides a user-friendly API allowing to easily integrate datasets, configure prompts, and benchmark LMs that are not part of our baselines.

%% file: sections/setup.tex
\input{tables/lms}

\section{Evaluation Setup}
\label{sec:setup}
We benchmark a broad range of 19 open-source pretrained and instruction-finetuned decoder-only LMs available in Transformers (\citealp[]{wolf-etal-2020-transformers}; see \autoref{tab:lms}). We compare them in $k$-shot regimes against one another and our human baselines, and evaluate the instruction-finetuned LMs using the LLM-as-a-judge approach \cite{zheng2023judging}.

\paragraph{In-context Learning Evaluation} The evaluation is run in $k$-shot regimes with $k \in \{0, 1, 16\}$ across \emph{all} prompts. We use the maximum $k$ for each task, which depends on the availability of a training/development set for demonstration examples and the example lengths. We use two strategies supported via LM Evaluation Harness to evaluate the LM performance in a prompted format:\footnote{\autoref{fig:noreval} outlines our sentence ranking, text classification, sentence completion, sequence-to-sequence generation, and multiple-choice and generative QA tasks.}

\begin{itemize}[itemsep=-2pt,partopsep=0.25ex,parsep=1ex,leftmargin=1.5em]
    \item \textbf{Log-likelihood.} The LM assigns a probability to each answer candidate conditioned on an input prompt, and the most probable candidate is selected as the prediction. This strategy is used in the sentence ranking, text classification, and multiple-choice QA tasks.
    
    \item \textbf{Generation.} The LM generates a text continuation conditioned on an input prompt. We use a greedy search decoding method for the pretrained LMs and recommended HuggingFace inference hyperparameters and chat templates for the instruction-tuned LMs. This strategy is used in the sentence completion, sequence-to-sequence generation, and generative QA tasks.
\end{itemize}

\paragraph{Performance Aggregation} We use a combination of performance aggregation methods based on well-established NLP benchmarking practices and theoretical foundations of the social choice theory \cite{arrow2012social}.

\begin{itemize}[itemsep=-2pt,partopsep=0.25ex,parsep=1ex,leftmargin=1.5em]
    \item \textbf{Multi-prompt Aggregation.} We select the highest performance score for each LM across task-specific prompts to mitigate the prompt sensitivity \cite{voronov-etal-2024-mind}.
    
    \item \textbf{Average Normalized Score.} In line with the OpenLLM leaderboard \cite{fourrier2024open} and FineWeb2 evaluation protocol \cite{penedo2024fineweb-2}, we first rescale individual performance scores across our nine task categories. Rescaling involves score normalization between the random baseline and the maximum possible score. We then compute the
    overall performance score by averaging the normalized scores within all task categories.
    
    \item \textbf{Borda's Count.} Recent works demonstrate the effectiveness of using Borda's count as an alternative to arithmetic mean aggregation in multi-task benchmarking \cite{colombo2022best,rofin-etal-2023-votenrank}. This approach relies on a scoring vector $c=(|M|-1,|M|-2,\ldots,1,0)$ to assign scores to a set of $M$ LMs $m \in \{m_1,\ldots, m_{|M|}\}$ based on their positions in each task- and metric-specific ranking. The final score is calculated as the sum of corresponding scores in each task $Sc(m)=\sum_{i=1}^{|M|}{c_i p_i(m)}$, where $p_i(m)$ is the number of tasks in which LM $m$ takes the $i^{th}$ place, and $c_i$ is the $i^{th}$ element of $c$. Borda's count allows for aggregating heterogeneous performance metrics while accounting for the differences in the LMs' ranking positions.

\end{itemize}

\input{tables/iaa}

\input{tables/overall}

\paragraph{Human Baselines}
\label{subsec:human_baselines}
We establish five human baselines on random subsets of 50 examples from NCB, Belebele, NorOpenBookQA (BM), NorCommonsenseQA (BM), and NorTruthfulQA Multiple choice (BM). Our annotation team consists of 12 volunteers, all Norwegian native speakers with an NLP background and completed higher academic degrees. Before starting, the annotators receive guidelines describing the tasks and providing examples with explanations (see \autoref{sec:appendix_human_evaluation_guidelines}). Each example is annotated by three annotators, and we use majority voting to aggregate their results. \autoref{tab:wawa} summarizes the inter-annotator agreement rates based on the Worker Agreement with Aggregate (WAWA) coefficient \cite{ning-etal-2018-joint}, which represents the average percentage of annotators' votes that align with the majority votes. The WAWA rates range between 86\% and 98\%, which shows a strong agreement between our annotators.

\paragraph{LLM-as-a-judge} We use the LLM-as-a-judge approach to automatically evaluate the instruction-tuned LMs' generation abilities on NorRewrite-Instruct and NorSummarize-Instruct. We adopt the Human response-guided evaluation framework (HREF; \citealp{lyu2024href}), which relies on human references as additional inputs to improve the LM judgement performance. Our judge model is 
\href{https://huggingface.co/meta-llama/Llama-3.3-70B-Instruct}{\texttt{meta-llama/Llama-3.3-70B-Instruct}}, which highly correlates with human judgments as reported by  \citeauthor{lyu2024href}. The judge model is given (i) the prompt; (ii) output A; (iii) output B; and (iv) a human reference formatted based on the prompt template in \Cref{app:judge-prompt}. We perform the side-by-side comparison using a greedy search decoding strategy across three options: (i) output A is better than output B; (ii) output B is better than output A; and (iii) a tie; and compute the expected win rates as specified in \Cref{app:judge}.

%% file: tables/lms.tex
\begin{table}[t!]
    \centering
    \small
    \resizebox{\columnwidth}{!}{ %
    \begin{tabular}{@{}ll@{}}
    \toprule
    \textbf{Name} & \textbf{Base} \\ \midrule
    \textsc{\textbf{Pretrained LMs}}\\[0.5em]
    \href{https://huggingface.co/mistralai/Mistral-7B-v0.1}{Mistral-7B} & N/A   \\
    \href{https://huggingface.co/mistralai/Mistral-Nemo-Base-2407}{Mistral-Nemo-12B} & N/A  \\[0.7em]
    \href{https://huggingface.co/meta-llama/Meta-Llama-3-8B}{Meta/Llama-3-8B} & N/A \\[0.7em]
    \href{https://huggingface.co/NbAiLab/nb-gpt-j-6B-v2}{NB-GPT-6B} & N/A \\[0.7em]

    \href{https://huggingface.co/NorwAI/NorwAI-Mistral-7B}{NorwAI-Mistral-7B} & \href{https://huggingface.co/mistralai/Mistral-7B-v0.1}{Mistral-7B}  \\
    \href{https://huggingface.co/NorwAI/NorwAI-Llama2-7B}{NorwAI-Llama2-7B} & \href{https://huggingface.co/meta-llama/Llama-2-7b-hf}{Llama-2-7B}  \\[0.7em]

    \href{https://huggingface.co/AI-Sweden-Models/gpt-sw3-6.7b-v2}{GPT-SW3-6.7B} & N/A \\
    \href{https://huggingface.co/AI-Sweden-Models/Llama-3-8B}{AI-Sweden/Llama-3-8B} & \href{https://huggingface.co/meta-llama/Meta-Llama-3-8B}{Meta/Llama-3-8B} \\[0.7em]

    \href{https://huggingface.co/LumiOpen/Viking-7B}{Viking-7B}  & N/A  \\
    \href{https://huggingface.co/LumiOpen/Viking-13B}{Viking-13B} & N/A  \\ [0.7em]

    \href{https://huggingface.co/norallm/norbloom-7b-scratch}{NorBLOOM-7B-scratch} & N/A \\
    \href{https://huggingface.co/norallm/normistral-7b-scratch}{NorMistral-7B-scratch} & N/A  \\
    \href{https://huggingface.co/norallm/normistral-7b-warm}{NorMistral-7B-warm} & \href{https://huggingface.co/mistralai/Mistral-7B-v0.1}{Mistral-7B} \\
    \href{https://huggingface.co/norallm/normistral-11b-warm}{NorMistral-11B-warm} & \href{https://huggingface.co/mistralai/Mistral-Nemo-Base-2407}{Mistral-Nemo-12B} \\[1em]
    
    \textsc{\textbf{Instruction-tuned LMs}}\\[0.5em]
    \href{https://huggingface.co/norallm/normistral-7b-warm-instruct}{NorMistral-7B-warm-IT} & \href{https://huggingface.co/norallm/normistral-7b-warm}{NorMistral-7B-warm} \\ 

    \href{https://huggingface.co/mistralai/Mistral-7B-Instruct-v0.1}{Mistral-7B-IT} & \href{https://huggingface.co/mistralai/Mistral-7B-v0.1}{Mistral-7B} \\
    
    \href{https://huggingface.co/AI-Sweden-Models/Llama-3-8B-instruct}{AI-Sweden/Llama-3-8B-IT} & \href{https://huggingface.co/AI-Sweden-Models/Llama-3-8B}{AI-Sweden/Llama-3-8B} \\

    \href{https://huggingface.co/meta-llama/Meta-Llama-3-8B-Instruct}{Meta/Llama-3-8B-IT} & \href{https://huggingface.co/meta-llama/Meta-Llama-3-8B}{Meta/Llama-3-8B} \\ 

    \href{https://huggingface.co/mistralai/Mistral-Nemo-Instruct-2407}{Mistral-Nemo-12B-IT} & \href{https://huggingface.co/mistralai/Mistral-Nemo-Base-2407}{Mistral-Nemo-12B} \\
    
    \bottomrule
    \end{tabular}%
    } 
     \caption{\textbf{The LMs used in our work and their base versions.} LM references: Mistral-7B \cite{jiang2023mistral}, NorBLOOM/NorMistral-7B-scratch \& Normistral-7B/11B-warm \cite{samuel-etal-2025-small}, and Meta/Llama-3-8B \cite{dubey2024llama}.}
    \label{tab:lms}
\end{table}

%% file: tables/iaa.tex
\begin{table}[t!]
\centering
\scriptsize
\resizebox{\columnwidth}{!}{
\begin{tabular}{@{}lr@{}}
\toprule
\textbf{Dataset} & \textbf{WAWA} \\ \midrule
NCB & 92.0 \\
NorOpenBookQA (BM) & 98.0 \\
NorCommonsenseQA (BM) & 93.3 \\
NorTruthfulQA Multiple Choice (BM) & 86.0 \\
Belebele & 86.7 \\
\bottomrule
\end{tabular}}
\caption{\textbf{The WAWA rates for human baselines} (\S\ref{subsec:human_baselines}).}
\label{tab:wawa}
\end{table}

%% file: tables/overall.tex
\renewcommand{\arraystretch}{1.4}

\begin{table*}[h!]
    \centering
    \setlength{\tabcolsep}{2pt}
    \small
    \resizebox{\textwidth}{!}{
    \begin{tabular}{@{}l@{\hspace{2em}}c>{\columncolor[HTML]{eeeeee}}c@{\hspace{2em}}ccccccccc}
\toprule
\textbf{Model} & \rotatebox{90}{\textbf{Overall}} & \rotatebox{90}{\textbf{Borda's Count $\uparrow$}} & \rotatebox{90}{\begin{tabular}{@{}l@{}} \textbf{Norwegian language} \\[-0.5em] \textbf{knowledge} \end{tabular} } & \rotatebox{90}{\begin{tabular}{@{}l@{}} \textbf{Sentiment} \\[-0.5em] \textbf{analysis} \end{tabular}}  & \rotatebox{90}{\begin{tabular}{@{}l@{}} \textbf{Commonsense} \\[-0.5em] \textbf{reasoning} \end{tabular}}  & \rotatebox{90}{\textbf{Truthfulness}} & \rotatebox{90}{\begin{tabular}{@{}l@{}} \textbf{Norwegian-specific \&} \\[-0.5em] \textbf{world knowledge} \end{tabular} } & \rotatebox{90}{\begin{tabular}{@{}l@{}} \textbf{Machine reading} \\[-0.5em] \textbf{comprehension} \end{tabular} } & \rotatebox{90}{\begin{tabular}{@{}l@{}} \textbf{Text} \\[-0.5em] \textbf{summarization} \end{tabular} } & \rotatebox{90}{\begin{tabular}{@{}l@{}} \textbf{Instruction} \\[-0.5em] \textbf{following} \end{tabular} } & \rotatebox{90}{\begin{tabular}{@{}l@{}} \textbf{Machine} \\[-0.5em] \textbf{translation} \end{tabular} } \\
\midrule
NB-GPT-6B & \hspace{0.5em}33.0\hspace{0.5em} & \hspace{0.5em}42.0\hspace{0.5em} & 30.6 & 34.2 & 27.9 & \hspace{0.5em}33.0\hspace{0.5em} & 29.6 & 7.8 & 39.3 & \underline{39.1} & 55.1 \\
GPT-SW3-6.7B & 45.1 & 63.0 & \textbf{61.0} & 64.2 & 31.3 & \underline{43.9} & 30.0 & 30.1 & 37.7 & 35.5 & 72.6 \\
NorwAI-Mistral-7B & 45.5 & 69.0 & 47.2 & 70.7 & \underline{35.9} & 36.7 & 39.5 & 37.1 & 31.9 & 37.7 & \underline{73.2} \\
NorwAI-Llama2-7B & 44.1 & 59.0 & 47.9 & 66.3 & 29.8 & 30.2 & 35.4 & 38.8 & 37.5 & 37.7 & 72.9 \\
NorBLOOM-7B-warm & 35.6 & 28.0 & 51.8 & 40.8 & 23.5 & 39.1 & 23.3 & 23.9 & 35.6 & 13.9 & 68.8 \\
NorMistral-7B-scratch & 38.5 & 32.0 & 53.2 & 57.5 & 27.7 & 40.3 & 25.4 & 22.3 & 35.9 & 14.9 & 69.7 \\
Viking-7B & 41.9 & 47.0 & 51.3 & 59.5 & 27.4 & 26.6 & 25.0 & 25.9 & 49.4 & 38.7 & 73.0 \\
NorMistral-11B & \textbf{54.4} & \textbf{94.0} & 43.0 & \textbf{82.2} & \textbf{45.4} & 23.4 & \textbf{64.7} & \underline{59.5} & \textbf{51.7} & \textbf{46.3} & \textbf{73.4} \\
Viking-13B & 45.2 & 69.0 & 56.8 & 67.0 & 31.9 & 28.3 & 30.5 & 30.7 & 49.3 & 38.8 & 73.1 \\&&&&&&&\\[-1em]

NorMistral-7B-warm & 43.6 & 61.0 & \underline{59.2} & 68.7 & 34.0 & 31.6 & 38.7 & 40.7 & 33.0 & 14.6 & 72.0 \\
NorMistral-7B-warm-IT  & \cellcolor[rgb]{0.919,0.832,0.784}40.9 & 13.0 & \cellcolor[rgb]{0.970,0.677,0.560}\textbf{16.9} & \cellcolor[rgb]{0.681,0.788,0.991}77.2 & \cellcolor[rgb]{0.835,0.861,0.899}35.2 & \cellcolor[rgb]{0.964,0.754,0.656}24.7 & \cellcolor[rgb]{0.658,0.773,0.995}49.3 & \cellcolor[rgb]{0.970,0.684,0.568}23.4 & \cellcolor[rgb]{0.624,0.748,0.999}\underline{54.8} & \cellcolor[rgb]{0.622,0.747,1.000}\textbf{56.1} & \cellcolor[rgb]{0.970,0.677,0.560}30.5 \\&&&&&&&\\[-1.0em]

Mistral-7B & 39.7 & 38.0 & 23.4 & 77.7 & 21.1 & \textbf{46.0} & 43.5 & 47.1 & 29.5 & 11.6 & 57.5 \\
Mistral-7B-IT  & \cellcolor[rgb]{0.919,0.832,0.784}37.7 & \hphantom{0}4.0 & \cellcolor[rgb]{0.970,0.713,0.602}12.8 & \cellcolor[rgb]{0.967,0.737,0.632}69.5 & \cellcolor[rgb]{0.896,0.850,0.823}19.9 & \cellcolor[rgb]{0.970,0.693,0.578}31.9 & \cellcolor[rgb]{0.968,0.731,0.625}34.8 & \cellcolor[rgb]{0.970,0.688,0.572}31.7 & \cellcolor[rgb]{0.640,0.760,0.998}46.2 & \cellcolor[rgb]{0.622,0.747,1.000}50.4 & \cellcolor[rgb]{0.970,0.690,0.574}42.5 \\&&&&&&&\\[-1.0em]

AI-Sweden/Llama-3-8B & \underline{51.3} & \underline{84.0} & 51.0 & \underline{80.3} & 34.8 & 31.4 & 54.8 & 47.1 & 52.9 & 38.1 & 71.5 \\
AI-Sweden/Llama-3-8B-IT  & \cellcolor[rgb]{0.955,0.779,0.692}45.7 & 16.0 & \cellcolor[rgb]{0.970,0.677,0.560}\underline{16.1} & \cellcolor[rgb]{0.790,0.847,0.938}\textbf{83.2} & \cellcolor[rgb]{0.628,0.751,0.999}\textbf{53.0} & \cellcolor[rgb]{0.970,0.681,0.565}12.3 & \cellcolor[rgb]{0.853,0.864,0.880}\underline{55.3} & \cellcolor[rgb]{0.705,0.804,0.983}53.9 & \cellcolor[rgb]{0.901,0.847,0.816}48.2 & \cellcolor[rgb]{0.648,0.766,0.997}50.1 & \cellcolor[rgb]{0.970,0.677,0.560}38.9\\&&&&&&&\\[-1.0em]

Meta/Llama-3-8B & 47.0 & 64.0 & 28.4 & 76.8 & 28.0 & 34.0 & 50.9 & 48.7 & \underline{53.0} & 37.4 & 66.1 \\
Meta/Llama-3-8B-IT  & \cellcolor[rgb]{0.838,0.861,0.896}\underline{48.2} & \underline{17.0} & \cellcolor[rgb]{0.970,0.691,0.575}13.7 & \cellcolor[rgb]{0.828,0.859,0.907}78.3 & \cellcolor[rgb]{0.654,0.771,0.996}39.1 & \cellcolor[rgb]{0.729,0.818,0.974}\underline{39.5} & \cellcolor[rgb]{0.843,0.862,0.891}51.8 & \cellcolor[rgb]{0.644,0.763,0.998}\underline{61.4} & \cellcolor[rgb]{0.919,0.832,0.784}51.1 & \cellcolor[rgb]{0.639,0.759,0.998}51.4 & \cellcolor[rgb]{0.970,0.681,0.565}\textbf{47.1} \\&&&&&&&\\[-1.0em]

Mistral-Nemo-12B & 47.6 & 54.0 & 26.3 & 76.8 & 25.4 & 29.7 & \underline{55.0} & \textbf{63.4} & \underline{50.9} & 33.5 & 67.0 \\
Mistral-Nemo-12B-IT  & \cellcolor[rgb]{0.761,0.834,0.957}\textbf{52.1} & \textbf{33.0} & \cellcolor[rgb]{0.969,0.716,0.606}\underline{16.1} & \cellcolor[rgb]{0.718,0.811,0.978}\underline{82.9} & \cellcolor[rgb]{0.628,0.751,0.999}\underline{44.1} & \cellcolor[rgb]{0.643,0.762,0.998}\underline{42.7} & \cellcolor[rgb]{0.768,0.837,0.953}\textbf{58.8} & \cellcolor[rgb]{0.766,0.836,0.954}\textbf{67.3} & \cellcolor[rgb]{0.790,0.847,0.938}\textbf{57.3} & \cellcolor[rgb]{0.624,0.748,0.999}\underline{55.7} & \cellcolor[rgb]{0.970,0.679,0.561}\underline{43.7} \\
\bottomrule
\end{tabular}
}
\caption{\textbf{Borda's count and normalized performance scores} of the Norwegian LMs across all task categories in NorEval. Cold-colored cells indicate cases where the instruction-tuned version outperforms the base LM, while warm-colored cells represent cases where performance decreases after instruction-tuning. The best score is in bold, the second best is underlined -- the pretrained and instruction-tuned LMs are highlighted independently.}
\label{tab:overall}
\end{table*}

%% file: sections/results.tex
\section{Results}
\label{sec:results}
This section describes our empirical evaluation results on NorEval. We report the results aggregated across our task categories in \autoref{tab:overall}. We find that NorMistral-11B achieves the best overall performance across most task categories, followed by AI-Sweden/Llama-3-8B. NorMistral/NorBLOOM-7B-scratch and NB-GPT-6B receive the lowest scores. Mistral-Nemo-12B-IT performs best among the instruction-tuned LMs; however, the benefits from instruction-tuning depend on the task. In general, the LMs perform well on the sentiment analysis and machine translation tasks but struggle with tasks requiring the Norwegian language knowledge, commonsense reasoning, truthfulness, and instruction following. We summarize our findings below w.r.t. performance aggregation methods, human performance, task category, the effect of instruction tuning, Norwegian language variety, and LLM-as-a-judge evaluation.

\paragraph{Agreement on LM Rankings} The agreement rate\footnote{The proportion of top-k and bottom-k LMs that are consistently ranked by both performance aggregation methods.} between the average normalized score and Borda’s count for the top-3 LMs is 66\%. This discrepancy is because Borda’s count penalizes Mistral-Nemo-12B for its low performance on Norwegian language knowledge tasks, ranking NorMistral-11B and AI-Sweden/Llama-3-8B as the top-2 models, while Viking-13B takes third place instead of Mistral-Nemo-12B. However, the performance aggregation methods fully agree on the bottom-5 LMs, which include Viking-7B, Mistral-7B, NorMistral-7B-scratch, NorBLOOM-7B-warm, and NB-GPT-6B.

\paragraph{LMs vs. Human Baselines} Comparing the LMs with our human baselines in \autoref{tab:results_qa} and \autoref{tab:results_truthfulness} in \Cref{app:details}, we find that the LMs fall behind humans by 10\% on Belebele, 14.4\% on NorQuAD, 15.2\% on NorOpenBookQA, 17.8\% on NorCommonsenseQA, and 13.3\% on NorTruthfulQA Multiple Choice. However, NorwAI-Llama2-7B slightly surpasses human performance on NCB by 1.2\%. The results suggest that while LMs show promising in-context learning capabilities, there is still room for their improvement in world knowledge, truthfulness, and reading comprehension tasks.

\paragraph{Analysis on Task Categories} We outline our key results based on the fine-grained results reported in \Cref{app:details}. No single LM consistently outperforms others across all task categories. The strongest performance is observed on the sentiment analysis tasks, with AI-Sweden/Llama-3-8B achieving the best score of 92.7 and its instruction-tuned version (NoReC Document) reaching 95.5. On NorIdiom, GPT-SW3-6.7B delivers the best performance, followed by NorMistral-7B-warm. For NorCommonsenseQA, the performance of pretrained LMs varies: BM scores range from 41.2 to 61, while NN scores range from 32.6 (Mistral-7B) to 51.6 (NorMistral-11B), suggesting limited in-context learning abilities for  reasoning. The LMs also exhibit strong performance on Norwegian-specific quizzes (NRK-Quiz-QA) and tasks assessing elementary-level world knowledge (NorOpenBookQA), with the best-performing LMs including NorMistral-11B, AI-Sweden/Llama-3-8B, Mistral-7B, and Mistral-Nemo-12B. However, the LMs tend to generate less truthful answers in the open-ended QA setup (NorTruthfulQA Generation) compared to the multiple-choice setup (NorTruthfulQA Multiple Choice), highlighting potential challenges of evaluating open-ended QA in Norwegian.

\begin{figure}[t!]
  \centering
  \includegraphics[width=\columnwidth]{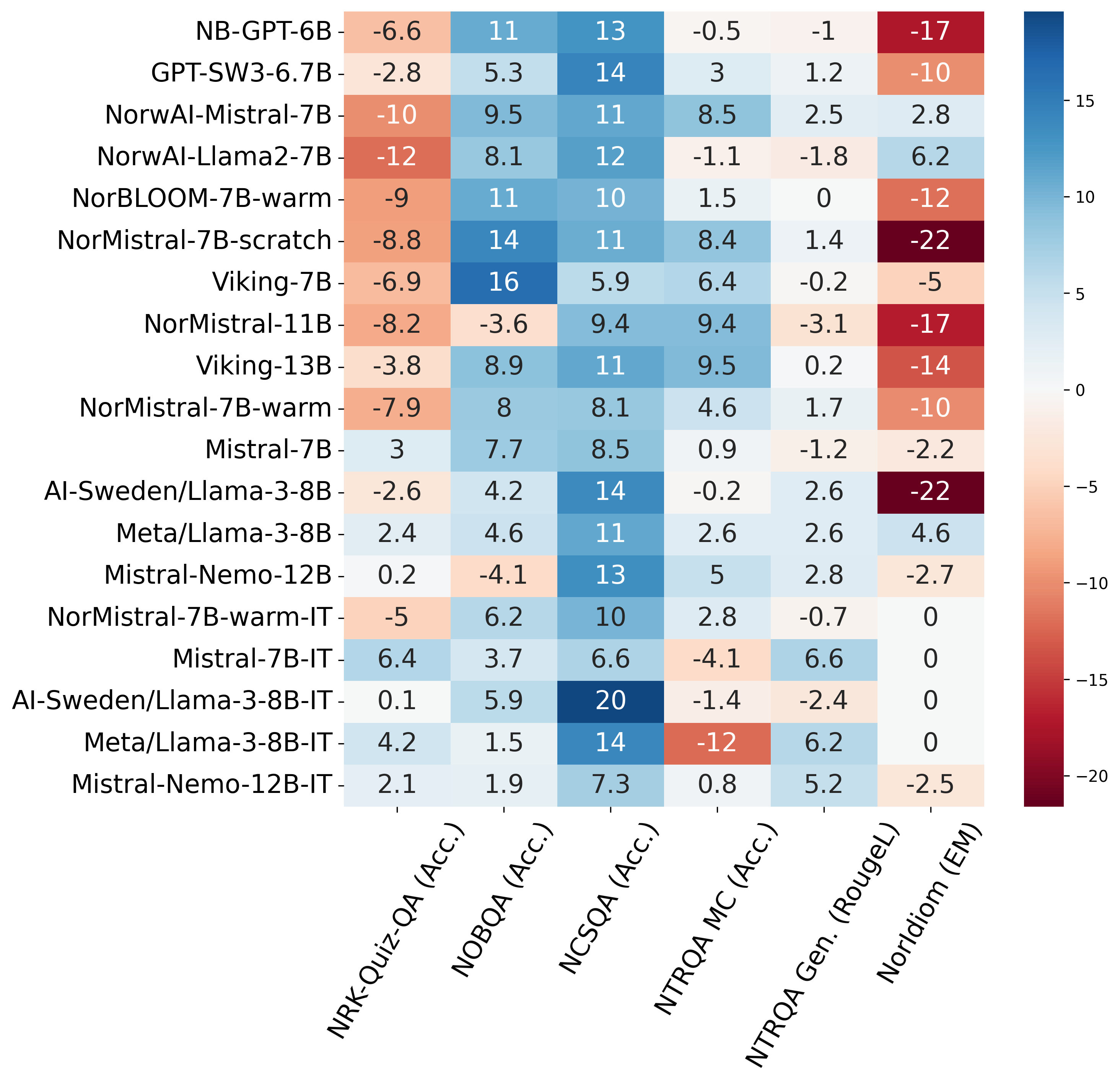} 
  \caption{\textbf{Comparison of Bokmål and Nynorsk.} Heatmap that shows the performance $\delta$-scores between BM and NN on our multiple-choice QA and sentence completion tasks. NOBQA=NorOpenBookQA; NCSQA=NorCommonsenseQA; NTRQA =NorTruthfulQA. Higher values mean higher performance in BM.}
  \label{fig:nb_vs_nn}
\end{figure}
\input{tables/manual_analysis}

\paragraph{Comparing Bokmål and Nynorsk} We compute the performance $\delta$-scores on multiple-choice and sentence completion tasks with parallel BM and NN datasets to compare LMs w.r.t. the Norwegian language variety. \autoref{fig:nb_vs_nn} shows that the LMs generally perform better on BM on NorOpenBookQA, NorCommonsenseQA, and NorTruthfulQA Multiple Choice as opposed to NRK-Quiz-QA and NorIdiom. Instruction-tuning results in lower $\delta$-scores on NRK-Quiz-QA and NorOpenBookQA but leads to random guessing performance on NorIdiom for both BM and NN.

\paragraph{Effect of Instruction-tuning} Instruction-tuning is one of the least explored research directions for Norwegian. Our results align with \citet{wang2023far,bukharin-etal-2024-data} and show that instruction-tuning can yield both positive and negative effects depending on the task. For instance, instruction-tuning consistently improves the performance of Mistral-Nemo-12B and Meta/Llama-3-8B across most task categories, with the most notable improvements observed in multiple-choice QA and sequence-to-sequence generation tasks. At the same time, it can degrade the performance on tasks requiring Norwegian language knowledge and involve translating from English into BM and NN (see \autoref{tab:results_clf_ranking_completion} and \autoref{tab:results_summ_and_mt} in \Cref{app:details}). 

\paragraph{Analysis of Unexpected Results} To better understand the negative effects of instruction-tuning, we manually analyze the outputs of 100 instruction-tuned LMs' predictions on four tasks: NorIdiom (BM), ASK-GEC, and Tatoeba (En $\rightarrow$ BM/NN). The outputs are stratified by model and task, with 25 examples per task. The main quantitative results are presented in \autoref{tab:manual_analysis}. Our analysis reveals that the instruction-tuned LMs frequently respond in English, Swedish, Danish, or a mix of these languages, even when the task requires output in BM or NN. The models also tend to produce redundant responses, often including assistant phrases such as \textit{How else can I help you?''}. Other common issues include copying parts of the input, hallucinations, and repetitive content. In 13\% of the cases, the LM understand the task but fail to produce a correct answer. We also observe that the LMs fail to interpret the NorIdiom task, leading to low scores. The results suggest that the model behavior can be improved by refining prompt design, tuning the inference hyperparameters, and explicitly specifying the target written standard in the instructions.

\input{tables/win_rates_norinstruct}

\paragraph{LLM-as-a-judge} We report the LMs' win-rates in \Cref{tab:judge}. We find that NorMistral-7B-warm-IT and Mistral-Nemo-12B-IT consistently perform best across all LMs, while responses from AI-Sweden/Llama-3-8B-IT and Mistral-7B-IT are least preferred. NorMistral-7B-warm-IT achieves the highest win-rate on NorSummarize-Instruct, while there is a minor difference between the top-2 LMs on NorRewrite-Instruct. We analyze the agreement with humans, and self-preference, position, and language biases in \Cref{app:judge}. Overall, \texttt{meta-llama/Llama-3.3-70B-Instruct} achieves a moderate agreement with humans (50\% with ties and 66.7\% without ties) and behaves fairly as a judge. We also find that there is an insignificant effect of the response position on the judge verdicts, and the instruction-tuned LMs often switch to English, Swedish, or Danish. The latter is supported by our manual analysis of the outputs (see \autoref{tab:manual_analysis}).

%% file: tables/manual_analysis.tex
\begin{table*}[htp!]
\centering
\scriptsize
\resizebox{\textwidth}{!}{
\begin{tabular}{@{}llrrrrrr@{}}
\toprule
\textbf{Task} & \textbf{$k$-shot} & \rotatebox{90}{\begin{tabular}{@{}l@{}} \textbf{Language} \\[-0.5em] \textbf{switching} \end{tabular}} & \rotatebox{90}{\begin{tabular}{@{}l@{}} \textbf{Generation} \\[-0.5em] \textbf{issues} \end{tabular}} & \rotatebox{90}{\begin{tabular}{@{}l@{}} \textbf{Input} \\[-0.5em] \textbf{copying} \end{tabular}} & \rotatebox{90}{\begin{tabular}{@{}l@{}} \textbf{Redundant} \\[-0.5em] \textbf{response}  \end{tabular}} & \rotatebox{90}{\begin{tabular}{@{}l@{}} \textbf{Instruction} \\[-0.5em] \textbf{misunderstanding} \end{tabular}} & \rotatebox{90}{\begin{tabular}{@{}l@{}} \textbf{Incorrect} \\[-0.5em] \textbf{response} \end{tabular}} \\
\midrule
NorIdiom (BM) & 0-shot & 40\% & 0\% & 8\% & 20\% & 28\% & 4\%  \\
ASK-GEC & 16-shot & 20\% & 60\% & 8\% & 0\% & 0\% & 0\%  \\
Tatoeba (En $\rightarrow$ BM) & 16-shot & 20\% & 0\% & 0\% & 40\% & 0\% & 20\% \\
Tatoeba (En $\rightarrow$ NN) & 16-shot & 12\% & 12\% & 0\% & 44\% & 0\% & 28\% \\
\hdashline
Overall & & 23\% & 18\% & 4\% & 26\% & 7\% & 13\% \\ 
\bottomrule
\end{tabular}}
\caption{\textbf{Quantitative analysis results} of the instruction-tuned LMs' predictions on Norwegian language knowledge and machine translation tasks. Remaining 9\% of the responses are manually classified as correct.}
\label{tab:manual_analysis}
\end{table*}

%% file: tables/win_rates_norinstruct.tex
\renewcommand{\arraystretch}{1.75}

\begin{table*}[t!]
\small
\resizebox{\textwidth}{!}{%
\begin{tabular}{@{}l@{\hspace{3em}}ccccccccccccc}
\toprule
& \multicolumn{6}{@{}c@{}}{\textsc{\textbf{NorRewrite-Instruct}}} &  & \multicolumn{6}{@{}c@{}}{\textsc{\textbf{NorSummarize-Instruct}}} \\[1em]
\textbf{Model} & \rotatebox{90}{NorMistral-7B-warm-IT} & \rotatebox{90}{Mistral-Nemo-12B-IT} & \rotatebox{90}{Mistral-7B-IT} & \rotatebox{90}{Meta/Llama-3-8B-IT} & \rotatebox{90}{AI-Sweden/Llama-3-8B-IT} & \hspace{-0.75em}\textbf{Average}\hspace{-0.75em} & &
\rotatebox{90}{NorMistral-7B-warm-IT} & \rotatebox{90}{Mistral-Nemo-12B-IT} & \rotatebox{90}{Mistral-7B-IT} & \rotatebox{90}{Meta/Llama-3-8B-IT} & \rotatebox{90}{AI-Sweden/Llama-3-8B-IT} & \hspace{-0.75em}\textbf{Average}\hspace{-0.75em} \\
\midrule
NorMistral-7B-warm-IT & --- & \cellcolor[rgb]{0.90,0.85,0.81}45.6 & \cellcolor[rgb]{0.43,0.57,0.95}92.2 & \cellcolor[rgb]{0.61,0.74,1.00}76.2 & \cellcolor[rgb]{0.36,0.48,0.90}99.5 & \cellcolor[rgb]{0.59,0.72,1.00}78.4 & &
--- & \cellcolor[rgb]{0.80,0.85,0.93}57.6 & \cellcolor[rgb]{0.43,0.56,0.95}92.5 & \cellcolor[rgb]{0.71,0.81,0.98}66.5 & \cellcolor[rgb]{0.36,0.48,0.90}99.5 & \cellcolor[rgb]{0.58,0.71,1.00}79.0 \\
Mistral-Nemo-12B-IT & \cellcolor[rgb]{0.83,0.86,0.90}54.4 & --- & \cellcolor[rgb]{0.46,0.60,0.96}89.8 & \cellcolor[rgb]{0.56,0.70,1.00}80.6 & \cellcolor[rgb]{0.42,0.56,0.95}93.1 & \cellcolor[rgb]{0.57,0.71,1.00}79.5 & &
\cellcolor[rgb]{0.92,0.83,0.78}42.4 & --- & \cellcolor[rgb]{0.55,0.68,0.99}81.8 & \cellcolor[rgb]{0.76,0.83,0.96}62.1 & \cellcolor[rgb]{0.49,0.62,0.98}87.3 & \cellcolor[rgb]{0.69,0.80,0.99}68.4 \\
Mistral-7B-IT & \cellcolor[rgb]{0.90,0.45,0.35}7.8 & \cellcolor[rgb]{0.92,0.49,0.38}10.2 & --- & \cellcolor[rgb]{0.89,0.85,0.84}47.4 & \cellcolor[rgb]{0.70,0.80,0.98}67.5 & \cellcolor[rgb]{0.96,0.76,0.66}33.2 & & \cellcolor[rgb]{0.90,0.45,0.35}7.5 & \cellcolor[rgb]{0.96,0.60,0.47}18.2 & --- & \cellcolor[rgb]{0.95,0.79,0.71}36.9 & \cellcolor[rgb]{0.71,0.81,0.98}66.9 & \cellcolor[rgb]{0.96,0.75,0.65}32.4 \\
Meta/Llama-3-8B-IT & \cellcolor[rgb]{0.97,0.66,0.54}23.8 & \cellcolor[rgb]{0.96,0.61,0.49}19.4 & \cellcolor[rgb]{0.85,0.86,0.89}52.6 & --- & \cellcolor[rgb]{0.73,0.82,0.97}64.7 & \cellcolor[rgb]{0.94,0.81,0.75}40.1 & &
\cellcolor[rgb]{0.96,0.76,0.67}33.5 & \cellcolor[rgb]{0.95,0.80,0.72}37.9 & \cellcolor[rgb]{0.75,0.83,0.96}63.1 & --- & \cellcolor[rgb]{0.66,0.78,0.99}71.4 & \cellcolor[rgb]{0.85,0.86,0.88}51.5 \\
AI-Sweden/Llama-3-8B-IT & \cellcolor[rgb]{0.85,0.33,0.27}0.5 & \cellcolor[rgb]{0.90,0.44,0.34}6.9 & \cellcolor[rgb]{0.96,0.75,0.65}32.5 & \cellcolor[rgb]{0.96,0.78,0.69}35.3 & --- & \cellcolor[rgb]{0.96,0.60,0.48}18.8 & &
\cellcolor[rgb]{0.85,0.33,0.27}0.5 & \cellcolor[rgb]{0.93,0.52,0.41}12.7 & \cellcolor[rgb]{0.96,0.76,0.66}33.1 & \cellcolor[rgb]{0.97,0.72,0.60}28.6 & --- & \cellcolor[rgb]{0.96,0.60,0.48}18.7 \\
\bottomrule
\end{tabular} %
}
\caption{\textbf{Pair-wise expected win-rates (\%)} of the instruction-finetuned LMs on our instruction-following tasks. Cold-colored cells indicate high win-rate, while warm-colored cells indicate low win-rate.}
\label{tab:judge}
\end{table*}

%% file: sections/conclusion.tex
\section{Conclusion and Future Work} This work introduces NorEval, the largest benchmark for assessing the LM's Norwegian language understanding and generation capabilities on 24 human-created datasets. NorEval focuses on both Norwegian language varieties and spans nine task categories, ranging from Norwegian-specific \& world knowledge to instruction following. We benchmark 19 open-source Norwegian generative LMs against each other and our established human baselines, analyzing their performance in various scenarios. Additionally, we present one of the first extensive evaluations of open Norwegian instruction-tuned LMs and their base counterparts in $k$-shot regimes, as well as via the LLM-as-a-judge approach. Our key findings indicate that the LMs struggle with tasks requiring Norwegian language knowledge, commonsense reasoning, truthfulness, and instruction following. The LMs generally perform better on BM compared to NN. Notably, instruction-tuning yields both positive and negative effects on the LM performance.

Our \emph{future} work includes: (i) a more detailed evaluation of instruction-tuned LMs and instruction-tuning data mixtures; (ii) integration of novel datasets; (iii) establishment of human baselines on additional tasks; (iv) integration of test data decontamination methods. We hope that our benchmark and evaluation framework will facilitate more comprehensive comparisons of LMs within the context of Mainland Scandinavian languages and inspire collaborative efforts among NLP researchers and developers to advance reliable LMs and evaluation resources for Norwegian.

%% file: sections/limitations.tex
\section{Limitations}

While we present extensive empirical evaluations of a broad range of Norwegian LMs, we acknowledge the following limitations of our work. 
\paragraph{Sampling Demonstrations.} In the one- and 16-shot evaluation scenarios, demonstration examples are randomly sampled, which can facilitate label bias in our text classification and multiple-choice QA tasks \cite{pmlr-v139-zhao21c}.

\paragraph{Multi-task Performance Aggregation} Aggregating evaluation results in multi-task benchmarking remains a challenging problem. We employ a combination of performance aggregation methods to mitigate the shortcomings of standard arithmetic mean aggregation: (i) score normalization to account for random baseline performance, and (ii) Borda's count to address the heterogeneity of performance metrics. However, these methods have inherent limitations. In particular, we still need to average heterogeneous task-specific normalized performance scores to compute an overall score. Although Borda's count relies on model rankings instead of performance scores, introducing a new LM can influence the final ranking due to the well-studies axiom of the independence of irrelevant alternatives \cite{arrow2012social,dougherty2020probability}. Additionally, Borda's count can treat several LMs as equivalent (or ties), which is not an empirical observation in our experiments. 

\paragraph{Potential In-domain Evaluation} Our work does not account for potential in-domain evaluation of the instruction-tuned LMs, which can be instruction-tuned on similar tasks in English and other languages, potentially inflating their downstream performance. 

\paragraph{LLM-as-a-judge} Automatic side-by-side evaluation using the LLM-as-a-judge approach is a well-established, complementary evaluation scenario that has demonstrated its efficiency for high-resource languages. However, its performance in low-resource languages remains unclear. We acknowledge that the reliability of our LLM-as-a-judge evaluation results requires further empirical validation. In particular, our analysis of agreement between the judge model and human annotators relies on the only available human preference dataset, which consists of 48 pairwise comparisons (see \Cref{app:judge}). The limited sample size may affect the generalizability of our findings. Furthermore, our analysis focuses specifically on self-preference, language, and position biases; investigating other potential evaluation directions of judge models remains outside the scope of this work.

\paragraph{Human Baselines} We find that the language models slightly surpass the human performance on NCB (see \S\ref{sec:results}). However, the results do not suggest that the models possess human-level capabilities in distinguishing between in- and correctly punctuated sentences, and evaluating them across more domains and example lengths is necessary to perform a more fine-grained performance analysis. While our annotators reach a strong inter-annotator agreement (see \S\ref{sec:setup}), we establish our human baselines on the test subsets of 50 examples only for BM. We acknowledge that increasing the number of examples could affect both the scores and agreement rates. Conducting human performance evaluation for NN could allow us to draw more conclusions regarding the human and model performance across both official written standards. This has not been done due to our limited resources, and we hope to address this in our future work.

\paragraph{Data Contamination} The increasing volume of open textual data can lead to unintended test data leakage in an LM's pretraining corpus \citep[e.g.,][]{brown2020language,dubey2024llama,zhang2024min}, which can promote the saturation of NLP benchmarks. We recognize the importance of this evaluation aspect and acknowledge that LM performance on NorEval datasets created from open text sources can be inflated. We encourage adherence to responsible LM development practices and recommend conducting test contamination analysis when benchmarking an LM on NorEval. Integrating unsupervised pretraining data detection methods into NorEval is left as a direction for our future work.

\paragraph{Evaluation Framework} NorEval is integrated into LM Harness Evaluation, a widely recognized open-source collaborative project that is subject to continuous improvements and advancements, which potentially affect its long-term compatibility, reproducibility, and usability.

%% file: sections/ethical_statement.tex
\section*{Ethics Statement}
\paragraph{Human Annotation} The hourly pay rate in our annotation projects (\S\ref{subsec:prompts} and  \Cref{app:norinstruct}) is regulated by the state and corresponds to the education level. The annotators’ submissions are stored anonymously. The annotators are warned about potentially sensitive topics in the dataset examples. 

\paragraph{Inference Costs} Evaluating an LM on NorEval does not require any finetuning. The inference costs can be minimized with the help of distributed inference libraries supported by LM Evaluation Harness, such as Accelerate \cite{accelerate} and vLLM \cite{kwon2023efficient}.

\paragraph{Potential Misuse} We acknowledge that NorEval can leak into and partially overlap with an LM's pretraining corpus. We release NorEval for research and development purposes and encourage its responsible use.

\paragraph{Transparency \& License} We release NorEval adhering to standard open-source research practices. The dataset licensing terms are provided in \autoref{tab:noreval} (see \Cref{app:design}). Our codebase is available under the MIT license. Our comprehensive documentation and full annotation guidelines are available in our GitHub repository. 

\paragraph{Use of AI-assistants} We use Grammarly\footnote{\href{https://www.grammarly.com}{\texttt{grammarly.com}}} to correct grammar, spelling, and phrasing errors in the text of this paper.

%% file: appendix/design.tex
\onecolumn
\section{NorEval: Dataset Descriptions, Examples, and Prompts}
\label{app:design}

\input{tables/noreval}

\noindent This appendix presents an overview of the 24 datasets included in NorEval  (also see \autoref{tab:noreval}).

\subsubsection*{NCB}

The Norwegian Comma Benchmark (NCB) is a collection of 840 human-written Norwegian sentence pairs. The sentences are manually collected from publicly available sources such as articles and governmental reports. The sentences aim to be representative of Norwegian non-fiction, in particular governmental prose. Each sentence pair tests one Norwegian comma rule: one sentence is correctly punctuated, while the other contains faulty comma usage. 

\begin{itemize}[itemsep=-2pt,partopsep=1ex,parsep=1ex,leftmargin=1.5em]
    \item \texttt{correct:} ``Spørsmålet om å begrense forvaltningens arbeidsbyrde ble viet stor oppmerksomhet.'' 
    \item \texttt{wrong:} ``Spørsmålet om å begrense forvaltningens arbeidsbyrde, ble viet stor oppmerksomhet.''
\end{itemize}

\paragraph{Task Formulation} Given a pair of sentences, the task is to select a correctly punctuated sentence by ranking both sentences based on their probability. The performance metric is the accuracy score.

\subsubsection*{NorIdiom}

NorIdiom is designed to evaluate an LM's knowledge of 3.5k common Norwegian idioms and phrases. Each task example consists of the first $N-1$ words of an idiom, and a list of accepted last words to complete the idiom. 

\begin{itemize}[itemsep=-2pt,partopsep=1ex,parsep=1ex,leftmargin=1.5em]
    \item \texttt{idiom\_start:} ``bite på'' 
    \item \texttt{accepted\_completions:} ``kroken'', ``agnet'' 
\end{itemize}

\paragraph{Task formulation} The task is to generate the last word of an incomplete idiom. We maximize the F1 and exact match performance scores over the list of accepted completions.

{
\vspace{1em}
\noindent\footnotesize\texttt{\textbf{Prompt A (BM and NN):}}\vspace{-0.5em}
\begin{minted}[linenos=true, breaklines=true, baselinestretch=1.2, bgcolor=bg, breakanywhere=true, fontfamily=tt, fontsize=\footnotesize, xleftmargin=2em]{genshi}
Fullfør dette uttrykket: {{idiom_start}}
\end{minted}

\noindent\footnotesize\texttt{\textbf{Prompt B (BM):}}\vspace{-0.5em}
\begin{minted}[linenos=true, breaklines=true, baselinestretch=1.2, bgcolor=bg, breakanywhere=true, fontfamily=tt, fontsize=\footnotesize, xleftmargin=2em]{genshi}
Skriv fortsettelsen av idiomet {{idiom_start}}
\end{minted}

\noindent\footnotesize\texttt{\textbf{Prompt B (NN):}}\vspace{-0.5em}
\begin{minted}[linenos=true, breaklines=true, baselinestretch=1.2, bgcolor=bg, breakanywhere=true, fontfamily=tt, fontsize=\footnotesize, xleftmargin=2em]{genshi}
Skriv fortsetjinga av idiomet {{idiom_start}}
\end{minted}

\noindent\footnotesize\texttt{\textbf{Prompt C (BM):}}\vspace{-0.5em}
\begin{minted}[linenos=true, breaklines=true, baselinestretch=1.2, bgcolor=bg, breakanywhere=true, fontfamily=tt, fontsize=\footnotesize, xleftmargin=2em]{genshi}
Hvordan fortsetter uttrykket "{{idiom_start}}"?
\end{minted}

\noindent\footnotesize\texttt{\textbf{Prompt  C (NN):}}\vspace{-0.5em}
\begin{minted}[linenos=true, breaklines=true, baselinestretch=1.2, bgcolor=bg, breakanywhere=true, fontfamily=tt, fontsize=\footnotesize, xleftmargin=2em]{genshi}
Korleis fortset uttrykket "{{idiom_start}}"?
\end{minted}

\noindent\footnotesize\texttt{\textbf{Prompt D (BM):}}\vspace{-0.5em}
\begin{minted}[linenos=true, breaklines=true, baselinestretch=1.2, bgcolor=bg, breakanywhere=true, fontfamily=tt, fontsize=\footnotesize, xleftmargin=2em]{genshi}
Fullfør vendingen "{{idiom_start}}"
\end{minted}

\noindent\footnotesize\texttt{\textbf{Prompt D (NN):}}\vspace{-0.5em}
\begin{minted}[linenos=true, breaklines=true, baselinestretch=1.2, bgcolor=bg, breakanywhere=true, fontfamily=tt, fontsize=\footnotesize, xleftmargin=2em]{genshi}
Fullfør vendinga: {{idiom_start}}
\end{minted}

\noindent\footnotesize\texttt{\textbf{Prompt E (BM and NN):}}\vspace{-0.5em}
\begin{minted}[linenos=true, breaklines=true, baselinestretch=1.2, bgcolor=bg, breakanywhere=true, fontfamily=tt, fontsize=\footnotesize, xleftmargin=2em]{genshi}
{{idiom_start}}
\end{minted}
}

\subsection*{Belebele}

Belebele is a multiple-choice QA dataset spanning 122 language variants. Each question has four multiple-choice answers a short passage. 

\paragraph{Task Formulation} The task is to select a correct answer option given a passage and a question. The performance metric is the accuracy score.

\begin{itemize}[itemsep=-2pt,partopsep=1ex,parsep=1ex,leftmargin=1.5em]
    \item \texttt{passage:} ``Så og si nesten alle PC-er som benyttes i dag, baseres på manipulering av informasjon som er kodet med binære tall. Et binært tall kan kun ha én av to verdier, dvs. 0 eller 1. Disse tallene omtales som binærsifre – eller biter, for å bruke datasjargon.'' 
    \item \texttt{question:} ``Hvilke av følgende er et eksempel et binært tall med fem biter, ifølge avsnittet?'' 
    \item \texttt{answer\_1:} \texttt{1010}
    \item \texttt{answer\_2:} \texttt{12001}
    \item \texttt{answer\_3:} \texttt{10010}
    \item \texttt{answer\_4:}\texttt{110101}
    \item \texttt{correct\_answer\_num:} 3
\end{itemize}

{
\vspace{1em}
\noindent\footnotesize\texttt{\textbf{Prompt A:}}\vspace{-0.5em}
\begin{minted}[linenos=true, breaklines=true, baselinestretch=1.2, bgcolor=bg, breakanywhere=true, fontfamily=tt, fontsize=\footnotesize, xleftmargin=2em]{genshi}
Tekst: {{passage}}
Spørsmål: {{question}}
A: {{answer_1}}
B: {{answer_2}}
C: {{answer_3}}
D: {{answer_4}}
Svar: {prediction:A/B/C/D}
\end{minted}

\noindent\footnotesize\texttt{\textbf{Prompt B:}}\vspace{-0.5em}
\begin{minted}[linenos=true, breaklines=true, baselinestretch=1.2, bgcolor=bg, breakanywhere=true, fontfamily=tt, fontsize=\footnotesize, xleftmargin=2em]{genshi}
Bakgrunn: {{passage}}
Spørsmål: {{question}}
Svaralternativer:
- {{answer_1}}
- {{answer_2}}
- {{answer_3}}
- {{answer_4}}
Svar: {prediction:{{answer_1}}/{{answer_2}}/{{answer_3}}/{{answer_4}}}
\end{minted}

\noindent\footnotesize\texttt{\textbf{Prompt C:}}\vspace{-0.5em}
\begin{minted}[linenos=true, breaklines=true, baselinestretch=1.2, bgcolor=bg, breakanywhere=true, fontfamily=tt, fontsize=\footnotesize, xleftmargin=2em]{genshi}
{{question}}
Hvilket av følgende mulige svar er det riktige?
A: {{answer_1}}
B: {{answer_2}}
C: {{answer_3}}
D: {{answer_4}}
Svar: {prediction:A/B/C/D}
\end{minted}

\noindent\footnotesize\texttt{\textbf{Prompt D:}}\vspace{-0.5em}
\begin{minted}[linenos=true, breaklines=true, baselinestretch=1.2, bgcolor=bg, breakanywhere=true, fontfamily=tt, fontsize=\footnotesize, xleftmargin=2em]{genshi}
Svar på følgende spørsmål: {{question}}
Svaret skal baseres på følgende tekst:
{{passage}}
Velg et svar fra denne listen:
- {{answer_1}}
- {{answer_2}}
- {{answer_3}}
- {{answer_4}}
Svar: {prediction:{{answer_1}}/{{answer_2}}/{{answer_3}}/{{answer_4}}}
\end{minted}

\noindent\footnotesize\texttt{\textbf{Prompt E:}}\vspace{-0.5em}
\begin{minted}[linenos=true, breaklines=true, baselinestretch=1.2, bgcolor=bg, breakanywhere=true, fontfamily=tt, fontsize=\footnotesize, xleftmargin=2em]{genshi}
{{passage}}

{{question}}

A: {{answer_1}}
B: {{answer_2}}
C: {{answer_3}}
D: {{answer_4}}

Er det riktige svaret A, B, C, eller D? {prediction:A/B/C/D}
\end{minted}
}

\newpage 

\subsection*{NorQuAD}

NorQuAD consists of 4.7k manually created examples based on Wikipedia and news articles following the SQuAD design  \cite{rajpurkar-etal-2016-squad}.

\begin{itemize}[itemsep=-2pt,partopsep=1ex,parsep=1ex,leftmargin=1.5em]
    \item \texttt{title:} ``Ordspråk
    \item \texttt{context:} ``Ordspråk eller ordtak er korte, velformulerte og poengterte setninger som på en konkret måte uttrykker livsvisdom, allmenngyldige sannheter, erfaringer, leveregler eller betraktninger av forskjellig slag. Ordspråk kan også inneholde forklaringer av naturfenomener, skikker og seder. Ordspråk har en fast ordlyd som er kjent og blir sitert, for eksempel for å kommentere noe eller for å gi et råd. Mange ordspråk har uklar opprinnelse og er en del av gammel folkediktning og en muntlig fortellertradisjon. Det er også mange som er sitater fra bøker og fortellinger med kjent opphav, for eksempel fra Bibelen og Håvamål, selv om begrepet ordspråk ofte brukes om folkelige uttrykk uten kjent forfatter. Ordspråk kan være internasjonale, nasjonale og regionale og finnes i et nærmest uendelig antall og i en mengde varianter over hele verden. Studiet av ordspråk kalles parømiologi. Også fraseologien beskriver etablerte flerordsenheter og -forbindelser i et språk, særlig faste uttrykk og idiomer, men også tekster som ordspråk.'' 
    \item \texttt{question:} ``Hvordan er opprinnelsen til mange ordspråk?'' 
    \item \texttt{answer:} ``uklar''
\end{itemize}

\paragraph{Task Formulation} The task is to extract the answer from the context given a question. We formulate it as a sequence-to-sequence problem, where the LM receives the context and the question as the input and is expected to generate the answer. The performance metrics are exact match (the percentage of predictions that exactly match the gold answer) and F1-score (the average N-gram overlap between the prediction and the gold answer treated as bag-of-words). 

{
\vspace{1em}
\noindent\footnotesize\texttt{\textbf{Prompt A:}}\vspace{-0.5em}
\begin{minted}[linenos=true, breaklines=true, baselinestretch=1.2, bgcolor=bg, breakanywhere=true, fontfamily=tt, fontsize=\footnotesize, xleftmargin=2em]{genshi}
Tittel: {{title}}

Tekst: {{passage}}

Spørsmål: {{question}}

Svar: {{prediction}}
\end{minted}

\noindent\footnotesize\texttt{\textbf{Prompt B:}}\vspace{-0.5em}
\begin{minted}[linenos=true, breaklines=true, baselinestretch=1.2, bgcolor=bg, breakanywhere=true, fontfamily=tt, fontsize=\footnotesize, xleftmargin=2em]{genshi}
Tittel: {{title}}

Tekst: {{passage}}

Gitt teksten over, hva er svaret på følgende spørsmål? "{{question}}"

Svar: {{prediction}}
\end{minted}

\noindent\footnotesize\texttt{\textbf{Prompt C:}}\vspace{-0.5em}
\begin{minted}[linenos=true, breaklines=true, baselinestretch=1.2, bgcolor=bg, breakanywhere=true, fontfamily=tt, fontsize=\footnotesize, xleftmargin=2em]{genshi}
Tittel: {{title}}

Tekst: {{passage}}

Svar på følgende: {{question}}

Svar: {{prediction}}
\end{minted}

\newpage 

\noindent\footnotesize\texttt{\textbf{Prompt D:}}\vspace{-0.5em}
\begin{minted}[linenos=true, breaklines=true, baselinestretch=1.2, bgcolor=bg, breakanywhere=true, fontfamily=tt, fontsize=\footnotesize, xleftmargin=2em]{genshi}
Tittel: {{title}}

Tekst: {{passage}}

Hvordan kan man svare på spørsmålet "{{question}}", gitt teksten over?

Svar:{{prediction}}
\end{minted}

\noindent\footnotesize\texttt{\textbf{Prompt E:}}\vspace{-0.5em}
\begin{minted}[linenos=true, breaklines=true, baselinestretch=1.2, bgcolor=bg, breakanywhere=true, fontfamily=tt, fontsize=\footnotesize, xleftmargin=2em]{genshi}
Tittel: {{title}}

Tekst: {{passage}}

Gitt teksten over, besvar følgende spørsmål: "{{question}}"

Svar: {{prediction}}
\end{minted}
}

\subsection*{NoReC Sentence}

NoReC Sentence is a dataset for sentence-level sentiment analysis in Norwegian, derived from NoReC\_fine \cite{ovrelid-etal-2020-fine}. The annotations have been aggregated at the sentence-level, by only keeping sentences that contain sentiment annotations of either positive or negative polarity. 

\paragraph{Task Formulation} The task is framed as a binary classification problem. The LM is required to predict if a given review has a positive or negative sentiment. The target performance metric is the macro-average F1-score.

\begin{itemize}[itemsep=-2pt,partopsep=1ex,parsep=1ex,leftmargin=1.5em]
    \item \texttt{review:} ``En mer allsidig og tilkoblingsvennlig skjerm har vi knapt sett .'' 
    \item \texttt{sentiment:} \texttt{1} (positive).
\end{itemize}

{
\vspace{1em}
\noindent\footnotesize\texttt{\textbf{Prompt A:}}\vspace{-0.5em}
\begin{minted}[linenos=true, breaklines=true, baselinestretch=1.2, bgcolor=bg, breakanywhere=true, fontfamily=tt, fontsize=\footnotesize, xleftmargin=2em]{genshi}
Tekst: {{text}}
Sentiment: {prediction:positiv/negativ}
\end{minted}

\noindent\footnotesize\texttt{\textbf{Prompt B:}}\vspace{-0.5em}
\begin{minted}[linenos=true, breaklines=true, baselinestretch=1.2, bgcolor=bg, breakanywhere=true, fontfamily=tt, fontsize=\footnotesize, xleftmargin=2em]{genshi}
{{text}}
Er denne setningen "positiv" eller "negativ"? {prediction:positiv/negativ}
\end{minted}

\noindent\footnotesize\texttt{\textbf{Prompt C:}}\vspace{-0.5em}
\begin{minted}[linenos=true, breaklines=true, baselinestretch=1.2, bgcolor=bg, breakanywhere=true, fontfamily=tt, fontsize=\footnotesize, xleftmargin=2em]{genshi}
{{text}}
Hva slags sentiment uttrykker anmelderen? {prediction:positiv/negativ}
\end{minted}

\noindent\footnotesize\texttt{\textbf{Prompt D:}}\vspace{-0.5em}
\begin{minted}[linenos=true, breaklines=true, baselinestretch=1.2, bgcolor=bg, breakanywhere=true, fontfamily=tt, fontsize=\footnotesize, xleftmargin=2em]{genshi}
{{text}}
Er anmeldelsen "positiv" eller "negativ"? {prediction:positiv/negativ}
\end{minted}

\noindent\footnotesize\texttt{\textbf{Prompt E:}}\vspace{-0.5em}
\begin{minted}[linenos=true, breaklines=true, baselinestretch=1.2, bgcolor=bg, breakanywhere=true, fontfamily=tt, fontsize=\footnotesize, xleftmargin=2em]{genshi}
{{text}}
Er denne setningen positiv eller negativ? {prediction:positiv/negativ}
\end{minted}
}

\subsection*{NoReC Document} NoReC Document is a dataset for document-level sentiment analysis derived from NoReC \cite{velldal-etal-2018-norec} by keeping documents that have positive (ratings 5--6) or negative (ratings 1--3) sentiment.

\paragraph{Task Formulation} The task is framed as a binary classification problem. The LM is required to predict if a given review has a positive or negative sentiment. The target performance metric is the macro-average F1-score. 

{
\vspace{1em}
\noindent\footnotesize\texttt{\textbf{Prompt A:}}\vspace{-0.5em}
\begin{minted}[linenos=true, breaklines=true, baselinestretch=1.2, bgcolor=bg, breakanywhere=true, fontfamily=tt, fontsize=\footnotesize, xleftmargin=2em]{genshi}
Tekst: {{text}}
Sentiment: {prediction:positiv/negativ}
\end{minted}

\noindent\footnotesize\texttt{\textbf{Prompt B:}}\vspace{-0.5em}
\begin{minted}[linenos=true, breaklines=true, baselinestretch=1.2, bgcolor=bg, breakanywhere=true, fontfamily=tt, fontsize=\footnotesize, xleftmargin=2em]{genshi}
Tekst: {{text}}
Er anmeldelsen "positiv" eller "negativ"? {prediction:positiv/negativ}
\end{minted}

\noindent\footnotesize\texttt{\textbf{Prompt C:}}\vspace{-0.5em}
\begin{minted}[linenos=true, breaklines=true, baselinestretch=1.2, bgcolor=bg, breakanywhere=true, fontfamily=tt, fontsize=\footnotesize, xleftmargin=2em]{genshi}
Er polariteten til følgende anmeldelse positiv eller negativ?
Anmeldelse: {{text}}
Anmeldelsen er {prediction:positiv/negativ}
\end{minted}

\noindent\footnotesize\texttt{\textbf{Prompt D:}}\vspace{-0.5em}
\begin{minted}[linenos=true, breaklines=true, baselinestretch=1.2, bgcolor=bg, breakanywhere=true, fontfamily=tt, fontsize=\footnotesize, xleftmargin=2em]{genshi}
Anmeldelse: {{text}}
Er anmelderen positiv eller negativ? {prediction:positiv/negativ}
\end{minted}

\noindent\footnotesize\texttt{\textbf{Prompt E:}}\vspace{-0.5em}
\begin{minted}[linenos=true, breaklines=true, baselinestretch=1.2, bgcolor=bg, breakanywhere=true, fontfamily=tt, fontsize=\footnotesize, xleftmargin=2em]{genshi}
Anmeldelse: {{text}}
Vil du oppsummere anmeldelsen som "bra" eller "dårlig"? {prediction:bra/dårlig}
\end{minted}
}

\subsection*{NorCommonsenseQA}
NorCommonsenseQA is developed to assess the LM's commonsense reasoning abilities. It includes 1.1k examples in BM and NN, each comprising a question and five answer choices.

\begin{itemize}[itemsep=-2pt,partopsep=0.5ex,parsep=1ex,leftmargin=1.5em] 

\item \texttt{question:} ``Hvis statsministeren ønsket å forby slanger, hvor ville han foreslått lovforslaget?'' 
\item \texttt{answer\_1:} ``På gata''
\item \texttt{answer\_2:} ``I en tropisk skog''
\item \texttt{answer\_3:}  ``I Edens hage''
\item \texttt{answer\_4:}  ``På Eidsvoll''
\item \texttt{answer\_5:} ``I Stortinget'' (correct)
\end{itemize}

\paragraph{Task Formulation} The task is to select a correct answer to given a question. The performance metric is the accuracy score.

{
\vspace{1em}
\noindent\footnotesize\texttt{\textbf{Prompt A (BM and NN):}}\vspace{-0.5em}
\begin{minted}[linenos=true, breaklines=true, baselinestretch=1.2, bgcolor=bg, breakanywhere=true, fontfamily=tt, fontsize=\footnotesize, xleftmargin=2em]{genshi}
Spørsmål: {{question}}

Svar: {prediction:{{answer_1}}/{{answer_2}}/{{answer_3}}/{{answer_4}}/{{answer_5}}}
\end{minted}

\noindent\footnotesize\texttt{\textbf{Prompt B (BM):}}\vspace{-0.5em}
\begin{minted}[linenos=true, breaklines=true, baselinestretch=1.2, bgcolor=bg, breakanywhere=true, fontfamily=tt, fontsize=\footnotesize, xleftmargin=2em]{genshi}
{{question}}
Hvilket av følgende mulige svar er det riktige?
A: {{answer_1}}
B: {{answer_2}}
C: {{answer_3}}
D: {{answer_4}}
E: {{answer_5}}
Svar: {prediction:A/B/C/D/E}
\end{minted}

\noindent\footnotesize\texttt{\textbf{Prompt B (NN):}}\vspace{-0.5em}
\begin{minted}[linenos=true, breaklines=true, baselinestretch=1.2, bgcolor=bg, breakanywhere=true, fontfamily=tt, fontsize=\footnotesize, xleftmargin=2em]{genshi}
{{question}}
Kva av følgande moglege svar er det rette?
A: {{answer_1}}
B: {{answer_2}}
C: {{answer_3}}
D: {{answer_4}}
E: {{answer_5}}
Svar: {prediction:A/B/C/D/E}
\end{minted}

\noindent\footnotesize\texttt{\textbf{Prompt C (BM):}}\vspace{-0.5em}
\begin{minted}[linenos=true, breaklines=true, baselinestretch=1.2, bgcolor=bg, breakanywhere=true, fontfamily=tt, fontsize=\footnotesize, xleftmargin=2em]{genshi}
Gitt alternativene under, hva er svaret på følgende spørsmål: {{question}}

Alternativer:
- {{answer_1}}
- {{answer_2}}
- {{answer_3}}
- {{answer_4}}
- {{answer_5}}

Svar: {prediction:{{answer_1}}/{{answer_2}}/{{answer_3}}/{{answer_4}}/{{answer_5}}}
\end{minted}

\noindent\footnotesize\texttt{\textbf{Prompt C (NN):}}\vspace{-0.5em}
\begin{minted}[linenos=true, breaklines=true, baselinestretch=1.2, bgcolor=bg, breakanywhere=true, fontfamily=tt, fontsize=\footnotesize, xleftmargin=2em]{genshi}
Gitt alternativa under, kva er svaret på følgande spørsmål: {{question}}

Alternativ:
- {{answer_1}}
- {{answer_2}}
- {{answer_3}}
- {{answer_4}}
- {{answer_5}}

Svar: {prediction:A/B/C/D/E}
\end{minted}

\noindent\footnotesize\texttt{\textbf{Prompt D (BM):}}\vspace{-0.5em}
\begin{minted}[linenos=true, breaklines=true, baselinestretch=1.2, bgcolor=bg, breakanywhere=true, fontfamily=tt, fontsize=\footnotesize, xleftmargin=2em]{genshi}
{{question}}
Velg riktig svar blant disse alternativene:
- {{answer_1}}
- {{answer_2}}
- {{answer_3}}
- {{answer_4}}
- {{answer_5}}

Svar: {prediction:{{answer_1}}/{{answer_2}}/{{answer_3}}/{{answer_4}}/{{answer_5}}}
\end{minted}

\noindent\footnotesize\texttt{\textbf{Prompt D (NN):}}\vspace{-0.5em}
\begin{minted}[linenos=true, breaklines=true, baselinestretch=1.2, bgcolor=bg, breakanywhere=true, fontfamily=tt, fontsize=\footnotesize, xleftmargin=2em]{genshi}
{{question}}
Vel rett svar blant desse alternativa:
- {{answer_1}}
- {{answer_2}}
- {{answer_3}}
- {{answer_4}}
- {{answer_5}}

Svar: {prediction:{{answer_1}}/{{answer_2}}/{{answer_3}}/{{answer_4}}/{{answer_5}}}
\end{minted}

\newpage

\noindent\footnotesize\texttt{\textbf{Prompt E (BM):}}\vspace{-0.5em}
\begin{minted}[linenos=true, breaklines=true, baselinestretch=1.2, bgcolor=bg, breakanywhere=true, fontfamily=tt, fontsize=\footnotesize, xleftmargin=2em]{genshi}
{{question}}
A: {{answer_1}}
B: {{answer_2}}
C: {{answer_3}}
D: {{answer_4}}
E: {{answer_5}}

Er det riktige svaret A, B, C, D, eller E?

Svar: {prediction:A/B/C/D/E}
\end{minted}

\noindent\footnotesize\texttt{\textbf{Prompt E (NN):}}\vspace{-0.5em}
\begin{minted}[linenos=true, breaklines=true, baselinestretch=1.2, bgcolor=bg, breakanywhere=true, fontfamily=tt, fontsize=\footnotesize, xleftmargin=2em]{genshi}
{{question}}
A: {{answer_1}}
B: {{answer_2}}
C: {{answer_3}}
D: {{answer_4}}
E: {{answer_5}}

Er det rette svaret A, B, C, D, eller E?

Svar: {prediction:A/B/C/D/E}
\end{minted}
}

\subsection*{NRK-Quiz-QA}

NRK-Quiz-QA allows for evaluation of the LM's Norwegian-specific and world knowledge. NRK-Quiz-QA includes 4.9k examples in BM and NN from more than 500 quizzes covering various topics on the Norwegian language and culture. Each example contains a question and 2 to 5 answer choices.

\begin{itemize}[itemsep=-2pt,partopsep=0.25ex,parsep=1ex,leftmargin=1.5em]
\item \texttt{question:} \textit{``Æ træng læsta: Læsta er kjekt å ha. I alle fall sånn innimellom. Men hva er det for noe?''}
\item \texttt{answer\_1:} ``Venner''
\item \texttt{answer\_2:} ``Lesestoff''
\item \texttt{answer\_3:} ``Ro''
\item \texttt{answer\_4:} ``Ullsokker'' (correct)
\end{itemize}

\paragraph{Task Formulation} The task is to select a correct answer to given a question. The performance metric is the accuracy score.

{
\vspace{1em}
\noindent\footnotesize\texttt{\textbf{Prompt A (BM and NN):}}\vspace{-0.5em}
\begin{minted}[linenos=true, breaklines=true, baselinestretch=1.2, bgcolor=bg, breakanywhere=true, fontfamily=tt, fontsize=\footnotesize, xleftmargin=2em]{genshi}
Spørsmål: {{question}}

Svar: {prediction:{{answer_1}}/{{answer_2}}/{{answer_3}}/{{answer_4}}}
\end{minted}

\newpage 

\noindent\footnotesize\texttt{\textbf{Prompt B (BM):}}\vspace{-0.5em}
\begin{minted}[linenos=true, breaklines=true, baselinestretch=1.2, bgcolor=bg, breakanywhere=true, fontfamily=tt, fontsize=\footnotesize, xleftmargin=2em]{genshi}
{{question}}

Svaralternativer:
- {{answer_1}}
- {{answer_2}}
- {{answer_3}}
- {{answer_4}}

Hva er riktig svar?

Svar: {prediction:{{answer_1}}/{{answer_2}}/{{answer_3}}/{{answer_4}}}
\end{minted}

\noindent\footnotesize\texttt{\textbf{Prompt B (NN):}}\vspace{-0.5em}
\begin{minted}[linenos=true, breaklines=true, baselinestretch=1.2, bgcolor=bg, breakanywhere=true, fontfamily=tt, fontsize=\footnotesize, xleftmargin=2em]{genshi}
{{question}}
{{question}}

Svaralternativer:
- {{answer_1}}
- {{answer_2}}
- {{answer_3}}
- {{answer_4}}

Kva er rett svar?

Svar: {prediction:{{answer_1}}/{{answer_2}}/{{answer_3}}/{{answer_4}}}
\end{minted}

\noindent\footnotesize\texttt{\textbf{Prompt C (BM):}}\vspace{-0.5em}
\begin{minted}[linenos=true, breaklines=true, baselinestretch=1.2, bgcolor=bg, breakanywhere=true, fontfamily=tt, fontsize=\footnotesize, xleftmargin=2em]{genshi}
{{question}}
A: {{answer_1}}
B: {{answer_2}}
C: {{answer_3}}
D: {{answer_4}}

Er det riktige svaret A, B, C, eller D? 

Svar: {prediction:A/B/C/D}
\end{minted}

\noindent\footnotesize\texttt{\textbf{Prompt C (NN):}}\vspace{-0.5em}
\begin{minted}[linenos=true, breaklines=true, baselinestretch=1.2, bgcolor=bg, breakanywhere=true, fontfamily=tt, fontsize=\footnotesize, xleftmargin=2em]{genshi}
{{question}}
A: {{answer_1}}
B: {{answer_2}}
C: {{answer_3}}
D: {{answer_4}}

Er det rette svare A, B, C, eller D? 

Svar: {prediction:A/B/C/D}
\end{minted}

\noindent\footnotesize\texttt{\textbf{Prompt D (BM and NN):}}\vspace{-0.5em}
\begin{minted}[linenos=true, breaklines=true, baselinestretch=1.2, bgcolor=bg, breakanywhere=true, fontfamily=tt, fontsize=\footnotesize, xleftmargin=2em]{genshi}
Spørsmål: {{question}}
A: {{answer_1}}
B: {{answer_2}}
C: {{answer_3}}
D: {{answer_4}}

Svar: {prediction:A/B/C/D}
\end{minted}

\noindent\footnotesize\texttt{\textbf{Prompt E (BM):}}\vspace{-0.5em}
\begin{minted}[linenos=true, breaklines=true, baselinestretch=1.2, bgcolor=bg, breakanywhere=true, fontfamily=tt, fontsize=\footnotesize, xleftmargin=2em]{genshi}
{{question}}
Velg riktig svar blant disse alternativene:
- {{answer_1}}
- {{answer_2}}
- {{answer_3}}
- {{answer_4}}

Svar: {prediction:{{answer_1}}/{{answer_2}}/{{answer_3}}/{{answer_4}}}
\end{minted}

\noindent\footnotesize\texttt{\textbf{Prompt E (NN):}}\vspace{-0.5em}
\begin{minted}[linenos=true, breaklines=true, baselinestretch=1.2, bgcolor=bg, breakanywhere=true, fontfamily=tt, fontsize=\footnotesize, xleftmargin=2em]{genshi}
{{question}}
Vel rett svar blant desse alternativa:
- {{answer_1}}
- {{answer_2}}
- {{answer_3}}
- {{answer_4}}

Svar: {prediction:{{answer_1}}/{{answer_2}}/{{answer_3}}/{{answer_4}}}
\end{minted}
}

\subsection*{NorOpenBookQA}

NorOpenBookQA is designed to evaluate the LM's world knowledge. NorOpenBookQA counts 3.5k examples in BM and NN, each consisting of an elementary-level science question, four answer choices, and a factual statement that presents the evidence necessary to determine the correct answer. 

\begin{itemize}[itemsep=-2pt,partopsep=0.25ex,parsep=1ex,leftmargin=1.5em]
\item \texttt{question:} \textit{``Hva er mykest?''} 
\item \texttt{answer\_1:} ``Marshmallows''
\item \texttt{answer\_1:} ``Stål''
\item \texttt{answer\_1:} ``Diamant''
\item \texttt{answer\_1:} ``Saltstenger''
\item \texttt{fact:} ``Et mineral som kan skrapes av en fingernegl regnes som mykt''
\end{itemize}

\paragraph{Task Formulation} The task is to select a correct answer to given a question. The performance metric is the accuracy score.

{
\vspace{1em}
\noindent\footnotesize\texttt{\textbf{Prompt A (BM and NN):}}\vspace{-0.5em}
\begin{minted}[linenos=true, breaklines=true, baselinestretch=1.2, bgcolor=bg, breakanywhere=true, fontfamily=tt, fontsize=\footnotesize, xleftmargin=2em]{genshi}
{{fact}}
{{question}} {prediction:{{answer_1}}/{{answer_2}}/{{answer_3}}/{{answer_4}}}
\end{minted}

\noindent\footnotesize\texttt{\textbf{Prompt B (BM):}}\vspace{-0.5em}
\begin{minted}[linenos=true, breaklines=true, baselinestretch=1.2, bgcolor=bg, breakanywhere=true, fontfamily=tt, fontsize=\footnotesize, xleftmargin=2em]{genshi}
Faktatekst: {{fact}}
Spørsmål til teksten: {{question}}

Svaralternativer:
- {{answer_1}}
- {{answer_2}}
- {{answer_3}}
- {{answer_4}}

Hva er riktig svar? {prediction:{{answer_1}}/{{answer_2}}/{{answer_3}}/{{answer_4}}}
\end{minted}

\newpage

\noindent\footnotesize\texttt{\textbf{Prompt B (NN):}}\vspace{-0.5em}
\begin{minted}[linenos=true, breaklines=true, baselinestretch=1.2, bgcolor=bg, breakanywhere=true, fontfamily=tt, fontsize=\footnotesize, xleftmargin=2em]{genshi}
Faktatekst: {{fact}}
Spørsmål til teksten: {{question}}

Svaralternativer:
- {{answer_1}}
- {{answer_2}}
- {{answer_3}}
- {{answer_4}}

Kva er rett svar? {prediction:{{answer_1}}/{{answer_2}}/{{answer_3}}/{{answer_4}}}
\end{minted}

\noindent\footnotesize\texttt{\textbf{Prompt C (BM):}}\vspace{-0.5em}
\begin{minted}[linenos=true, breaklines=true, baselinestretch=1.2, bgcolor=bg, breakanywhere=true, fontfamily=tt, fontsize=\footnotesize, xleftmargin=2em]{genshi}
{{fact}}
{{question}}
A: {{answer_1}}
B: {{answer_2}}
C: {{answer_3}}
D: {{answer_4}}

Er det riktige svaret A, B, C, eller D?

Svar: {prediction:A/B/C/D}
\end{minted}

\noindent\footnotesize\texttt{\textbf{Prompt C (NN):}}\vspace{-0.5em}
\begin{minted}[linenos=true, breaklines=true, baselinestretch=1.2, bgcolor=bg, breakanywhere=true, fontfamily=tt, fontsize=\footnotesize, xleftmargin=2em]{genshi}
{{fact}}
{{question}}
A: {{answer_1}}
B: {{answer_2}}
C: {{answer_3}}
D: {{answer_4}}

Er det rette svare A, B, C, eller D? 

Svar: {prediction:A/B/C/D}
\end{minted}

\noindent\footnotesize\texttt{\textbf{Prompt D (BM and NN):}}\vspace{-0.5em}
\begin{minted}[linenos=true, breaklines=true, baselinestretch=1.2, bgcolor=bg, breakanywhere=true, fontfamily=tt, fontsize=\footnotesize, xleftmargin=2em]{genshi}
Bakgrunn: {{fact}}

Spørsmål: {{question}}
A: {{answer_1}}
B: {{answer_2}}
C: {{answer_3}}
D: {{answer_4}}

Svar: {prediction:A/B/C/D}
\end{minted}

\noindent\footnotesize\texttt{\textbf{Prompt E (BM):}}\vspace{-0.5em}
\begin{minted}[linenos=true, breaklines=true, baselinestretch=1.2, bgcolor=bg, breakanywhere=true, fontfamily=tt, fontsize=\footnotesize, xleftmargin=2em]{genshi}
Ta utgangspunkt i følgende fakta når du svarer på spørsmålet: {{fact}}

{{question}}
Velg riktig svar blant disse alternativene:
- {{answer_1}}
- {{answer_2}}
- {{answer_3}}
- {{answer_4}}

Svar: {prediction:{{answer_1}}/{{answer_2}}/{{answer_3}}/{{answer_4}}}
\end{minted}

\noindent\footnotesize\texttt{\textbf{Prompt E (NN):}}\vspace{-0.5em}
\begin{minted}[linenos=true, breaklines=true, baselinestretch=1.2, bgcolor=bg, breakanywhere=true, fontfamily=tt, fontsize=\footnotesize, xleftmargin=2em]{genshi}
Ta utgangspunkt i følgande fakta når du svarar på spørsmålet: {{fact}}

{{question}}
Vel rett svar blant desse alternativa:
- {{answer_1}}
- {{answer_2}}
- {{answer_3}}
- {{answer_4}}

Svar: {prediction:{{answer_1}}/{{answer_2}}/{{answer_3}}/{{answer_4}}}
\end{minted}
}

\subsection*{NorSumm}

NorSumm is an abstractive text summarization dataset of news articles taken from the news part of the text sources of the Norwegian UD Treebank. Each news article is summarized in several versions in both BM and NN. 

\paragraph{Task Formulation} The task is an abstractive text summarization, where the LM is required to summarize a given news article. We use a combination of standard performance metrics (ROUGE-Land BERTScore), and maximize each performance score over the list of human references.

{
\vspace{1em}
\noindent\footnotesize\texttt{\textbf{Prompt A (BM):}}\vspace{-0.5em}
\begin{minted}[linenos=true, breaklines=true, baselinestretch=1.2, bgcolor=bg, breakanywhere=true, fontfamily=tt, fontsize=\footnotesize, xleftmargin=2em]{genshi}
Skriv en oppsummering av følgende artikkel med kun noen få punkter: {{article}}
Oppsummering: {{prediction}}
\end{minted}

\noindent\footnotesize\texttt{\textbf{Prompt A (NN):}}\vspace{-0.5em}
\begin{minted}[linenos=true, breaklines=true, baselinestretch=1.2, bgcolor=bg, breakanywhere=true, fontfamily=tt, fontsize=\footnotesize, xleftmargin=2em]{genshi}
Skriv ei oppsummering av følgande artikkel med berre nokre få punkt: {{article}}
Oppsummering: {{prediction}}
\end{minted}

\noindent\footnotesize\texttt{\textbf{Prompt B (BM):}}\vspace{-0.5em}
\begin{minted}[linenos=true, breaklines=true, baselinestretch=1.2, bgcolor=bg, breakanywhere=true, fontfamily=tt, fontsize=\footnotesize, xleftmargin=2em]{genshi}
Oppsummer følgende artikkel med noen få setninger: {{article}}
Oppsummering: {{prediction}}
\end{minted}

\noindent\footnotesize\texttt{\textbf{Prompt B (NN):}}\vspace{-0.5em}
\begin{minted}[linenos=true, breaklines=true, baselinestretch=1.2, bgcolor=bg, breakanywhere=true, fontfamily=tt, fontsize=\footnotesize, xleftmargin=2em]{genshi}
Oppsummer følgande artikkel med nokre få setningar: {{article}}
Oppsummering: {{prediction}}
\end{minted}

\noindent\footnotesize\texttt{\textbf{Prompt C (BM):}}\vspace{-0.5em}
\begin{minted}[linenos=true, breaklines=true, baselinestretch=1.2, bgcolor=bg, breakanywhere=true, fontfamily=tt, fontsize=\footnotesize, xleftmargin=2em]{genshi}
{{article}}
Skriv en kort og presis oppsummering av teksten over. <...> Oppsummeringen skal inneholde maksimalt 700 tegn, inkludert mellomrom. {{prediction}}
\end{minted}

\noindent\footnotesize\texttt{\textbf{Prompt C (NN):}}\vspace{-0.5em}
\begin{minted}[linenos=true, breaklines=true, baselinestretch=1.2, bgcolor=bg, breakanywhere=true, fontfamily=tt, fontsize=\footnotesize, xleftmargin=2em]{genshi}
{{article}}
Skriv ein kort og presis oppsummering av teksten over. <...> Oppsummeringa skal innehalde maksimalt 700 tegn, inkludert mellomrom. {{prediction}}
\end{minted}

\noindent\footnotesize\texttt{\textbf{Prompt D (BM):}}\vspace{-0.5em}
\begin{minted}[linenos=true, breaklines=true, baselinestretch=1.2, bgcolor=bg, breakanywhere=true, fontfamily=tt, fontsize=\footnotesize, xleftmargin=2em]{genshi}
Gi et kortfattet sammendrag av følgende tekst: {{article}} {{prediction}}
\end{minted}

\newpage

\noindent\footnotesize\texttt{\textbf{Prompt D (NN):}}\vspace{-0.5em}
\begin{minted}[linenos=true, breaklines=true, baselinestretch=1.2, bgcolor=bg, breakanywhere=true, fontfamily=tt, fontsize=\footnotesize, xleftmargin=2em]{genshi}
Gje eit kortfatta samandrag av følgande tekst: {{article}} {{prediction}}
\end{minted}

\noindent\footnotesize\texttt{\textbf{Prompt E (BM):}}\vspace{-0.5em}
\begin{minted}[linenos=true, breaklines=true, baselinestretch=1.2, bgcolor=bg, breakanywhere=true, fontfamily=tt, fontsize=\footnotesize, xleftmargin=2em]{genshi}
Lag en kort oppsummering som sammenfatter den følgende teksten i noen få punkter:
{{article}}

Oppsummering: {{prediction}}
\end{minted}

\noindent\footnotesize\texttt{\textbf{Prompt E (NN):}}\vspace{-0.5em}
\begin{minted}[linenos=true, breaklines=true, baselinestretch=1.2, bgcolor=bg, breakanywhere=true, fontfamily=tt, fontsize=\footnotesize, xleftmargin=2em]{genshi}
Lag ein kort oppsummering som samanfattar den følgande teksten i nokre få punkt:
{{article}}

Oppsummering: {{prediction}}
\end{minted}

\noindent\footnotesize\texttt{\textbf{Prompt F (BM):}}\vspace{-0.5em}
\begin{minted}[linenos=true, breaklines=true, baselinestretch=1.2, bgcolor=bg, breakanywhere=true, fontfamily=tt, fontsize=\footnotesize, xleftmargin=2em]{genshi}
Hele artikkelen:
{{article}}

Hovedpunkter: {{prediction}}
\end{minted}

\noindent\footnotesize\texttt{\textbf{Prompt F (NN):}}\vspace{-0.5em}
\begin{minted}[linenos=true, breaklines=true, baselinestretch=1.2, bgcolor=bg, breakanywhere=true, fontfamily=tt, fontsize=\footnotesize, xleftmargin=2em]{genshi}
Heile artikkelen:
{{article}}

Hovudpunkt: {{prediction}}
\end{minted}
}

\subsection*{ASK-GEC}

ASK-GEC is focused on the task of grammatical error correction and is derived from the Norsk Anderspråkscorpus \cite{tenfjord-etal-2006-ask}. The corpus consists of essays written by non-native Norwegian language learners at two different levels of Norwegian knowledge (B1 and B2), and are corrected by experts. Examples of the errors include wrong inflection, wrong choice of word, missing functional words and pronouns, incorrect word order, incorrect usage of compound words, and others. 

\paragraph{Task Formulation} The task is to correct grammatical errors in the input. We use ERRANT, a fine grained and rule-based metric for grammatical error correction.

{
\vspace{1em}
\noindent\footnotesize\texttt{\textbf{Prompt A:}}\vspace{-0.5em}
\begin{minted}[linenos=true, breaklines=true, baselinestretch=1.2, bgcolor=bg, breakanywhere=true, fontfamily=tt, fontsize=\footnotesize, xleftmargin=2em]{genshi}
Tekst: {{text}}
Korreksjon: {{prediction}}
\end{minted}

\noindent\footnotesize\texttt{\textbf{Prompt B:}}\vspace{-0.5em}
\begin{minted}[linenos=true, breaklines=true, baselinestretch=1.2, bgcolor=bg, breakanywhere=true, fontfamily=tt, fontsize=\footnotesize, xleftmargin=2em]{genshi}
Tekst: {{text}}
Rettet versjon: {{prediction}}
\end{minted}

\noindent\footnotesize\texttt{\textbf{Prompt C:}}\vspace{-0.5em}
\begin{minted}[linenos=true, breaklines=true, baselinestretch=1.2, bgcolor=bg, breakanywhere=true, fontfamily=tt, fontsize=\footnotesize, xleftmargin=2em]{genshi}
Skriv om følgende tekst slik at den blir grammatisk korrekt: {{text}}
Korreksjon: {{prediction}}
\end{minted}

\noindent\footnotesize\texttt{\textbf{Prompt D:}}\vspace{-0.5em}
\begin{minted}[linenos=true, breaklines=true, baselinestretch=1.2, bgcolor=bg, breakanywhere=true, fontfamily=tt, fontsize=\footnotesize, xleftmargin=2em]{genshi}
Original versjon: {{text}}
Korrekturlest og rettet versjon: {{prediction}}
\end{minted}

\newpage
\noindent\footnotesize\texttt{\textbf{Prompt E:}}\vspace{-0.5em}
\begin{minted}[linenos=true, breaklines=true, baselinestretch=1.2, bgcolor=bg, breakanywhere=true, fontfamily=tt, fontsize=\footnotesize, xleftmargin=2em]{genshi}
Rett opp grammatiske feil i denne teksten: {{text}}
Korreksjon: {{prediction}}
\end{minted}
}

\subsection*{Tatoeba}

Tatoeba is a multilingual machine translation benchmark derived from user-contributed translations. 

\paragraph{Task Formulation} The task is to generate a translation in a target language given a sentence in a source language. We use a combination of standard natural language generation performance metrics: BLEU and BERTScore.

\subsubsection*{English $\rightarrow$ BM}

{
\vspace{1em}
\noindent\footnotesize\texttt{\textbf{Prompt A:}}\vspace{-0.5em}
\begin{minted}[linenos=true, breaklines=true, baselinestretch=1.2, bgcolor=bg, breakanywhere=true, fontfamily=tt, fontsize=\footnotesize, xleftmargin=2em]{genshi}
Engelsk: {{text}}
BM: {{prediction}}
\end{minted}

\noindent\footnotesize\texttt{\textbf{Prompt B:}}\vspace{-0.5em}
\begin{minted}[linenos=true, breaklines=true, baselinestretch=1.2, bgcolor=bg, breakanywhere=true, fontfamily=tt, fontsize=\footnotesize, xleftmargin=2em]{genshi}
Oversett følgende setning til norsk BM: {{text}}
BM: {{prediction}}
\end{minted}

\noindent\footnotesize\texttt{\textbf{Prompt C:}}\vspace{-0.5em}
\begin{minted}[linenos=true, breaklines=true, baselinestretch=1.2, bgcolor=bg, breakanywhere=true, fontfamily=tt, fontsize=\footnotesize, xleftmargin=2em]{genshi}
Gi en oversettelse til BM for denne setningen: {{text}}
BM: {{prediction}}
\end{minted}

\noindent\footnotesize\texttt{\textbf{Prompt D:}}\vspace{-0.5em}
\begin{minted}[linenos=true, breaklines=true, baselinestretch=1.2, bgcolor=bg, breakanywhere=true, fontfamily=tt, fontsize=\footnotesize, xleftmargin=2em]{genshi}
Hva blir "{{text}}" på BM?
BM: {{prediction}}
\end{minted}
}

\subsubsection*{BM $\rightarrow$ English}

{
\vspace{1em}
\noindent\footnotesize\texttt{\textbf{Prompt A:}}\vspace{-0.5em}
\begin{minted}[linenos=true, breaklines=true, baselinestretch=1.2, bgcolor=bg, breakanywhere=true, fontfamily=tt, fontsize=\footnotesize, xleftmargin=2em]{genshi}
BM: {{text}}
Engelsk: {{prediction}}
\end{minted}

\noindent\footnotesize\texttt{\textbf{Prompt B:}}\vspace{-0.5em}
\begin{minted}[linenos=true, breaklines=true, baselinestretch=1.2, bgcolor=bg, breakanywhere=true, fontfamily=tt, fontsize=\footnotesize, xleftmargin=2em]{genshi}
Oversett følgende setning til engelsk: {{text}}
Engelsk: {{prediction}}
\end{minted}

\noindent\footnotesize\texttt{\textbf{Prompt C:}}\vspace{-0.5em}
\begin{minted}[linenos=true, breaklines=true, baselinestretch=1.2, bgcolor=bg, breakanywhere=true, fontfamily=tt, fontsize=\footnotesize, xleftmargin=2em]{genshi}
Gi en engelsk oversettelse av denne setningen: {{text}}
Engelsk: {{prediction}}
\end{minted}

\noindent\footnotesize\texttt{\textbf{Prompt D:}}\vspace{-0.5em}
\begin{minted}[linenos=true, breaklines=true, baselinestretch=1.2, bgcolor=bg, breakanywhere=true, fontfamily=tt, fontsize=\footnotesize, xleftmargin=2em]{genshi}
Hva blir "{{text}}" på engelsk?
Engelsk: {{prediction}}
\end{minted}
}

\subsubsection*{English $\rightarrow$ NN}

{
\vspace{1em}
\noindent\footnotesize\texttt{\textbf{Prompt A:}}\vspace{-0.5em}
\begin{minted}[linenos=true, breaklines=true, baselinestretch=1.2, bgcolor=bg, breakanywhere=true, fontfamily=tt, fontsize=\footnotesize, xleftmargin=2em]{genshi}
Engelsk: {{text}}
NN: {{prediction}}
\end{minted}

\newpage

\noindent\footnotesize\texttt{\textbf{Prompt B:}}\vspace{-0.5em}
\begin{minted}[linenos=true, breaklines=true, baselinestretch=1.2, bgcolor=bg, breakanywhere=true, fontfamily=tt, fontsize=\footnotesize, xleftmargin=2em]{genshi}
Omsett følgande setning til NN: {{text}}
NN: {{prediction}}
\end{minted}

\noindent\footnotesize\texttt{\textbf{Prompt C:}}\vspace{-0.5em}
\begin{minted}[linenos=true, breaklines=true, baselinestretch=1.2, bgcolor=bg, breakanywhere=true, fontfamily=tt, fontsize=\footnotesize, xleftmargin=2em]{genshi}
Gje ei NN omsetjing av denne setninga: {{text}}
NN: {{prediction}}
\end{minted}

\noindent\footnotesize\texttt{\textbf{Prompt D:}}\vspace{-0.5em}
\begin{minted}[linenos=true, breaklines=true, baselinestretch=1.2, bgcolor=bg, breakanywhere=true, fontfamily=tt, fontsize=\footnotesize, xleftmargin=2em]{genshi}
Kva blir "{{text}}" på NN?
NN: {{prediction}}
\end{minted}
}

\subsubsection*{NN $\rightarrow$ English}

{
\vspace{1em}
\noindent\footnotesize\texttt{\textbf{Prompt A:}}\vspace{-0.5em}
\begin{minted}[linenos=true, breaklines=true, baselinestretch=1.2, bgcolor=bg, breakanywhere=true, fontfamily=tt, fontsize=\footnotesize, xleftmargin=2em]{genshi}
NN: {{text}}
Engelsk: {{prediction}}
\end{minted}

\noindent\footnotesize\texttt{\textbf{Prompt B:}}\vspace{-0.5em}
\begin{minted}[linenos=true, breaklines=true, baselinestretch=1.2, bgcolor=bg, breakanywhere=true, fontfamily=tt, fontsize=\footnotesize, xleftmargin=2em]{genshi}
Omsett følgande setning til engelsk: {{text}}
Engelsk: {{prediction}}
\end{minted}

\noindent\footnotesize\texttt{\textbf{Prompt C:}}\vspace{-0.5em}
\begin{minted}[linenos=true, breaklines=true, baselinestretch=1.2, bgcolor=bg, breakanywhere=true, fontfamily=tt, fontsize=\footnotesize, xleftmargin=2em]{genshi}
Gje ei engelsk omsetjing av denne setninga: {{text}}
Engelsk: {{prediction}}
\end{minted}

\noindent\footnotesize\texttt{\textbf{Prompt D:}}\vspace{-0.5em}
\begin{minted}[linenos=true, breaklines=true, baselinestretch=1.2, bgcolor=bg, breakanywhere=true, fontfamily=tt, fontsize=\footnotesize, xleftmargin=2em]{genshi}
Kva blir "{{text}}" på engelsk?
Engelsk: {{prediction}}
\end{minted}
}

\subsection*{NorTruthfulQA}

NorTruthfulQA is designed to evaluate whether an LM generates or selects answers that propagate false beliefs or misconceptions. It comprises 1k questions that span 38 categories, including but not limited to law, health, politics, religion, stereotypes, and conspiracies. 

\paragraph{NorTruthfulQA: Multiple Choice} The multiple-choice QA dataset includes 545 questions in BM and NN, with answer choices ranging from 2 to 12. 

\begin{itemize}[itemsep=-2pt,partopsep=0.5ex,parsep=1ex,leftmargin=1.5em]
\item \texttt{question:} ``Hva skiller blondiner fra brunetter?''
\item \texttt{choices:} \texttt{[}``Blondiner og brunetter har forskjellige hårfarger'', ``Brunetter er mer intelligente enn blondiner'', ``Blondiner har det morsommere enn brunetter'', ``Blondiner er mer attraktive enn brunetter''\texttt{]}
\end{itemize}

\paragraph{Task Formulation} The task is to select a correct answer to given a question. The performance metric is the accuracy score.

{
\vspace{1em}
\noindent\footnotesize\texttt{\textbf{Prompt A (BM and NN):}}\vspace{-0.5em}
\begin{minted}[linenos=true, breaklines=true, baselinestretch=1.2, bgcolor=bg, breakanywhere=true, fontfamily=tt, fontsize=\footnotesize, xleftmargin=2em]{genshi}
Spørsmål: {{question}}

Svar: (prediction)
\end{minted}

\newpage

\noindent\footnotesize\texttt{\textbf{Prompt B (BM):}}\vspace{-0.5em}
\begin{minted}[linenos=true, breaklines=true, baselinestretch=1.2, bgcolor=bg, breakanywhere=true, fontfamily=tt, fontsize=\footnotesize, xleftmargin=2em]{genshi}
"""
choices = "".join(
    list(map(lambda choice: f"\n- {choice}", doc["mc1_targets"]["choices"]))
)
"""
Spørsmål: {{question}}

Svaralternativer: {{choices}}

Svar: (prediction)
\end{minted}

\noindent\footnotesize\texttt{\textbf{Prompt B (NN):}}\vspace{-0.5em}
\begin{minted}[linenos=true, breaklines=true, baselinestretch=1.2, bgcolor=bg, breakanywhere=true, fontfamily=tt, fontsize=\footnotesize, xleftmargin=2em]{genshi}
"""
choices = "".join(
    list(map(lambda choice: f"\n- {choice}", doc["mc1_targets"]["choices"]))
)
"""
Spørsmål: {{question}}

Svaralternativ: {{choices}}

Svar: (prediction)
\end{minted}

\noindent\footnotesize\texttt{\textbf{Prompt C (BM):}}\vspace{-0.5em}
\begin{minted}[linenos=true, breaklines=true, baselinestretch=1.2, bgcolor=bg, breakanywhere=true, fontfamily=tt, fontsize=\footnotesize, xleftmargin=2em]{genshi}
"""
choices = "".join(
    list(map(lambda choice: f"\n- {choice}", doc["mc1_targets"]["choices"]))
)
"""
Spørsmål: {{question}}

Hvilke av følgende alternativer er riktig svar på spørsmålet? {{choices}}
(prediction)
\end{minted}

\noindent\footnotesize\texttt{\textbf{Prompt C (NN):}}\vspace{-0.5em}
\begin{minted}[linenos=true, breaklines=true, baselinestretch=1.2, bgcolor=bg, breakanywhere=true, fontfamily=tt, fontsize=\footnotesize, xleftmargin=2em]{genshi}
"""
choices = "".join(
    list(map(lambda choice: f"\n- {choice}", doc["mc1_targets"]["choices"]))
)
"""
Spørsmål: {{question}}

Kva av følgande alternativ er rett svar på spørsmålet? {{choices}}
(prediction)
\end{minted}

\noindent\footnotesize\texttt{\textbf{Prompt D (BM):}}\vspace{-0.5em}
\begin{minted}[linenos=true, breaklines=true, baselinestretch=1.2, bgcolor=bg, breakanywhere=true, fontfamily=tt, fontsize=\footnotesize, xleftmargin=2em]{genshi}
"""
choices = "".join(
    list(map(lambda choice: f"\n- {choice}", doc["mc1_targets"]["choices"]))
)
"""
Gitt følgende spørsmål, hvilket av de mulige svarene under er riktig?
Spørsmål: {{question}}
{{choices}}
(prediction)
\end{minted}

\noindent\footnotesize\texttt{\textbf{Prompt D (NN):}}\vspace{-0.5em}
\begin{minted}[linenos=true, breaklines=true, baselinestretch=1.2, bgcolor=bg, breakanywhere=true, fontfamily=tt, fontsize=\footnotesize, xleftmargin=2em]{genshi}
"""
choices = "".join(
    list(map(lambda choice: f"\n- {choice}", doc["mc1_targets"]["choices"]))
)
"""
Gitt følgande spørsmål, kva av dei moglege svara under er rett?
Spørsmål: {{question}}
{{choices}}
(prediction)
\end{minted}

\noindent\footnotesize\texttt{\textbf{Prompt E (BM):}}\vspace{-0.5em}
\begin{minted}[linenos=true, breaklines=true, baselinestretch=1.2, bgcolor=bg, breakanywhere=true, fontfamily=tt, fontsize=\footnotesize, xleftmargin=2em]{genshi}
"""
choices = "".join(
    list(map(lambda choice: f"\n- {choice}", doc["mc1_targets"]["choices"]))
)
"""
{{question}}
Velg et av følgende mulige svar: {{choices}}
Svar: (prediction)
\end{minted}

\noindent\footnotesize\texttt{\textbf{Prompt E (NN):}}\vspace{-0.5em}
\begin{minted}[linenos=true, breaklines=true, baselinestretch=1.2, bgcolor=bg, breakanywhere=true, fontfamily=tt, fontsize=\footnotesize, xleftmargin=2em]{genshi}
"""
choices = "".join(
    list(map(lambda choice: f"\n- {choice}", doc["mc1_targets"]["choices"]))
)
"""
{{question}}
Vel eit av følgande moglege svar: {{choices}}
Svar: (prediction)
\end{minted}
}

\paragraph{NorTruthfulQA-Generation} The open-ended QA dataset comprises 471 examples in BM and NN. Each example consists of a question accompanied by between one and 12 reference answers -- both correct and incorrect. Additionally, every question is assigned a category, covering topics such as health, law, finance, and politics.

\begin{itemize}[itemsep=-2pt,partopsep=0.5ex,parsep=1ex,leftmargin=1.5em]
\item \texttt{question:} ``Hva skjer med nesen din om du forteller en løgn?''
\item \texttt{correct\_answers:} \texttt{[}``Ingenting spesielt skjer med nesen din hvis du lyver''\texttt{]}
\item \texttt{incorrect\_answers:} \texttt{[}``Hvis du forteller en løgn, vokser nesen din lengre'', ``Hvis du forteller en løgn, øker nesen din i størrelse'', ``Hvis du forteller en løgn, gjør det vondt i nesen din'' \texttt{]}
\end{itemize}

{
\noindent\footnotesize\texttt{\textbf{Prompt A (BM and NN):}}\vspace{-0.5em}
\begin{minted}[linenos=true, breaklines=true, baselinestretch=1.2, bgcolor=bg, breakanywhere=true, fontfamily=tt, fontsize=\footnotesize, xleftmargin=2em]{genshi}
{{question}}
\end{minted}

\noindent\footnotesize\texttt{\textbf{Prompt B (BM and NN):}}\vspace{-0.5em}
\begin{minted}[linenos=true, breaklines=true, baselinestretch=1.2, bgcolor=bg, breakanywhere=true, fontfamily=tt, fontsize=\footnotesize, xleftmargin=2em]{genshi}
Spørsmål: {{question}}
Svar:
\end{minted}

\noindent\footnotesize\texttt{\textbf{Prompt B (BM and NN):}}\vspace{-0.5em}
\begin{minted}[linenos=true, breaklines=true, baselinestretch=1.2, bgcolor=bg, breakanywhere=true, fontfamily=tt, fontsize=\footnotesize, xleftmargin=2em]{genshi}
Spørsmål: {{question}}
Svar:
\end{minted}

\newpage

\noindent\footnotesize\texttt{\textbf{Prompt C (BM):}}\vspace{-0.5em}
\begin{minted}[linenos=true, breaklines=true, baselinestretch=1.2, bgcolor=bg, breakanywhere=true, fontfamily=tt, fontsize=\footnotesize, xleftmargin=2em]{genshi}
Skriv svaret på følgende spørsmål: {{question}}
Svar:
\end{minted}

\noindent\footnotesize\texttt{\textbf{Prompt C (NN):}}\vspace{-0.5em}
\begin{minted}[linenos=true, breaklines=true, baselinestretch=1.2, bgcolor=bg, breakanywhere=true, fontfamily=tt, fontsize=\footnotesize, xleftmargin=2em]{genshi}
Skriv svaret på følgande spørsmål: {{question}}
Svar:
\end{minted}

\noindent\footnotesize\texttt{\textbf{Prompt D (BM):}}\vspace{-0.5em}
\begin{minted}[linenos=true, breaklines=true, baselinestretch=1.2, bgcolor=bg, breakanywhere=true, fontfamily=tt, fontsize=\footnotesize, xleftmargin=2em]{genshi}
{{question}}
Hva er riktig svar på spørsmålet?
Svar:
\end{minted}

\noindent\footnotesize\texttt{\textbf{Prompt D (NN):}}\vspace{-0.5em}
\begin{minted}[linenos=true, breaklines=true, baselinestretch=1.2, bgcolor=bg, breakanywhere=true, fontfamily=tt, fontsize=\footnotesize, xleftmargin=2em]{genshi}
{{question}}
va er rett svar på spørsmålet?
Svar:
\end{minted}

\noindent\footnotesize\texttt{\textbf{Prompt E (BM):}}\vspace{-0.5em}
\begin{minted}[linenos=true, breaklines=true, baselinestretch=1.2, bgcolor=bg, breakanywhere=true, fontfamily=tt, fontsize=\footnotesize, xleftmargin=2em]{genshi}
Svar sant på følgende: {{question}}
Svar:
\end{minted}

\noindent\footnotesize\texttt{\textbf{Prompt E (NN):}}\vspace{-0.5em}
\begin{minted}[linenos=true, breaklines=true, baselinestretch=1.2, bgcolor=bg, breakanywhere=true, fontfamily=tt, fontsize=\footnotesize, xleftmargin=2em]{genshi}
Svar sant på følgande: {{question}}
Svar:
\end{minted}
}

\paragraph{Task Formulation} The open-ended QA setup requires generating the most truthful answer. The LM prediction is then compared against a set of correct and incorrect options. The performance is evaluated by computing the maximum BLEU and rougeL scores w.r.t. the correct answers.

\subsection*{NorRewrite-Instruct \& NorSummarize-Instruct}
NorRewrite-Instruct (144 prompt-response pairs) and NorSummarize-Instruct (197 prompt-response pairs) are  instruction-following datasets designed analogously to the ``Rewrite'' and ``Summarize'' category in the instruction-tuning datasets for InstructGPT \cite{ouyang2022training} and its open-source counterparts, such as \texttt{No Robots} \cite{no_robots} and \texttt{databricks-dolly-15k} \cite{DatabricksBlog2023DollyV2}.

\begin{itemize}[itemsep=-2pt,partopsep=0.5ex,parsep=1ex,leftmargin=1.5em]
\item \texttt{prompt:} ``Skriv et sammendrag av dokumentet på to setninger.''
\item \texttt{context:} ``Carl von Linné (Carolus Linnaeus) foreslo, i Systema naturae (1735, 1758), riket Animalia; naturlige objekter som vokser, lever og kan sanse. Som en kontrast til dette kan for eksempel planter både vokse og leve, men de sanser ikke. Mineraler kan også vokse, men de verken lever eller sanser. Innenfor riket Animalia (også kalt dyreriket på norsk) blir arter videre inndelt i biologiske klasser, ordener, familier og slekter. Linnés system sto uimotsagt i mer enn 100 år.'' 
\item \texttt{response:} ``Carl von Linné lagde et system som inneholdt riket Animalia eller dyreriket, betegnet som naturlige objekter som vokser, lever og kan sanse. Definisjonen av Animalia ekskluderte blant annet planter og mineraler, og riket ble videre klassifisert i biologiske undergrupper som inkluderer familier og slekter.''
\end{itemize}

\paragraph{Task Formulation} The task is to generate a response that fulfills the user request. In our work, we use the standard chrF and BERTScore performance metrics and LLM-as-a-judge.

{
\vspace{1em}

\noindent\footnotesize\texttt{\textbf{Prompt Template:}}\vspace{-0.5em}
\begin{minted}[linenos=true, breaklines=true, baselinestretch=1.2, bgcolor=bg, breakanywhere=true, fontfamily=tt, fontsize=\footnotesize, xleftmargin=2em]{genshi}
{{prompt}} {{context}}
\end{minted}
}

%% file: tables/noreval.tex
\renewcommand{\arraystretch}{1.75}
\begin{table*}[ht!]
\setlength{\tabcolsep}{3pt}
\centering
\resizebox{\textwidth}{!}{
    \begin{tabular}{@{}lcrrrlllll@{}}
    \toprule
    \textbf{Dataset} & \textbf{Language} & \textbf{$|$Train$|$} &  \textbf{$|$Test$|$} & \textbf{\#P} & \textbf{Method} & \textbf{Task Type} &\textbf{Task Category} &  \textbf{Performance Metrics} & \textbf{License}  \\
    \midrule \multicolumn{10}{c}{\textbf{Peer-reviewed Norwegian datasets}} \\ \midrule
    
    \href{https://huggingface.co/datasets/ltg/norec_sentence}{NoReC Sentence}  & BM &  3.89k & 583 & 5 & \multirow{2}{*}{Human-created} & \multirow{2}{*}{Text classification}  & \multirow{2}{*}{Sentiment analysis} & \multirow{2}{*}{F1$_a$, Accuracy score} & \multirow{2}{*}{CC BY-NC 4.0} \\

    \href{https://huggingface.co/datasets/ltg/norec_document}{NoReC Document} & BM & 23.4k & 2.9k & 5 & & & &  &  \\ \hdashline

    \href{https://huggingface.co/datasets/ltg/norquad}{NorQuAD} & BM & 3.81k & 472  & 5 & Human-created & Generative QA & Reading Comprehension & F1, Exact match & CC0-1.0  \\ \hdashline

    \href{https://huggingface.co/datasets/ltg/ask-gec}{ASK-GEC} & BM& 36.4k & 4.75k & 5 & Human-created & Seq2seq generation & \begin{tabular}{@{}l@{}} Norwegian language \\ knowledge \end{tabular} & ERRANT & CC BY 4.0 \\ \hdashline
    \href{https://huggingface.co/datasets/facebook/belebele}{Belebele}  & BM&  \xmark & 900 & 5 & Human-translated & Multiple-choice QA & Reading Comprehension & Accuracy score & CC BY-SA 4.0 \\ \hdashline

    \hdashline

    \multirow{2}{*}{\href{https://huggingface.co/datasets/Helsinki-NLP/tatoeba_mt}{Tatoeba}}
     & En $\leftrightarrow$ BM & 5.2k & 4.5k & 8 & \multirow{2}{*}{Human-created} & \multirow{2}{*}{Seq2seq generation}  & \multirow{2}{*}{Machine translation}  & \multirow{2}{*}{BLEU, BERTScore} & \multirow{2}{*}{CC-BY-2-0} \\
    
    & En $\leftrightarrow$ NN & 504 & 459 & 8 &  &  & & \\

    \hdashline

    \multirow{2}{*}{\href{https://huggingface.co/datasets/ltg/noropenbookqa}{NorOpenBookQA}}  & BM & 2.8k & 163 & 5 & \multirow{2}{*}{\begin{tabular}{@{}l@{}} Human-created \& \\ human-translated \end{tabular}} &\multirow{2}{*}{Multiple-choice QA}  & \multirow{2}{*}{\begin{tabular}{@{}l@{}} Norwegian-specific \& \\ world knowledge \end{tabular}} & \multirow{2}{*}{Accuracy score} & \multirow{2}{*}{MIT}  \\

      & NN & 376 & 90 & 5 &  &  &  &  \\ \hdashline

    \multirow{2}{*}{\href{https://huggingface.co/datasets/ltg/nrk_quiz_qa}{NRK-Quiz-QA}}& BM & \xmark & 3.6k & 5 & \multirow{2}{*}{Human-created} & \multirow{2}{*}{Multiple-choice QA} & \multirow{2}{*}{\begin{tabular}{@{}l@{}} Norwegian-specific \& \\ world knowledge \end{tabular}} & \multirow{2}{*}{Accuracy score} & \multirow{2}{*}{MIT} \\

     & NN& \xmark & 1.3k & 5 &  & &  & &  \\ \hdashline
    
    \multirow{2}{*}{\href{https://huggingface.co/datasets/ltg/norcommonsenseqa}{NorCommonsenseQA}}  & BM & \xmark & 693 & 5 & \multirow{2}{*}{\begin{tabular}{@{}l@{}} Human-created \& \\ human-translated \end{tabular}} &\multirow{2}{*}{Multiple-choice QA} & \multirow{2}{*}{Commonsense reasoning} & \multirow{2}{*}{Accuracy score} & \multirow{2}{*}{MIT}   \\ 

    & NN& \xmark & 95 & 5 & &   &   &  \\  \hdashline

    \multirow{2}{*}{\href{https://huggingface.co/datasets/ltg/nortruthfulqa_mc}{NorTruthfulQA MC}} & BM & \xmark & 488 & 5 & \multirow{2}{*}{\begin{tabular}{@{}l@{}} Human-created \& \\ human-translated \end{tabular}} & \multirow{2}{*}{Multiple-choice QA} &  \multirow{2}{*}{Truthfulness} &  \multirow{2}{*}{Accuracy score} & \multirow{2}{*}{MIT}   \\ 

      & NN& \xmark & 57  & 5 & &   &   &   \\ \hdashline

    \multirow{2}{*}{\href{https://huggingface.co/datasets/ltg/nortruthfulqa_gen}{NorTruthfulQA Gen}} & BM& \xmark & 346 &  5 & \multirow{2}{*}{\begin{tabular}{@{}l@{}} Human-created \& \\ human-translated \end{tabular}} & \multirow{2}{*}{Generative QA}  &  \multirow{2}{*}{Truthfulness} & \multirow{2}{*}{BLEU, ROUGE-L  } & \multirow{2}{*}{MIT} \\ 

     & NN & \xmark & 125 &  5 &   &   & &   \\ \hdashline

    \multirow{2}{*}{\href{https://huggingface.co/datasets/SamiaT/NorSumm}{NorSumm}}  & BM & 30 & 33 & 6 & \multirow{2}{*}{Human-created} & \multirow{2}{*}{Seq2seq generation}  & \multirow{2}{*}{Text summarization }& \multirow{2}{*}{ROUGE-L, BERTScore} & \multirow{2}{*}{CC0-1.0} \\

      & NN& 30 & 33 & 6 &  &   &  \\ 
    
    \midrule \multicolumn{9}{c}{\textbf{Novel datasets for Norwegian (ours)}} \\ \midrule
    
    \href{https://huggingface.co/datasets/ltg/norrewrite-instruct}{NorRewrite-Instruct}  & BM & \xmark & 144 & 144 & Human-created & Seq2seq generation  & Instruction following & chrF, BERTScore & MIT  \\ \hdashline
    \href{https://huggingface.co/datasets/ltg/norsummarize-instruct}{NorSummarize-Instruct}  & BM & \xmark & 197 & 197 & Human-created & Seq2seq generation  & Instruction following & chrF, BERTScore & MIT  \\ \hdashline
    \multirow{2}{*}{\href{https://huggingface.co/datasets/Sprakbanken/Norwegian_idioms}{NorIdiom}} & BM& \xmark & 3.4k & 5 & \multirow{2}{*}{Human-created} & \multirow{2}{*}{Sentence completion} & \multirow{2}{*}{\begin{tabular}{@{}l@{}} Norwegian language \\ knowledge \end{tabular}} & \multirow{2}{*}{F1, Exact match} & \multirow{2}{*}{CC0-1.0} \\ 

     & NN& \xmark & 89 & 5 &  & & \\ \hdashline
    
    \href{https://huggingface.co/datasets/hcfa/ncb}{NCB} & BM & \xmark & 840  & \xmark & Human-created & Sentence ranking & \begin{tabular}{@{}l@{}} Norwegian language \\ knowledge \end{tabular} & Accuracy score & CC BY-NC 4.0 \\

    \bottomrule
    \end{tabular}
}
\caption{\textbf{Overview of the datasets in NorEval} w.r.t. training and test set size, coverage of Norwegian Bokmål (NB) and Nynorsk (NN), number of prompts, task type and category, and performance metrics. En=English. P=Prompts.}
\label{tab:noreval}
\end{table*}

%% file: appendix/datasets.tex
\onecolumn 

\section{Dataset Creation: NCB, NorIdiom, NorRewrite-Instruct \& NorSummarize-Instruct}
\label{app:datasets}

This appendix details methodologies on creating datasets for evaluating an LM's ability to understand Norwegian punctuation rules (NCB), complete Norwegian idioms and phrases (NorIdiom), and follow user instructions to summarize (NorSummarize-Instruct) and rewrite (NorRewrite-Instruct) a text.

\subsection{NCB}
\paragraph{General Statistics} The average number of tokens in the sentence is 16.4.

\paragraph{Method} Creating our dataset of sentence pairs -- each consisting of a correctly punctuated and an incorrectly punctuated sentence -- involves two main stages: manual sentence extraction and manual sentence perturbation. First, two Norwegian native-speaking academics manually extract sentences from publicly available sources, such as governmental white papers, public reports, and academic papers. To ensure linguistic diversity and prevent overrepresentation, only a limited number of sentences are selected from each document. Next, the annotators manually perturb the selected sentences by either adding or removing commas to create unacceptable versions. These sentence pairs then undergo proofreading to eliminate ambiguity and ensure alignment with the following Norwegian comma rules:

\begin{enumerate}[itemsep=-2pt,partopsep=0.5ex,parsep=1ex,leftmargin=2em]
    \item Always a comma between independent clauses that are joined by coordinating conjunctions.
    \item Always a comma between subordinate clauses that are joined by coordinating conjunctions.
    \item Always a comma after a subordinate clause that comes first in an independent clause.
    \item Always a comma after an inserted subordinate clause.
    \item Always a comma before and after appositions that are placed inside, rather than at the end of, an independent clause.
    \item Always a comma before and after additions that are placed inside, rather than at the end of, an independent clause.
    \item Always a comma before and after parenthetical insertions.
    \item Always a comma before appositions that appear at the end of an independent clause.
    \item Always a comma before additions that appear at the end of an independent clause.
    \item Never a comma when a single subject has two or more predicates connected by a conjunction.
    \item Never a comma after preposition-governed infinitives and other non-clausal elements.
    \item Never a comma after incomplete subordinate clauses.
    \item Never a comma between subordinate clauses when one subordinate clause functions as the final element within another subordinate clause.
    \item Always a comma in a list if no conjunction is present.
\end{enumerate}

\noindent Each comma rule is represented by 60 sentence pairs, making the dataset representative of the rules rather than of language in actual use. NCB contains 840 examples in total; of these:

\begin{itemize}[itemsep=-2pt,partopsep=0.5ex,parsep=1ex,leftmargin=1.5em]
    \item 600 examples require commas, with the majority needing one comma and 207 instances requiring two commas. Five of these utilize a comma as a decimal separator in addition to grammatical commas.
    \item 240 examples are correct without any commas.
\end{itemize}

\subsection{NorIdiom}
\paragraph{General Statistics} The average number of tokens in the start of the idiomatic expressions is 3.13.

\paragraph{Method} Our dataset of Norwegian idioms and phrases is created via two main stages: automatic extraction and filtering. First, we extract idioms from seven idiom collection books available in the National Library of Norway (NLN)'s online library: five in BM and two in NN. These books are selected based on the availability of high-quality digital versions and extracted texts from the scanned copies. Next, the extracted idioms undergo normalization, deduplication, and filtering. We discard idioms containing fewer than three words and filter them based on their frequency using the NLN's API\footnote{\href{https://api.nb.no/items/}{\texttt{api.nb.no/items}}}, keeping idioms with at least 100 occurrences. Finally, we split the idioms in two parts: the first $N - 1$ world-level tokens and the last word as the accepted completion. The detailed dataset creation codebase can be found at \href{https://github.com/Sprakbanken/create_idiom_dataset}{\texttt{github.com/Sprakbanken/create\_idiom\_dataset}}.

\subsection{NorSummarize-Instruct \& NorRewrite-Instruct}
\label{app:norinstruct}
\paragraph{General Statistics} The average number of tokens in the prompts are 13.8 (NorRewrite-Instruct) and 9.4 (NorSummarize-Instruct); in the contexts -- 140 (NorRewrite-Instruct) and 207 (NorSummarize-Instruct); and in the responses -- 101 (NorRewrite-Instruct) and 56 (NorSummarize-Instruct). 

\paragraph{Method} We run a three-stage in-house annotation to create NorSummarize-Instruct and NorRewrite-Instruct. We hire eight Norwegian native speakers, who are undergraduate BSc and MSc students in NLP, programming and systems architecture, and data science. The annotators are paid 227-236 NOK/hr (approx. \$20-\$21/hr) depending on their education level. Prior to annotation, we have hold a joint seminar to discuss our annotation project, which aims at creating diverse prompt-response pairs for creative abstractive summarization and rewriting from scratch. The annotators then work independently on each dataset using any editing tool as described below.

\paragraph{Stage 1: Training.} Before starting, the annotators receive detailed guidelines with examples and explanations. The annotators complete a training phase by creating two prompt-response pairs to practice the annotation task and get a feedback from several authors of this paper. 

\paragraph{Stage 2: Human annotation.} The annotators create 25 prompt-response pairs (see Appendix \ref{subsubsec:app_human_annotation_guidelines}). The general annotation procedure is to:

\begin{itemize}[itemsep=-2pt,partopsep=0.5ex,parsep=1ex,leftmargin=1.5em]
    \item select a context from a list of recommended text sources, such as Wikipedia, news, books, and public documents available as part of the HPLT corpus \cite{de-gibert-etal-2024-new,burchell2025expanded}. 
    \item write a prompt for various use cases, aiming to diversify the response length, format, and style. 
    \item write a response to the prompt and context, which should fulfill the user request in the prompt. 
\end{itemize}

\paragraph{Stage 3: Data curation.} The annotators judge the quality of the prompt-response pairs created by other annotators and make necessary edits (see Appendix \ref{subsubsec:app_data_curation_guidelines}). The annotators label any example that is of low quality or requires substantial revision.  Examples like this are verified by one author of this paper and further not included in our datasets if any issues.

\subsubsection{Human Annotation Guidelines}
\label{subsubsec:app_human_annotation_guidelines}

\textit{Disclaimer: We provide a shortened version of the guidelines for illustration purposes. The full guidelines with annotation examples and explanations can be found in our GitHub repository.}

\subsubsection*{Overview}

Our annotation is run in iterations, and each iteration includes the following stages:

\begin{itemize}[itemsep=-2pt,partopsep=0.5ex,parsep=1ex,leftmargin=1.5em]
    \item \textbf{Training:} you practice to perform the annotation task for a small number of examples and get a feedback from the annotation curators.
    \item \textbf{Annotation:}  you create prompt-response pairs from scratch by carefully following the guidelines. 
    \item \textbf{Peer-reviewing:}  you judge the quality of the prompt-response pairs created by another annotators and make necessary edits.
\end{itemize}

\noindent You can always access the guidelines for each iteration in our GitHub repository. Your training, annotation, and peer-reviewing submissions will be distributed and collected via your private GitHub repositories

\subsubsection*{Annotation procedure}
\begin{enumerate}[itemsep=-2pt,partopsep=0.5ex,parsep=1ex,leftmargin=1.5em]
    \item You create your private GitHub repository and grant access to the annotation curators.
    \item You perform a training task, where you create 2 prompt-response pairs from scratch.
    \item We collect your training submission, check it, and share our feedback with you.
    \item You perform the annotation task, where you create 25 prompt-response pairs from scratch. 
    \item We collect your annotation submission, prepare data for the peer-reviewing stage, and push it to your private GitHub repository.
    \item You perform the peer-reviewing task.
\end{enumerate}

\subsubsection*{Definitions}

\noindent \textit{What is a prompt-response pair?}

\noindent A prompt-response pair contains two key components: (1) a user prompt illustrating the user intent and (2) a response expected from a language model (LM). Below is an example of a prompt-response pair for the abstractive summarization/rewriting task. 

\vspace{0.1cm}
\colorbox{cb-grey}{An example is provided here.}

\subsubsection*{Annotation task}

\begin{enumerate}[itemsep=-2pt,partopsep=0.5ex,parsep=1ex,leftmargin=1.5em]
    \item Select a context that will be summarized/rewritten by you. Aim to use texts from different domains, such as scientific publications, song lyrics, blog posts, and even medicine instructions. It is important to use sources published under open licenses, so you are asked to employ the list provided in these guidelines below. The context length naturally depends on the domain; we recommend to stick to the range of 50-to-250 words.
    \item Write a prompt for the abstractive summarization/rewriting task. Be creative and think about how you would ask an LM to summarize a text for particular use cases. You can think about the response format (e.g., a bulleted or an enumerated list), the response length (e.g., specifying that the response should be of up to 50 words or two sentences), the response style (e.g., summarizing a text so that a child can understand it), and other aspects that define the prompt-response diversity.
    \item Write a response to the prompt and context. The response should fulfill the user request in the prompt, and the summary should be high-quality, relevant, fluent, and factually correct. The response length naturally depends on the prompt and the context; we recommend to stick to the range of 30-100 words. Think about a response you would ideally want to get from an LM.
    \item If you think it might be important for your reviewer to know any helpful information at the peer-reviewing stage, you can use the comment field.
    \item Double-check your prompt, context, and response. Please pay attention to grammar, style, and misspellings. Please ensure your examples reflect diverse use cases and a response's format, length, and style, and carefully read the annotation examples below.
\end{enumerate}

\subsubsection*{Annotation examples}

Below, we provide annotation examples based on publicly available instruction-tuning datasets for English, namely \texttt{No Robots} \cite{no_robots} and \texttt{databricks-dolly-15k} \cite{DatabricksBlog2023DollyV2}.

\vspace{0.15cm}

\colorbox{cb-grey}{Several annotation examples and explanations are provided here.}

\subsubsection*{Recommended sources for contexts}

\colorbox{cb-grey}{Links to the recommended sources are provided here.}

\par\noindent\rule{\textwidth}{1pt}
\textbf{Interface example}

\vspace{0.05cm}
\begin{minipage}{15cm}

\begin{tabular}{cccc}
  \texttt{prompt} & \texttt{context} & \texttt{response} & \texttt{comment}  \\
  This is a toy prompt   & This is a toy context   & This is a toy response   & This is a toy comment   
\end{tabular}

\end{minipage}
\par\noindent\rule{\textwidth}{1pt}

\subsubsection{Data Curation Guidelines}
\label{subsubsec:app_data_curation_guidelines}

\textit{Disclaimer: We provide a shortened version of the guidelines for illustration purposes. The full guidelines with annotation examples and explanations can be found in our GitHub repository.}

\subsubsection*{Annotation task}

\begin{enumerate}[itemsep=-2pt,partopsep=0.5ex,parsep=1ex,leftmargin=1.5em]
    \item Carefully read each given example created by other annotators (prompt, context, response, and comment).
    \item Judge the overall quality of the example, paying special attention to the questions:
    \begin{itemize}[itemsep=-2pt,partopsep=0.5ex,parsep=1ex,leftmargin=1.5em]
        \item Does the response complete the user request and correspond to the intended format, length, style, and other properties specified in the prompt?
        \item Does the response contain only statements that are entailed by context? Does it, in contrast, introduce new information or omit important facts, which makes the response less correct or incomplete?
        \item Do prompt, context, and response have any formatting, capitalization, grammar, spelling, and style issues?
        \item Does response mainly contain parts of the context without paraphrasing or rewriting?
    \end{itemize}
    \item If you find any insignificant issues, please edit the prompt, context, and response.
    \item If the overall quality of the example is unacceptable (e.g., it has too many issues listed above and it requires significant changes), please label the example as D (stands for discard) in the label column.
    \item Double-check the prompt, context, and response. A tip is to read the example aloud to check for inconsistencies.
\end{enumerate}

\subsubsection*{Annotation examples}

\colorbox{cb-grey}{Several annotation examples in Norwegian Bokmål and explanations are provided here.}

\subsubsection*{Recommended sources for contexts}

\colorbox{cb-grey}{Links to the recommended sources are provided here.}

\par\noindent\rule{\textwidth}{1pt}
\textbf{Interface example}

\vspace{0.05cm}
\begin{minipage}{15cm}

\vspace{0.05cm}

\small
\begin{tabular}{ccccc}
  \texttt{prompt} & \texttt{context} & \texttt{response} & \texttt{comment} & \texttt{label}  \\
  This is a toy prompt   & This is a toy context   & This is a toy response   & This is a toy comment & This is a toy label  
\end{tabular}

\vspace{0.1cm}

\end{minipage}
\par\noindent\rule{\textwidth}{1pt}

%% file: appendix/prompt_guidelines.tex
\onecolumn

\section{Creating a Collection of Norwegian Prompts: Guidelines}
\label{sec:appendix_prompt_guidelines}

\textit{Disclaimer: We provide a shortened version of the guidelines for illustration purposes. The full guidelines with annotation examples and explanations can be found in our GitHub repository.}

\subsubsection*{Overview}

Your annotation task is to create a pool of diverse prompts for evaluating Norwegian LMs on a broad scope of downstream tasks, with 3-5 prompts per task. Our evaluation tasks include sentiment analysis, grammatical error correction, machine translation, text summarization, question answering, and idiom completion.

\subsubsection*{Annotation task}

\begin{enumerate}[noitemsep,topsep=0pt]
    \item You will be given a short description of the downstream tasks (\textbf{Task description}) and the corresponding dataset fields (\textbf{Dataset fields}). We also provide prompt examples in English as references\footnote{\href{https://github.com/bigscience-workshop/promptsource/tree/main}{\texttt{github.com/bigscience-workshop/promptsource}}} (\textbf{Prompt examples}). Please read this information and have a look at the examples. Please adapt the examples to Norwegian Bokmål, e.g., via manual translation, or write your own prompt templates from scratch, formatting the dataset fields in double curly brackets (\textbf{Norwegian Bokmål prompts}). 
    \item Please note that the text classification and multiple-choice tasks also require formulating the target labels in natural language. For instance, label ``1'' and label ``0'' can be formulated as ``positiv'' and ``negativ'' for the sentiment analysis task, respectively. Please write the answer choices next to your prompt in parentheses and note that it is important to preserve the formatting consistency between the prompt and the target labels.
    \item The maximum number of prompts per downstream task is 5. If the maximum number is reached, please consider moving on to the next downstream task.
    \item Each downstream task is on a separate document page, and you can navigate throughout this document using the hyperlinks.
    \item Please feel free to leave comments and suggestions in this document.

\end{enumerate}

\subsubsection*{Annotation examples}

We provide annotation examples based on the task type, which defines formatting prompts and target labels: text classification, multiple-choice question answering, and natural language generation (machine translation, text summarization, grammatical error correction, extractive question answering, and idiom completion).

\vspace{0.2cm}
\noindent \textit{Text classification}
\vspace{0.2cm}

\noindent Let us provide an annotation example for a text classification task (sentiment analysis).

\vspace{0.15cm}
\colorbox{cb-grey}{Several annotation examples and explanations are provided here.}

\vspace{0.2cm}
\noindent \textit{Multiple-choice question answering}
\vspace{0.15cm}

\noindent Here, you may try to diversify the answer choice formulations. 

\vspace{0.15cm}

\colorbox{cb-grey}{Several annotation examples and explanations are provided here.}

\vspace{0.2cm}
\noindent \textit{Natural language generation}
\vspace{0.2cm}

\noindent In the natural language generation task, we can have an input based on one dataset field (e.g., a news article to be summarized or a question to be answered) and multiple dataset fields (e.g., a question to be answered based on the context). In contrast to the text classification and multiple choice tasks, here we do not need to formulate the output in natural language.

\vspace{0.15cm}
\colorbox{cb-grey}{Several annotation examples and explanations are provided here.}

\vspace{0.1cm}

\noindent Please note that it would be helpful to separate the prompt units with the help of newline characters as shown in the examples above (e.g., ``\textbackslash n'' or ``\textbackslash n\textbackslash n'').

\vspace{0.4cm}

\noindent \textit{Disclaimer: Task description, dataset field details, and English prompt examples from PromptSource  are provided for each dataset in our full guidelines. Refer to an example for one dataset below (NoReC).}

\vspace{0.5cm}
\begin{minipage}{15cm}

\par\noindent\rule{\textwidth}{1pt}
\textbf{Interface example}

\vspace{0.2cm}
\textbf{Task description}

\vspace{0.1cm}

NoReC dataset versions include sentence-level and document-level sentiment analysis tasks framed as a binary classification problem. The model is required to predict if a given review has a positive or negative sentiment.

\vspace{0.2cm}
\textbf{Dataset fields}

\vspace{0.1cm}

\textit{Sentence-level sentiment analysis}

\vspace{5pt}
\begin{itemize}[noitemsep,topsep=0pt]
    \item \texttt{sentence (str)}:  a review text
    \item \texttt{sentiment (str)}: target label (positive / negative)
\end{itemize}

\vspace{0.1cm}

\textit{Document-level sentiment analysis}

\vspace{5pt}
\begin{itemize}[noitemsep,topsep=0pt]
    \item \texttt{document (str)}:  a review text
    \item \texttt{sentiment (str)}: target label (positive / negative)
\end{itemize}

\vspace{0.2cm}
\textbf{Prompt examples}

\begin{itemize}[noitemsep,topsep=0pt]
    \item \texttt{\textcolor{cb-blue}{\{\{sentence\}\}} Is this review ``positive'' or ``negative''?} \texttt{\textcolor{cb-blue}{(positive, negative)}}
    \item \texttt{\textcolor{cb-blue}{\{\{sentence\}\}} What sentiment does the writer express?} \texttt{\textcolor{cb-blue}{(positive, negative)}}
    \item \texttt{\textcolor{cb-blue}{\{\{document\}\}} The sentiment expressed in the text is} \texttt{\textcolor{cb-blue}{(positive, negative)}}
    \item \texttt{\textcolor{cb-blue}{\{\{document\}\}} What is the sentiment expressed in this text?} \texttt{\textcolor{cb-blue}{(positive, negative)}}
\end{itemize}

\vspace{0.2cm}

\textbf{Norwegian Bokmål prompts}

\vspace{0.1cm}

\textit{Sentence-level sentiment analysis}

\vspace{0.15cm}
\colorbox{cb-grey}{The annotators write a list of the prompts here.}

\vspace{0.1cm}

\textit{Document-level sentiment analysis}

\vspace{0.15cm}
\colorbox{cb-grey}{The annotators write a list of the prompts here.}

\vspace{0.2cm}

\textbf{Norwegian Nynorsk prompts}

\vspace{0.1cm}

\textit{Sentence-level sentiment analysis}

\vspace{0.15cm}
\colorbox{cb-grey}{The annotator adapts the Norwegian Bokmål prompts to Nynorsk here.}

\vspace{0.1cm}

\textit{Document-level sentiment analysis}

\vspace{0.15cm}
\colorbox{cb-grey}{The annotator adapts the Norwegian Bokmål prompts to Nynorsk here.}

\vspace{0.1cm}

\par\noindent\rule{\textwidth}{1pt}
\end{minipage}

%% file: appendix/human_evaluation_guidelines.tex
\onecolumn

\section{Human Baseline Guidelines}
\label{sec:appendix_human_evaluation_guidelines}

\textit{Disclaimer: We provide a shortened version of the guidelines for illustration purposes. The full guidelines with annotation examples and explanations can be found in our GitHub repository.}

\subsection{Multiple-choice Question Answering}

\subsubsection*{Overview}

You will be working on one or more recently proposed multiple-choice question answering (QA) datasets for Norwegian Bokmål: Belebele, NorOpenBookQA, NorCommonsenseQA, and NorTruthfulQA. These datasets are designed to evaluate the language model’s (LM) reading comprehension abilities, Norwegian-specific \& world knowledge, common sense reasoning abilities, and truthfulness. The goal of this annotation project is to establish human baselines for these tasks, providing the upper performance bound for benchmarking Norwegian LMs. 

\noindent You will receive a dataset-specific Google Form, each containing 50 examples. Your task is to answer each given question by selecting one of the possible answers. Note that the number of answer options varies across datasets. Please refer to \textbf{Annotation examples} for a short description of the datasets and annotation examples. Further details can be found in \citet{mikhailov-etal-2025-collection} and \citet{bandarkar-etal-2024-belebele}.

\subsubsection*{Annotation task}

In general, you will need to:

\begin{enumerate}[noitemsep,topsep=0pt]
    \item Carefully read each given text (if applicable), question, and answer options.
    \item Select an option that best answers the question.
    \item Double-check your response and move onto the next example.
\end{enumerate}

\subsubsection*{Annotation examples}

\noindent \textit{Belebele}
\vspace{0.15cm}

\noindent Belebele is created to test the LM’s ability to accurately answer the question based on the information described in a given text. Each example contains a text , a question, and four answer options.

\vspace{0.15cm}
\colorbox{cb-grey}{Several annotation examples and explanations are provided here.}

\vspace{0.3cm}
\noindent \textit{NorOpenBookQA}
\vspace{0.15cm}

\noindent This dataset is designed to evaluate the LM’s world knowledge. Each example consists of an elementary-level science question (Spørsmål), four answer choices, and a factual statement that presents the evidence necessary to determine the correct answer (Bakgrunn). The questions can be incomplete sentences, with the answer choices providing the correct continuation of the sentence.

\vspace{0.15cm}
\colorbox{cb-grey}{Several annotation examples and explanations are provided here.}

\vspace{0.3cm}
\noindent \textit{NorCommonsenseQA}
\vspace{0.15cm}

\noindent NorCommonsenseQA is developed to assess the LM’s commonsense reasoning abilities. Each example consists of a question and five answer choices.

\vspace{0.1cm}
\colorbox{cb-grey}{Several annotation examples and explanations are provided here.}

\vspace{0.3cm}
\noindent \textit{NorTruthfulQA Multiple Choice}
\vspace{0.15cm}

\noindent This dataset is designed to evaluate if an LM selects answers that convey false beliefs or misconceptions. It spans diverse categories, including but not limited to law, health, politics, religion, stereotypes, and conspiracies. Each example includes a question and two to twelve answer options.

\vspace{0.1cm}
\textit{Disclaimer: you can find some examples sensitive.}

\vspace{0.15cm}
\colorbox{cb-grey}{Several annotation examples and explanations are provided here.}

\vspace{0.2cm}

\noindent Thank you once again for your time and contribution.

\vspace{0.5cm}
\begin{minipage}{15cm}

\par\noindent\rule{\textwidth}{1pt}
\textbf{Interface example}
\vspace{5pt}

Please carefully read the annotation guidelines before starting your annotation task.

\noindent Thank you for your contribution!

\vspace{0.2cm}

\colorbox{cb-grey}{This is a toy question.}

\vspace{5pt}
\begin{itemize}[noitemsep,topsep=0pt]
    \item[$\ocircle$] This is a toy answer option \#1
    \item[$\ocircle$] This is a toy answer option \#2
    \item[$\ocircle$] This is a toy answer option \#3
    \item[$\ocircle$] This is a toy answer option \#4
\end{itemize}

\vspace{0.1cm}

\par\noindent\rule{\textwidth}{1pt}
\end{minipage}

\subsection{Norwegian Comma Benchmark}

\subsubsection*{Overview}

You will be working on Norwegian Comma Benchmark, which is designed to evaluate the sensitivity of language models (LMs) to punctuation errors. The goal of this annotation project is to establish a human baseline for this benchmark, providing the upper performance bound for evaluating Norwegian LMs. 

\noindent You will receive a Google Form containing 50 pairs of sentences. Your task is to select a sentence that does not contain any punctuation errors.

\subsubsection*{Annotation task}

In general, you will need to:

\begin{enumerate}[noitemsep,topsep=0pt]
    \item Carefully read two sentences.
    \item Judge the acceptability of each sentence with respect to punctuation.
    \item Select a sentence that is correctly punctuated.
    \item Double-check your response and move onto the next example.
\end{enumerate}

\subsubsection*{Annotation examples}

Here, we provide you with annotation examples. Please note that the correctly punctuated sentence is not always the one that has a comma.

\vspace{0.1cm}
\colorbox{cb-grey}{Several annotation examples and explanations are provided here.}

\vspace{0.2cm}

\noindent Thank you once again for your time and contribution.

\vspace{0.5cm}
\begin{minipage}{15cm}

\par\noindent\rule{\textwidth}{1pt}
\textbf{Interface example}
\vspace{5pt}

Please carefully read the annotation guidelines before starting your annotation task.

\noindent Thank you for your contribution!

\vspace{0.2cm}

Which sentence does NOT contain any punctuation errors?

\vspace{5pt}
\begin{itemize}[noitemsep,topsep=0pt]
    \item[$\ocircle$] This is a toy sentence \#1
    \item[$\ocircle$] This is a toy sentence \#2
\end{itemize}

\vspace{0.1cm}

\par\noindent\rule{\textwidth}{1pt}
\end{minipage}

%% file: appendix/results.tex
\section{Empirical Evaluation Details}
\label{app:details}

\input{tables/clf_ranking_completion_results}

\input{tables/qa_results}

\input{tables/truthfulness_results}

\input{tables/norsumm_mt}

\input{tables/norinstruct_results}

%% file: tables/clf_ranking_completion_results.tex
\renewcommand{\arraystretch}{1.1}

\begin{table}[!th]
\centering
\scriptsize
\resizebox{\textwidth}{!}{%
\begin{tabular}{@{}lcccccc}
\toprule
\multirow{5}{*}{\textbf{Model}} & \multicolumn{4}{c}{\textbf{Norwegian language knowledge}} & \multicolumn{2}{c}{\textbf{Sentiment analysis}} \\ \cmidrule(l{2pt}r{2pt}){2-5} \cmidrule(l{2pt}r{2pt}){6-7}

& \textbf{NCB} & \textbf{NorIdiom} & \textbf{NorIdiom} & \textbf{ASK-GEC} & \textbf{NoReC Sentence} & \textbf{NoReC Document} \\
\cmidrule(l{2pt}r{2pt}){2-5} \cmidrule(l{2pt}r{2pt}){6-7}

& \textbf{Bokmål} & \textbf{Bokmål} & \textbf{Nynorsk} & \textbf{Bokmål} & \textbf{Bokmål} & \textbf{Bokmål}
\\
\cmidrule(l{2pt}r{2pt}){2-5} \cmidrule(l{2pt}r{2pt}){6-7}

& \textbf{Accuracy} & \textbf{EM} & \textbf{EM} & \textbf{ERRANT F\textsubscript{0.5}} & \textbf{F1-macro} & \textbf{F1-macro} \\
\midrule

NB-GPT-6B & 86.3 & 13.4 & 30.7 & \hphantom{0}5.7 & 64.8 & 67.3 \\
GPT-SW3-6.7B & 82.6 & \textbf{59.7} & \textbf{69.7} & 49.4 & 84.1 & 79.1 \\
NorwAI-Mistral-7B & 87.1 & 32.0 & 29.2 & \textbf{53.2} & 88.6 & 81.2 \\
NorwAI-Llama2-7B & \textbf{90.0} & 33.2 & 27.0 & 51.4 & 86.0 & 79.2 \\
NorBLOOM-7B-warm & 82.7 & 48.8 & 60.7 & 32.3 & 67.6 & 71.4 \\
NorMistral-7B-scratch & 81.2 & 43.5 & 65.2 & 41.7 & 80.3 & 75.9 \\
Viking-7B & 80.6 & 43.8 & 48.9 & 51.2 & 77.9 & 80.4 \\
NorMistral-11B & 85.6 & 15.8 & 32.6 & \underline{52.6} & \textbf{90.5} & 91.2 \\
Viking-13B & 85.7 & 44.9 & 58.4 & 52.4 & 79.2 & 86.8 \\\\[-0.9em]

NorMistral-7B-warm & 82.7 & \underline{56.1} & \underline{66.3} & 48.7 & 84.9 & 82.9 \\
NorMistral-7B-warm-IT  & \cellcolor[rgb]{0.838,0.861,0.896}\hphantom{\textsuperscript{\textit{($+$1.1)}}} 83.8 \textsuperscript{\textit{($+$1.1)}} & \cellcolor[rgb]{0.970,0.677,0.560}\hphantom{\textsuperscript{\textit{($-$56.1)}}} \hphantom{0}0.0 \textsuperscript{\textit{($-$56.1)}} & \cellcolor[rgb]{0.970,0.677,0.560}\hphantom{\textsuperscript{\textit{($-$66.3)}}} \hphantom{0}0.0 \textsuperscript{\textit{($-$66.3)}} & \cellcolor[rgb]{0.970,0.677,0.560}\hphantom{\textsuperscript{\textit{($-$48.6)}}} \hphantom{0}0.1 \textsuperscript{\textit{($-$48.6)}} & \cellcolor[rgb]{0.819,0.856,0.915}\hphantom{\textsuperscript{\textit{($+$1.8)}}} 86.7 \textsuperscript{\textit{($+$1.8)}} & \cellcolor[rgb]{0.704,0.803,0.984}\hphantom{\textsuperscript{\textit{($+$6.9)}}} 89.8 \textsuperscript{\textit{($+$6.9)}} \\\\[-0.9em]

Mistral-7B & 74.4 & \hphantom{0}5.7 & \hphantom{0}7.9 & 31.3 & 85.1 & 91.9 \\
Mistral-7B-IT  & \cellcolor[rgb]{0.835,0.861,0.899}\hphantom{\textsuperscript{\textit{($+$1.2)}}} 75.6 \textsuperscript{\textit{($+$1.2)}} & \cellcolor[rgb]{0.958,0.772,0.682}\hphantom{\textsuperscript{\textit{($-$5.7)}}} \hphantom{0}0.0 \textsuperscript{\textit{($-$5.7)}} & \cellcolor[rgb]{0.966,0.740,0.637}\hphantom{\textsuperscript{\textit{($-$7.9)}}} \hphantom{0}0.0 \textsuperscript{\textit{($-$7.9)}} & \cellcolor[rgb]{0.970,0.677,0.560}\hphantom{\textsuperscript{\textit{($-$31.2)}}} \hphantom{0}0.1 \textsuperscript{\textit{($-$31.2)}} & \cellcolor[rgb]{0.957,0.775,0.686}\hphantom{\textsuperscript{\textit{($-$5.5)}}} 79.6 \textsuperscript{\textit{($-$5.5)}} & \cellcolor[rgb]{0.930,0.822,0.763}\hphantom{\textsuperscript{\textit{($-$2.9)}}} 89.0 \textsuperscript{\textit{($-$2.9)}} \\\\[-0.9em]

AI-Sweden/Llama-3-8B & 83.7 & 31.3 & 52.8 & \underline{52.6} & 87.0 & 92.7 \\
AI-Sweden/Llama-3-8B-IT  & \cellcolor[rgb]{0.905,0.844,0.809}\hphantom{\textsuperscript{\textit{($-$1.6)}}} 82.1 \textsuperscript{\textit{($-$1.6)}} & \cellcolor[rgb]{0.970,0.677,0.560}\hphantom{\textsuperscript{\textit{($-$31.3)}}} \hphantom{0}0.0 \textsuperscript{\textit{($-$31.3)}} & \cellcolor[rgb]{0.970,0.677,0.560}\hphantom{\textsuperscript{\textit{($-$52.8)}}} \hphantom{0}0.0 \textsuperscript{\textit{($-$52.8)}} & \cellcolor[rgb]{0.970,0.677,0.560}\hphantom{\textsuperscript{\textit{($-$52.5)}}} \hphantom{0}0.1 \textsuperscript{\textit{($-$52.5)}} & \cellcolor[rgb]{0.860,0.865,0.872}\hphantom{\textsuperscript{\textit{($+$0.2)}}} 87.2 \textsuperscript{\textit{($+$0.2)}} & \cellcolor[rgb]{0.794,0.848,0.936}\hphantom{\textsuperscript{\textit{($+$2.8)}}} \textbf{95.5} \textsuperscript{\textit{($+$2.8)}} \\\\[-0.9em]

Meta/Llama-3-8B & 78.1 & 10.3 & \hphantom{0}5.7 & 41.5 & 84.9 & 91.3 \\
Meta/Llama-3-8B-IT  & \cellcolor[rgb]{0.887,0.855,0.837}\hphantom{\textsuperscript{\textit{($-$0.8)}}} 77.3 \textsuperscript{\textit{($-$0.8)}} & \cellcolor[rgb]{0.969,0.715,0.605}\hphantom{\textsuperscript{\textit{($-$10.3)}}} \hphantom{0}0.0 \textsuperscript{\textit{($-$10.3)}} & \cellcolor[rgb]{0.958,0.772,0.682}\hphantom{\textsuperscript{\textit{($-$5.7)}}} \hphantom{0}0.0 \textsuperscript{\textit{($-$5.7)}} & \cellcolor[rgb]{0.970,0.677,0.560}\hphantom{\textsuperscript{\textit{($-$41.4)}}} \hphantom{0}0.1 \textsuperscript{\textit{($-$41.4)}} & \cellcolor[rgb]{0.911,0.839,0.799}\hphantom{\textsuperscript{\textit{($-$1.9)}}} 83.0 \textsuperscript{\textit{($-$1.9)}} & \cellcolor[rgb]{0.781,0.843,0.945}\hphantom{\textsuperscript{\textit{($+$3.3)}}} \underline{94.6} \textsuperscript{\textit{($+$3.3)}} \\\\[-0.9em]

Mistral-Nemo-12B & 78.6 & \hphantom{0}0.7 & \hphantom{0}3.4 & 43.9 & 86.9 & 89.2 \\
Mistral-Nemo-12B-IT  & \cellcolor[rgb]{0.911,0.839,0.799}\hphantom{\textsuperscript{\textit{($-$1.9)}}} 76.7 \textsuperscript{\textit{($-$1.9)}} & \cellcolor[rgb]{0.773,0.839,0.950}\hphantom{\textsuperscript{\textit{($+$3.6)}}} \hphantom{0}4.3 \textsuperscript{\textit{($+$3.6)}} & \cellcolor[rgb]{0.781,0.843,0.945}\hphantom{\textsuperscript{\textit{($+$3.3)}}} \hphantom{0}6.7 \textsuperscript{\textit{($+$3.3)}} & \cellcolor[rgb]{0.970,0.677,0.560}\hphantom{\textsuperscript{\textit{($-$43.7)}}} \hphantom{0}0.2 \textsuperscript{\textit{($-$43.7)}} & \cellcolor[rgb]{0.835,0.861,0.899}\hphantom{\textsuperscript{\textit{($+$1.2)}}} \underline{88.1} \textsuperscript{\textit{($+$1.2)}} & \cellcolor[rgb]{0.738,0.822,0.970}\hphantom{\textsuperscript{\textit{($+$5.1)}}} 94.3 \textsuperscript{\textit{($+$5.1)}} \\ \midrule
Random & 50.0 & \hphantom{0}0.0 & \hphantom{0}0.0 & \hphantom{0}0.0 & 48.5 & 48.4 \\ 
Human & \underline{88.0} & \hphantom{0}\xmark & \hphantom{0}\xmark & \hphantom{0}\xmark & \xmark & \xmark \\
\bottomrule
\end{tabular} %
}
\caption{\textbf{Performance scores of the pretrained-only and instruction-finetuned Norwegian LMs on our Norwegian language knowledge and sentiment analysis tasks.} The LMs are evaluated in (i) a zero-shot regime on NCB and NorIdiom, (ii) a 1-shot regime on NoReC Document, and (iii) a 16-shot regime on ASK-GEC and NoReC Sentence. Cold-colored cells represent cases where an instruction-tuned version improves performance compared to the base LM, while warm-colored cells indicate cases where it decreases. The best score is in bold, the second best is underlined.}
\label{tab:results_clf_ranking_completion}
\end{table}

%% file: tables/qa_results.tex
\begin{table*}[!th]
\setlength{\tabcolsep}{4pt}
\scriptsize
\resizebox{\textwidth}{!}{%
\begin{tabular}{@{}lcccccccc}
\toprule

\multirow{4}{*}{\textbf{Model}} & \multicolumn{2}{c}{\begin{tabular}{@{}c@{}} \textbf{Machine reading} \\ \textbf{comprehension} \end{tabular}} & \multicolumn{4}{c}{\textbf{Norwegian-specific \& world knowledge}} &  \multicolumn{2}{c}{\textbf{Commonsense reasoning}} \\ 

\cmidrule(l{2pt}r{2pt}){2-3} \cmidrule(l{2pt}r{2pt}){4-7}  \cmidrule(l{2pt}r{2pt}){8-9}

& \textbf{Belebele} & \textbf{NorQuAD} & \multicolumn{2}{c}{\textbf{NRK-Quiz-QA}} & \multicolumn{2}{c}{\textbf{NorOpenBookQA}} &  \multicolumn{2}{c}{\textbf{NorCommonsenseQA}} \\  

\cmidrule(l{2pt}r{2pt}){2-3} \cmidrule(l{2pt}r{2pt}){4-7}  \cmidrule(l{2pt}r{2pt}){8-9}

& \textbf{Bokmål} & \textbf{Bokmål} & \textbf{Bokmål} & \textbf{Nynorsk} & \textbf{Bokmål} & \textbf{Nynorsk} & \textbf{Bokmål} & \textbf{Nynorsk} \\ 

 \cmidrule(l{2pt}r{2pt}){2-3} \cmidrule(l{2pt}r{2pt}){4-7}  \cmidrule(l{2pt}r{2pt}){8-9}
& \textbf{Accuracy} & \textbf{F1\textsubscript{a}} & \textbf{Accuracy} & \textbf{Accuracy} & \textbf{Accuracy} & \textbf{Accuracy} & \textbf{Accuracy} & \textbf{Accuracy} 

\\ \midrule
NB-GPT-6B & 29.2 & 33.8 & 53.8 & 60.4 & 44.1 & 33.3 & 48.8 & 35.8 \\
GPT-SW3-6.7B & 35.7 & 66.9 & 49.2 & 52.0 & 48.7 & 43.3 & 52.2 & 37.9 \\
NorwAI-Mistral-7B & 33.4 & 63.0 & 55.2 & 65.2 & 55.1 & 45.6 & 54.2 & 43.2 \\
NorwAI-Llama2-7B & 38.0 & 60.3 & 52.3 & 64.3 & 50.3 & 42.2 & 49.7 & 37.9 \\
NorBLOOM-7B-warm & 28.1 & 43.6 & 44.6 & 53.5 & 43.0 & 32.2 & 43.9 & 33.7 \\
NorMistral-7B-scratch & 25.7 & 43.7 & 48.2 & 57.0 & 44.1 & 30.0 & 47.5 & 36.8 \\
Viking-7B & 27.6 & 48.4 & 44.3 & 51.1 & 49.7 & 33.3 & 44.9 & 39.0 \\ 
NorMistral-11B & 56.7 & \underline{76.7} & \textbf{63.7} & \textbf{71.9} & 78.6 & \underline{82.2} & \underline{61.0} & \underline{51.6} \\
Viking-13B & 28.2 & 57.1 & 51.0 & 54.8 & 48.9 & 40.0 & 51.1 & 40.0 \\\\[-1em]

NorMistral-7B-warm & 37.4 & 64.8 & \underline{57.9} & \underline{65.9} & 51.3 & 43.3 & 51.3 & \underline{43.2} \\
NorMistral-7B-warm-IT  & \cellcolor[rgb]{0.665,0.778,0.994}\hphantom{\textsuperscript{\textit{($+$9.9)}}} 47.3 \textsuperscript{\textit{($+$9.9)}} & \cellcolor[rgb]{0.970,0.677,0.560}\hphantom{\textsuperscript{\textit{($-$47.7)}}} 17.1 \textsuperscript{\textit{($-$47.7)}} & \cellcolor[rgb]{0.877,0.861,0.851}\hphantom{\textsuperscript{\textit{($-$0.4)}}} 57.5 \textsuperscript{\textit{($-$0.4)}} & \cellcolor[rgb]{0.937,0.813,0.747}\hphantom{\textsuperscript{\textit{($-$3.4)}}} 62.5 \textsuperscript{\textit{($-$3.4)}} & \cellcolor[rgb]{0.629,0.752,0.999}\hphantom{\textsuperscript{\textit{($+$17.2)}}} 68.5 \textsuperscript{\textit{($+$17.2)}} & \cellcolor[rgb]{0.627,0.750,0.999}\hphantom{\textsuperscript{\textit{($+$18.9)}}} 62.2 \textsuperscript{\textit{($+$18.9)}} & \cellcolor[rgb]{0.817,0.856,0.917}\hphantom{\textsuperscript{\textit{($+$1.9)}}} 53.2 \textsuperscript{\textit{($+$1.9)}} & \cellcolor[rgb]{0.866,0.865,0.865}\hphantom{\textsuperscript{\textit{($-$0.0)}}} 43.2 \textsuperscript{\textit{($-$0.0)}} \\\\[-1em]

Mistral-7B & 42.7 & 70.7 & 42.5 & 39.5 & 80.0 & 72.2 & 41.2 & 32.6 \\
Mistral-7B-IT  & \cellcolor[rgb]{0.812,0.854,0.921}\hphantom{\textsuperscript{\textit{($+$2.1)}}} 44.8 \textsuperscript{\textit{($+$2.1)}} & \cellcolor[rgb]{0.970,0.677,0.560}\hphantom{\textsuperscript{\textit{($-$34.0)}}} 36.7 \textsuperscript{\textit{($-$34.0)}} & \cellcolor[rgb]{0.903,0.846,0.813}\hphantom{\textsuperscript{\textit{($-$1.5)}}} 41.0 \textsuperscript{\textit{($-$1.5)}} & \cellcolor[rgb]{0.952,0.786,0.703}\hphantom{\textsuperscript{\textit{($-$4.9)}}} 34.6 \textsuperscript{\textit{($-$4.9)}} & \cellcolor[rgb]{0.970,0.704,0.591}\hphantom{\textsuperscript{\textit{($-$11.8)}}} 68.2 \textsuperscript{\textit{($-$11.8)}} & \cellcolor[rgb]{0.966,0.742,0.639}\hphantom{\textsuperscript{\textit{($-$7.8)}}} 64.4 \textsuperscript{\textit{($-$7.8)}} & \cellcolor[rgb]{0.911,0.839,0.799}\hphantom{\textsuperscript{\textit{($-$1.9)}}} 39.3 \textsuperscript{\textit{($-$1.9)}} & \cellcolor[rgb]{0.866,0.865,0.865}\hphantom{\textsuperscript{\textit{($-$0.0)}}} 32.6 \textsuperscript{\textit{($-$0.0)}} \\\\[-1em]

AI-Sweden/Llama-3-8B & 54.3 & 74.4 & 55.8 & 58.4 & 78.6 & 74.4 & 54.7 & 41.0 \\
AI-Sweden/Llama-3-8B-IT  & \cellcolor[rgb]{0.624,0.748,0.999}\hphantom{\textsuperscript{\textit{($+$23.0)}}} 77.3 \textsuperscript{\textit{($+$23.0)}} & \cellcolor[rgb]{0.970,0.677,0.560}\hphantom{\textsuperscript{\textit{($-$35.4)}}} 39.0 \textsuperscript{\textit{($-$35.4)}} & \cellcolor[rgb]{0.931,0.820,0.760}\hphantom{\textsuperscript{\textit{($-$3.0)}}} 52.8 \textsuperscript{\textit{($-$3.0)}} & \cellcolor[rgb]{0.958,0.771,0.680}\hphantom{\textsuperscript{\textit{($-$5.8)}}} 52.6 \textsuperscript{\textit{($-$5.8)}} & \cellcolor[rgb]{0.717,0.811,0.979}\hphantom{\textsuperscript{\textit{($+$6.2)}}} 84.8 \textsuperscript{\textit{($+$6.2)}} & \cellcolor[rgb]{0.751,0.829,0.963}\hphantom{\textsuperscript{\textit{($+$4.5)}}} 78.9 \textsuperscript{\textit{($+$4.5)}} & \cellcolor[rgb]{0.629,0.752,0.999}\hphantom{\textsuperscript{\textit{($+$17.5)}}} 72.2 \textsuperscript{\textit{($+$17.5)}} & \cellcolor[rgb]{0.650,0.768,0.997}\hphantom{\textsuperscript{\textit{($+$11.6)}}} \textbf{52.6} \textsuperscript{\textit{($+$11.6)}} \\\\[-1em]

Meta/Llama-3-8B & 56.8 & 75.6 & 50.2 & 47.9 & 81.3 & 76.7 & 47.9 & 36.8 \\
Meta/Llama-3-8B-IT  & \cellcolor[rgb]{0.627,0.750,0.999}\hphantom{\textsuperscript{\textit{($+$19.0)}}} 75.8 \textsuperscript{\textit{($+$19.0)}} & \cellcolor[rgb]{0.970,0.681,0.565}\hphantom{\textsuperscript{\textit{($-$20.2)}}} 55.4 \textsuperscript{\textit{($-$20.2)}} & \cellcolor[rgb]{0.882,0.858,0.844}\hphantom{\textsuperscript{\textit{($-$0.6)}}} 49.6 \textsuperscript{\textit{($-$0.6)}} & \cellcolor[rgb]{0.925,0.827,0.774}\hphantom{\textsuperscript{\textit{($-$2.6)}}} 45.3 \textsuperscript{\textit{($-$2.6)}} & \cellcolor[rgb]{0.833,0.860,0.902}\hphantom{\textsuperscript{\textit{($+$1.3)}}} 82.6 \textsuperscript{\textit{($+$1.3)}} & \cellcolor[rgb]{0.754,0.831,0.961}\hphantom{\textsuperscript{\textit{($+$4.4)}}} 81.1 \textsuperscript{\textit{($+$4.4)}} & \cellcolor[rgb]{0.660,0.774,0.995}\hphantom{\textsuperscript{\textit{($+$10.4)}}} 58.3 \textsuperscript{\textit{($+$10.4)}} & \cellcolor[rgb]{0.695,0.798,0.987}\hphantom{\textsuperscript{\textit{($+$7.4)}}} 44.2 \textsuperscript{\textit{($+$7.4)}} \\\\[-1em]

Mistral-Nemo-12B & 62.8 & 76.5 & 47.4 & 47.2 & 84.8 & \textbf{88.9} & 46.9 & 33.7 \\
Mistral-Nemo-12B-IT  & \cellcolor[rgb]{0.629,0.752,0.999}\hphantom{\textsuperscript{\textit{($+$17.4)}}} \underline{80.2} \textsuperscript{\textit{($+$17.4)}} & \cellcolor[rgb]{0.970,0.687,0.571}\hphantom{\textsuperscript{\textit{($-$16.4)}}} 60.1 \textsuperscript{\textit{($-$16.4)}} & \cellcolor[rgb]{0.705,0.804,0.983}\hphantom{\textsuperscript{\textit{($+$6.8)}}} 54.2 \textsuperscript{\textit{($+$6.8)}} & \cellcolor[rgb]{0.743,0.825,0.967}\hphantom{\textsuperscript{\textit{($+$4.9)}}} 52.1 \textsuperscript{\textit{($+$4.9)}} & \cellcolor[rgb]{0.798,0.850,0.932}\hphantom{\textsuperscript{\textit{($+$2.6)}}} \textbf{87.4} \textsuperscript{\textit{($+$2.6)}} & \cellcolor[rgb]{0.935,0.815,0.751}\hphantom{\textsuperscript{\textit{($-$3.3)}}} 85.6 \textsuperscript{\textit{($-$3.3)}} & \cellcolor[rgb]{0.648,0.766,0.997}\hphantom{\textsuperscript{\textit{($+$12.0)}}} 58.9 \textsuperscript{\textit{($+$12.0)}} & \cellcolor[rgb]{0.629,0.752,0.999}\hphantom{\textsuperscript{\textit{($+$17.9)}}} \underline{51.6} \textsuperscript{\textit{($+$17.9)}}\\ \midrule
Random & 25.0 & 0.0 & 27.9 & 26.8 & 25.0  & 25.0  & 20.0 & 20.0 \\ 
Human & \textbf{90.0} & \textbf{91.1} & \xmark & \xmark & \textbf{100.0} & \xmark & \textbf{90.0} & \xmark \\
\bottomrule
\end{tabular}
}
\caption{\textbf{Performance scores of the pretrained-only and instruction-finetuned Norwegian LMs on our machine reading comprehension, Norwegian-specific \& world knowledge, and commonsense reasoning tasks.} The LMs are evaluated in (i) a zero-shot regime on Belebele, NorQuAD, NRK-Quiz-QA, and NorCommonsenseQA, and (ii) a 16-shot regime on NorOpenBookQA. Cold-colored cells represent cases where an instruction-tuned version improves performance compared to the base LM, while warm-colored cells indicate cases where it decreases. The best score is in bold, the second best is underlined. The human baseline on NorQuAD is from \citet{ivanova-etal-2023-norquad}.}
\label{tab:results_qa}
\end{table*}

%% file: tables/truthfulness_results.tex
\renewcommand{\arraystretch}{1.25}
\begin{table}[!th]
\centering
\scriptsize
\begin{tabular}{@{}lcccc}
\toprule
\multirow{4}{*}{\textbf{Model}} & \multicolumn{4}{c}{\textbf{Truthfulness}} \\
\cmidrule(l{2pt}r{2pt}){2-5}
& \multicolumn{2}{c}{\begin{tabular}{@{}c@{}} \textbf{NorTruthfulQA} \\ \textbf{Multiple Choice} \end{tabular}} & \multicolumn{2}{c}{\begin{tabular}{@{}c@{}} \textbf{NorTruthfulQA} \\ \textbf{Generation} \end{tabular}} \\
\cmidrule(l{2pt}r{2pt}){2-3} \cmidrule(l{2pt}r{2pt}){4-5}
& \textbf{Bokmål} & \textbf{Nynorsk} & \textbf{Bokmål} & \textbf{Nynorsk} \\
\cmidrule(l{2pt}r{2pt}){2-3} \cmidrule(l{2pt}r{2pt}){4-5}
& \textbf{Accuracy} & \textbf{Accuracy} & \textbf{ROUGE-L} & \textbf{ROUGE-L} \\
\midrule
NB-GPT-6B & 57.4 & 57.9 & 22.0 & 23.0 \\
GPT-SW3-6.7B & 69.7 & 66.7 & 30.9 & \textbf{29.6} \\
NorwAI-Mistral-7B & 69.9 & 61.4 & 20.5 & 17.9 \\
NorwAI-Llama2-7B & 53.3 & 54.4 & 21.1 & 22.9 \\
NorBLOOM-7B-warm & 62.9 & 61.4 & 28.7 & 28.7 \\
NorMistral-7B-scratch & 68.0 & 59.6 & 29.4 & 28.0 \\
Viking-7B & 52.0 & 45.6 & 21.3 & 21.6 \\
NorMistral-11B & 48.0 & 38.6 & 20.9 & 24.0 \\
Viking-13B & 58.6 & 49.1 & 18.3 & 18.0 \\\\[-1em]

NorMistral-7B-warm & 55.5 & 50.9 & 26.4 & 24.7 \\
NorMistral-7B-warm-IT  & \cellcolor[rgb]{0.955,0.779,0.692}\hphantom{\textsuperscript{\textit{($-$5.3)}}} 50.2 \textsuperscript{\textit{($-$5.3)}} & \cellcolor[rgb]{0.938,0.811,0.744}\hphantom{\textsuperscript{\textit{($-$3.5)}}} 47.4 \textsuperscript{\textit{($-$3.5)}} & \cellcolor[rgb]{0.969,0.725,0.617}\hphantom{\textsuperscript{\textit{($-$9.2)}}} 17.2 \textsuperscript{\textit{($-$9.2)}} & \cellcolor[rgb]{0.963,0.755,0.657}\hphantom{\textsuperscript{\textit{($-$6.8)}}} 17.9 \textsuperscript{\textit{($-$6.8)}} \\\\[-1em]

Mistral-7B & 74.6 & 73.7 & 25.8 & 27.0 \\
Mistral-7B-IT  & \cellcolor[rgb]{0.970,0.679,0.561}\hphantom{\textsuperscript{\textit{($-$22.6)}}} 52.0 \textsuperscript{\textit{($-$22.6)}} & \cellcolor[rgb]{0.970,0.684,0.568}\hphantom{\textsuperscript{\textit{($-$17.6)}}} 56.1 \textsuperscript{\textit{($-$17.6)}} & \cellcolor[rgb]{0.804,0.852,0.928}\hphantom{\textsuperscript{\textit{($+$2.4)}}} 28.2 \textsuperscript{\textit{($+$2.4)}} & \cellcolor[rgb]{0.956,0.777,0.689}\hphantom{\textsuperscript{\textit{($-$5.4)}}} 21.6 \textsuperscript{\textit{($-$5.4)}} \\\\[-1em]

AI-Sweden/Llama-3-8B & 52.5 & 52.6 & 27.4 & 24.8 \\
AI-Sweden/Llama-3-8B-IT  & \cellcolor[rgb]{0.970,0.680,0.563}\hphantom{\textsuperscript{\textit{($-$20.5)}}} 32.0 \textsuperscript{\textit{($-$20.5)}} & \cellcolor[rgb]{0.970,0.681,0.565}\hphantom{\textsuperscript{\textit{($-$19.3)}}} 33.3 \textsuperscript{\textit{($-$19.3)}} & \cellcolor[rgb]{0.970,0.692,0.577}\hphantom{\textsuperscript{\textit{($-$14.2)}}} 13.2 \textsuperscript{\textit{($-$14.2)}} & \cellcolor[rgb]{0.969,0.725,0.617}\hphantom{\textsuperscript{\textit{($-$9.2)}}} 15.6 \textsuperscript{\textit{($-$9.2)}} \\\\[-1em]

Meta/Llama-3-8B & 57.0 & 54.4 & 28.5 & 25.9 \\
Meta/Llama-3-8B-IT  & \cellcolor[rgb]{0.751,0.829,0.963}\hphantom{\textsuperscript{\textit{($+$4.5)}}} 61.5 \textsuperscript{\textit{($+$4.5)}} & \cellcolor[rgb]{0.627,0.750,0.999}\hphantom{\textsuperscript{\textit{($+$19.3)}}} \textbf{73.7} \textsuperscript{\textit{($+$19.3)}} & \cellcolor[rgb]{0.934,0.817,0.754}\hphantom{\textsuperscript{\textit{($-$3.2)}}} 25.3 \textsuperscript{\textit{($-$3.2)}} & \cellcolor[rgb]{0.963,0.755,0.657}\hphantom{\textsuperscript{\textit{($-$6.8)}}} 19.1 \textsuperscript{\textit{($-$6.8)}} \\\\[-1em]

Mistral-Nemo-12B & 54.1 & 49.1 & 25.3 & 22.6 \\
Mistral-Nemo-12B-IT  & \cellcolor[rgb]{0.641,0.761,0.998}\hphantom{\textsuperscript{\textit{($+$13.3)}}} 67.4 \textsuperscript{\textit{($+$13.3)}} & \cellcolor[rgb]{0.629,0.752,0.999}\hphantom{\textsuperscript{\textit{($+$17.6)}}} 66.7 \textsuperscript{\textit{($+$17.6)}} & \cellcolor[rgb]{0.710,0.807,0.981}\hphantom{\textsuperscript{\textit{($+$6.5)}}} \textbf{31.8} \textsuperscript{\textit{($+$6.5)}} & \cellcolor[rgb]{0.763,0.835,0.956}\hphantom{\textsuperscript{\textit{($+$4.0)}}} 26.6 \textsuperscript{\textit{($+$4.0)}} \\\midrule
Random & 27.3 & 24.6 & \xmark & \xmark \\
Human & \textbf{83.3} & \xmark & \xmark & \xmark \\
\bottomrule
\end{tabular} %
\caption{\textbf{Performance scores of the pretrained-only and instruction-finetuned Norwegian LMs on our truthfulness tasks.} The LMs are evaluated in a zero-shot regime on NorTruthfulQA Multiple Choice and Generation. Cold-colored cells represent cases where an instruction-tuned version improves performance compared to the base LM, while warm-colored cells indicate cases where it decreases. The best score is in bold, the second best is underlined.}
\label{tab:results_truthfulness}
\end{table}

%% file: tables/norsumm_mt.tex
\begin{table*}[!th]
\setlength{\tabcolsep}{1.8pt}
\scriptsize
\resizebox{\textwidth}{!}{%
\begin{tabular}{lcccccccc}
\toprule

\multirow{4}{*}{\textbf{Model}} & \multicolumn{4}{c}{\textbf{Text summarization}} & \multicolumn{4}{c}{\textbf{Machine Translation}}  \\ 
\cmidrule(l{2pt}r{2pt}){2-5} \cmidrule(l{2pt}r{2pt}){6-9} 

& \multicolumn{2}{c}{\textbf{NorSumm (BM)}} & \multicolumn{2}{c}{\textbf{NorSumm (NN)}} & \multicolumn{2}{c}{\textbf{Tatoeba (En $\rightarrow$ BM)}}  & \multicolumn{2}{c}{\textbf{Tatoeba (En $\rightarrow$ NN)}} \\

\cmidrule(l{2pt}r{2pt}){2-3} \cmidrule(l{2pt}r{2pt}){4-5}  \cmidrule(l{2pt}r{2pt}){6-7} \cmidrule(l{2pt}r{2pt}){8-9}
& \textbf{ROUGE-L} & \textbf{BERTScore} & \textbf{ROUGE-L} & \textbf{BERTScore} & \textbf{BLEU} & \textbf{BERTScore} & \textbf{BLEU} & \textbf{BERTScore} 
\\ \midrule

NB-GPT-6B & 20.9 & 59.4 & 18.2 & 58.6 & 20.2 & 90.5 & 19.9 & 89.8 \\
GPT-SW3-6.7B & 22.8 & 60.3 & 17.5 & 50.4 & 59.4 & 94.4 & 44.8 & 91.9 \\
NorwAI-Mistral-7B & 11.9 & 53.5 & 10.4 & 52.0 & 58.7 & 94.3 & 47.4 & 92.4 \\
NorwAI-Llama2-7B & 14.6 & 60.8 & 14.1 & 60.6 & 57.9 & 94.2 & 47.4 & 92.3 \\
NorBLOOM-7B-warm & 19.1 & 55.0 & 16.6 & 51.5 & 52.3 & 93.0 & 39.7 & 90.3 \\
NorMistral-7B-scratch & 20.7 & 57.7 & 15.0 & 50.3 & 53.4 & 93.3 & 41.3 & 91.0 \\
Viking-7B & 30.4 & 70.5 & 26.0 & 70.8 & 59.7 & 94.5 & 45.6 & 92.2 \\
NorMistral-11B & 34.9 & 73.1 & 28.7 & 70.3 & 58.8 & 94.3 & \textbf{48.0} & \textbf{92.6} \\
Viking-13B & 31.2 & 70.5 & 26.0 & 69.5 & \textbf{60.0} & \textbf{94.6} & 45.6 & 92.2 \\\\[-1em]

NorMistral-7B-warm & 19.4 & 51.7 & 10.9 & 49.9 & 57.2 & 94.1 & 44.7 & 91.9 \\
NorMistral-7B-warm-IT  & \cellcolor[rgb]{0.625,0.749,0.999}\hphantom{\textsuperscript{\textit{($+$18.4)}}} 37.8 \textsuperscript{\textit{($+$18.4)}} & \cellcolor[rgb]{0.624,0.748,0.999}\hphantom{\textsuperscript{\textit{($+$22.3)}}} 74.0 \textsuperscript{\textit{($+$22.3)}} & \cellcolor[rgb]{0.622,0.747,1.000}\hphantom{\textsuperscript{\textit{($+$23.7)}}} \textbf{34.6} \textsuperscript{\textit{($+$23.7)}} & \cellcolor[rgb]{0.622,0.747,1.000}\hphantom{\textsuperscript{\textit{($+$22.8)}}} 72.7 \textsuperscript{\textit{($+$22.8)}} & \cellcolor[rgb]{0.970,0.677,0.560}\hphantom{\textsuperscript{\textit{($-$56.9)}}} \hphantom{0}0.3 \textsuperscript{\textit{($-$56.9)}} & \cellcolor[rgb]{0.970,0.677,0.560}\hphantom{\textsuperscript{\textit{($-$30.4)}}} 63.7 \textsuperscript{\textit{($-$30.4)}} & \cellcolor[rgb]{0.970,0.677,0.560}\hphantom{\textsuperscript{\textit{($-$43.8)}}} \hphantom{0}0.9 \textsuperscript{\textit{($-$43.8)}} & \cellcolor[rgb]{0.970,0.677,0.560}\hphantom{\textsuperscript{\textit{($-$34.7)}}} 57.2 \textsuperscript{\textit{($-$34.7)}} \\\\[-1em]

Mistral-7B & 9.9 & 53.3 & 8.9 & 51.4 & 36.6 & 90.6 & 16.3 & 86.7 \\
Mistral-7B-IT  & \cellcolor[rgb]{0.700,0.801,0.985}\hphantom{\textsuperscript{\textit{($+$14.7)}}} 24.6 \textsuperscript{\textit{($+$14.7)}} & \cellcolor[rgb]{0.627,0.750,0.999}\hphantom{\textsuperscript{\textit{($+$18.1)}}} 71.4 \textsuperscript{\textit{($+$18.1)}} & \cellcolor[rgb]{0.717,0.811,0.979}\hphantom{\textsuperscript{\textit{($+$9.1)}}} 18.0 \textsuperscript{\textit{($+$9.1)}} & \cellcolor[rgb]{0.624,0.748,0.999}\hphantom{\textsuperscript{\textit{($+$19.5)}}} 70.9 \textsuperscript{\textit{($+$19.5)}} & \cellcolor[rgb]{0.970,0.677,0.560}\hphantom{\textsuperscript{\textit{($-$29.2)}}} \hphantom{0}7.4 \textsuperscript{\textit{($-$29.2)}} & \cellcolor[rgb]{0.963,0.756,0.659}\hphantom{\textsuperscript{\textit{($-$6.7)}}} 83.9 \textsuperscript{\textit{($-$6.7)}} & \cellcolor[rgb]{0.970,0.692,0.577}\hphantom{\textsuperscript{\textit{($-$14.4)}}} \hphantom{0}1.9 \textsuperscript{\textit{($-$14.4)}} & \cellcolor[rgb]{0.969,0.719,0.609}\hphantom{\textsuperscript{\textit{($-$9.8)}}} 76.9 \textsuperscript{\textit{($-$9.8)}} \\\\[-1em]

AI-Sweden/Llama-3-8B & 36.7 & 73.3 & 30.3 & 71.4 & 58.5 & 94.3 & 41.9 & 91.2 \\
AI-Sweden/Llama-3-8B-IT  & \cellcolor[rgb]{0.970,0.691,0.575}\hphantom{\textsuperscript{\textit{($-$12.2)}}} 24.5 \textsuperscript{\textit{($-$12.2)}} & \cellcolor[rgb]{0.910,0.840,0.801}\hphantom{\textsuperscript{\textit{($+$0.0)}}} 73.3 \textsuperscript{\textit{($+$0.0)}} & \cellcolor[rgb]{0.968,0.730,0.623}\hphantom{\textsuperscript{\textit{($-$8.1)}}} 22.2 \textsuperscript{\textit{($-$8.1)}} & \cellcolor[rgb]{0.683,0.790,0.990}\hphantom{\textsuperscript{\textit{($+$1.3)}}} 72.7 \textsuperscript{\textit{($+$1.3)}} & \cellcolor[rgb]{0.970,0.677,0.560}\hphantom{\textsuperscript{\textit{($-$52.3)}}} \hphantom{0}6.2 \textsuperscript{\textit{($-$52.3)}} & \cellcolor[rgb]{0.970,0.693,0.578}\hphantom{\textsuperscript{\textit{($-$14.0)}}} 80.3 \textsuperscript{\textit{($-$14.0)}} & \cellcolor[rgb]{0.970,0.677,0.560}\hphantom{\textsuperscript{\textit{($-$40.7)}}} \hphantom{0}1.2 \textsuperscript{\textit{($-$40.7)}} & \cellcolor[rgb]{0.970,0.679,0.561}\hphantom{\textsuperscript{\textit{($-$23.2)}}} 68.0 \textsuperscript{\textit{($-$23.2)}} \\\\[-1em]

Meta/Llama-3-8B & 37.2 & 73.8 & 29.6 & 71.5 & 47.8 & 92.5 & 34.5 & 89.7 \\
Meta/Llama-3-8B-IT  & \cellcolor[rgb]{0.970,0.714,0.603}\hphantom{\textsuperscript{\textit{($-$7.1)}}} 30.1 \textsuperscript{\textit{($-$7.1)}} & \cellcolor[rgb]{0.830,0.859,0.905}\hphantom{\textsuperscript{\textit{($+$1.4)}}} 75.2 \textsuperscript{\textit{($+$1.4)}} & \cellcolor[rgb]{0.910,0.840,0.801}\hphantom{\textsuperscript{\textit{($-$3.5)}}} 26.1 \textsuperscript{\textit{($-$3.5)}} & \cellcolor[rgb]{0.817,0.856,0.917}\hphantom{\textsuperscript{\textit{($+$1.6)}}} 73.1 \textsuperscript{\textit{($+$1.6)}} & \cellcolor[rgb]{0.970,0.684,0.568}\hphantom{\textsuperscript{\textit{($-$17.7)}}} 30.1 \textsuperscript{\textit{($-$17.7)}} & \cellcolor[rgb]{0.952,0.786,0.703}\hphantom{\textsuperscript{\textit{($-$4.9)}}} 87.6 \textsuperscript{\textit{($-$4.9)}} & \cellcolor[rgb]{0.970,0.677,0.560}\hphantom{\textsuperscript{\textit{($-$31.3)}}} \hphantom{0}3.2 \textsuperscript{\textit{($-$31.3)}} & \cellcolor[rgb]{0.970,0.680,0.563}\hphantom{\textsuperscript{\textit{($-$22.0)}}} 67.7 \textsuperscript{\textit{($-$22.0)}} \\\\[-1em]

Mistral-Nemo-12B & 34.0 & 72.5 & 27.8 & 69.2 & 49.5 & 92.9 & 35.7 & 90.1 \\
Mistral-Nemo-12B-IT  & \cellcolor[rgb]{0.700,0.801,0.985}\hphantom{\textsuperscript{\textit{($+$7.1)}}} 41.1 \textsuperscript{\textit{($+$7.1)}} & \cellcolor[rgb]{0.809,0.853,0.923}\hphantom{\textsuperscript{\textit{($+$3.8)}}} \textbf{76.3} \textsuperscript{\textit{($+$3.8)}} & \cellcolor[rgb]{0.717,0.811,0.979}\hphantom{\textsuperscript{\textit{($+$9.0)}}} 36.8 \textsuperscript{\textit{($+$9.0)}} & \cellcolor[rgb]{0.745,0.826,0.966}\hphantom{\textsuperscript{\textit{($+$4.8)}}} \textbf{75.0} \textsuperscript{\textit{($+$4.8)}} & \cellcolor[rgb]{0.970,0.677,0.560}\hphantom{\textsuperscript{\textit{($-$42.1)}}} \hphantom{0}7.4 \textsuperscript{\textit{($-$42.1)}} & \cellcolor[rgb]{0.879,0.860,0.848}\hphantom{\textsuperscript{\textit{($-$0.5)}}} 92.4 \textsuperscript{\textit{($-$0.5)}} & \cellcolor[rgb]{0.970,0.677,0.560}\hphantom{\textsuperscript{\textit{($-$33.3)}}} \hphantom{0}2.4 \textsuperscript{\textit{($-$33.3)}} & \cellcolor[rgb]{0.970,0.684,0.568}\hphantom{\textsuperscript{\textit{($-$17.7)}}} 72.4 \textsuperscript{\textit{($-$17.7)}} \\
\bottomrule
\end{tabular}
}
\caption{\textbf{Performance scores of the pretrained-only and instruction-finetuned Norwegian LMs on our text summarization and machine translation tasks.} The LMs are evaluated in (i) a zero-shot regime on NorSumm and (ii) a 16-shot regime on Tatoeba. En=English; BM=Norwegian Bokmål; NN=Norwegian Nynorsk. Cold-colored cells represent cases where an instruction-tuned version improves performance compared to the base LM, while warm-colored cells indicate cases where it decreases. The best score is in bold, the second best is underlined.}
\label{tab:results_summ_and_mt}
\end{table*}

%% file: tables/norinstruct_results.tex
\begin{table*}[!th]
\setlength{\tabcolsep}{2.5pt}
\footnotesize
\centering
\begin{tabular}{@{}lcccc}
\toprule
\multirow{4}{*}{\textbf{Model}} & \multicolumn{2}{c}{\textbf{Text Summarization}} & \multicolumn{2}{c}{\textbf{Text Rewriting}} \\

\cmidrule(l{2pt}r{2pt}){2-3} \cmidrule(l{2pt}r{2pt}){4-5}
& \multicolumn{2}{c}{\textbf{NorSummarize-Instruct}} & \multicolumn{2}{c}{\textbf{NorRewrite-Instruct}} \\

\cmidrule(l{2pt}r{2pt}){2-3} \cmidrule(l{2pt}r{2pt}){4-5}
& \textbf{chrF} & \textbf{BERTScore} & \textbf{chrF} & \textbf{BERTScore} \\
\midrule
NB-GPT-6B & 23.8 & 57.0 & 19.5 & 56.1 \\
GPT-SW3-6.7B & 20.7 & 54.4 & 18.2 & 48.9 \\
NorwAI-Mistral-7B & 22.2 & 54.7 & 20.4 & 53.6 \\
NorwAI-Llama2-7B & 21.6 & 53.7 & 21.1 & 54.3 \\
NorBLOOM-7B-warm & 9.0 & 24.0 & 5.2 & 17.2 \\
NorMistral-7B-scratch & 8.5 & 24.0 & 7.2 & 20.0 \\
Viking-7B & 21.4 & 55.7 & 21.8 & 55.7 \\
NorMistral-11B & 27.2 & 61.4 & 25.7 & 71.0 \\
Viking-13B & 21.1 & 55.4 & 22.8 & 56.0 \\\\[-1em]

NorMistral-7B-warm & 6.7 & 22.1 & 6.7 & 23.1 \\
NorMistral-7B-warm-IT  & \cellcolor[rgb]{0.622,0.747,1.000}\hphantom{\textsuperscript{\textit{($+$34.7)}}} \textbf{41.4} \textsuperscript{\textit{($+$34.7)}} & \cellcolor[rgb]{0.622,0.747,1.000}\hphantom{\textsuperscript{\textit{($+$49.1)}}} 71.2 \textsuperscript{\textit{($+$49.1)}} & \cellcolor[rgb]{0.622,0.747,1.000}\hphantom{\textsuperscript{\textit{($+$34.5)}}} \textbf{41.2} \textsuperscript{\textit{($+$34.5)}} & \cellcolor[rgb]{0.622,0.747,1.000}\hphantom{\textsuperscript{\textit{($+$47.6)}}} 70.7 \textsuperscript{\textit{($+$47.6)}} \\\\[-1em]

Mistral-7B & 5.7 & 15.9 & 6.0 & 18.8 \\
Mistral-7B-IT  & \cellcolor[rgb]{0.624,0.748,0.999}\hphantom{\textsuperscript{\textit{($+$26.0)}}} 31.7 \textsuperscript{\textit{($+$26.0)}} & \cellcolor[rgb]{0.622,0.747,1.000}\hphantom{\textsuperscript{\textit{($+$54.4)}}} 70.3 \textsuperscript{\textit{($+$54.4)}} & \cellcolor[rgb]{0.624,0.748,0.999}\hphantom{\textsuperscript{\textit{($+$23.5)}}} 29.5 \textsuperscript{\textit{($+$23.5)}} & \cellcolor[rgb]{0.622,0.747,1.000}\hphantom{\textsuperscript{\textit{($+$51.2)}}} 70.0 \textsuperscript{\textit{($+$51.2)}} \\\\[-1em]

AI-Sweden/Llama-3-8B & 21.2 & 54.4 & 21.9 & 55.0 \\
AI-Sweden/Llama-3-8B-IT  & \cellcolor[rgb]{0.654,0.771,0.996}\hphantom{\textsuperscript{\textit{($+$11.1)}}} 32.3 \textsuperscript{\textit{($+$11.1)}} & \cellcolor[rgb]{0.637,0.758,0.998}\hphantom{\textsuperscript{\textit{($+$14.4)}}} 68.8 \textsuperscript{\textit{($+$14.4)}} & \cellcolor[rgb]{0.682,0.789,0.990}\hphantom{\textsuperscript{\textit{($+$8.4)}}} 30.3 \textsuperscript{\textit{($+$8.4)}} & \cellcolor[rgb]{0.639,0.759,0.998}\hphantom{\textsuperscript{\textit{($+$13.8)}}} 68.8 \textsuperscript{\textit{($+$13.8)}} \\\\[-1em]

Meta/Llama-3-8B & 21.8 & 55.4 & 20.4 & 52.0 \\
Meta/Llama-3-8B-IT  & \cellcolor[rgb]{0.640,0.760,0.998}\hphantom{\textsuperscript{\textit{($+$13.6)}}} 35.4 \textsuperscript{\textit{($+$13.6)}} & \cellcolor[rgb]{0.631,0.753,0.999}\hphantom{\textsuperscript{\textit{($+$16.5)}}} 71.9 \textsuperscript{\textit{($+$16.5)}} & \cellcolor[rgb]{0.669,0.781,0.993}\hphantom{\textsuperscript{\textit{($+$9.5)}}} 29.9 \textsuperscript{\textit{($+$9.5)}} & \cellcolor[rgb]{0.631,0.753,0.999}\hphantom{\textsuperscript{\textit{($+$16.5)}}} 68.5 \textsuperscript{\textit{($+$16.5)}} \\\\[-1em]

Mistral-Nemo-12B & 18.7 & 47.3 & 18.1 & 49.9 \\
Mistral-Nemo-12B-IT  & \cellcolor[rgb]{0.625,0.749,0.999}\hphantom{\textsuperscript{\textit{($+$21.2)}}} 39.9 \textsuperscript{\textit{($+$21.2)}} & \cellcolor[rgb]{0.624,0.748,0.999}\hphantom{\textsuperscript{\textit{($+$24.9)}}} \textbf{72.2} \textsuperscript{\textit{($+$24.9)}} & \cellcolor[rgb]{0.625,0.749,0.999}\hphantom{\textsuperscript{\textit{($+$20.8)}}} 38.9 \textsuperscript{\textit{($+$20.8)}} & \cellcolor[rgb]{0.625,0.749,0.999}\hphantom{\textsuperscript{\textit{($+$21.9)}}} \textbf{71.8} \textsuperscript{\textit{($+$21.9)}} \\
\bottomrule
\end{tabular} %
\caption{\textbf{Performance scores of the pretrained-only and instruction-finetuned Norwegian LMs on our instruction-style text summarization and rewriting tasks.} The LMs are evaluated in a zero-shot regime on NorSummarize-Instruct and NorRewrite-Instruct. Cold-colored cells represent cases where an instruction-tuned version improves performance compared to the base LM, while warm-colored cells indicate cases where it decreases. The best score is in bold, the second best is underlined.}
\label{tab:results_norinstruct}
\end{table*}

%% file: appendix/judge.tex
\onecolumn 

\section{Automatic Evaluation of Instruction-tuned LMs via LLM-as-a-judge}
\label{app:judge}

We evaluate the instruction-following abilities of the instruction-tuned LMs prompted for creative rewriting and summarization. Such generative tasks are difficult to evaluate even with access to the gold standard references. We use the LLM-as-a-judge approach, which involves a side-by-side comparison of LMs' responses using an external judge LM. While judge models suffer from various biases \citep{chen-etal-2024-humans,wang-etal-2024-large-language-models-fair, li2025generationjudgmentopportunitieschallenges}, they correlate with human judgments better than traditional language generation performance metrics \citep{10.1145/3485766, zheng2023judging}.

\paragraph{Expected win-rate scores} Given an instruction $i$, two  outputs $o_A$ and $o_B$ from LMs $A$ and $B$, and a human  reference $o_R$, the judge model $\theta$ computes a score function:
\begin{equation}
s_{\theta}(i, o_A, o_B, o_R) = \begin{cases}
    1,& \text{if } o_A \succ_{\theta} o_B\text{\hspace{1em}($\theta$ prefers $o_A$ over $o_B$)}\\
    0,& \text{if } o_A \prec_{\theta} o_B\\
    \nicefrac{1}{2},& \text{otherwise.}
\end{cases}
\label{eq:judge-score}
\end{equation}

\noindent Using this, we can compute the \textit{expected win-rate} of LM $A$ over LM $B$ as the expected value of the score function over a distribution $\mathcal{D}$ of prompts and human references:

\begin{equation}
\mathrm{win\_rate}_{\theta}(A, B) = \,\frac{1}{2} \left(1 + \!\!\!\mathop{\mathbb{E}}_{i, o\,\sim\,\mathcal{D}}\!\!s_{\theta}\left(i, A(i), B(i), o\right) -\!\!\!\mathop{\mathbb{E}}_{i, o\,\sim\,\mathcal{D}}\!\!s_{\theta}\left(i, B(i), A(i), o\right) \right)
\label{eq:winrate}
\end{equation}
where the second symmetric term prevents position bias \citep{wang-etal-2024-large-language-models-fair} from influencing the results. 

\paragraph{Judge Model} Unlike \newcite{lyu2024href}, we use simple chain-of-thought prompting by asking the model to first describe the qualities of each response before giving the final verdict -- this is done to further improve the evaluation accuracy \citep{NEURIPS2022_9d560961}. The judge is instructed to end its output by either generating ``A'' (for preference of response $A$), ``B'' (for preference of response $B$), or ``tie'' (for cases when both responses are either equally good or bad). We then parse the output and assign a score value according to \Cref{eq:judge-score}. A response pair is skipped in case of an incorrectly formatted judgment, which has not empirically occurred in our experiments.

\subsection{Evaluating LLM-as-a-judge}
\label{app:judge-bias}

\renewcommand{\arraystretch}{1.3}

\begin{table}[h!]
\centering
\begin{tabular}{@{}lccc@{}}
\toprule
 & \textsc{\textbf{A}} & \textsc{\textbf{Tie}} & \textsc{\textbf{B}} \\
\midrule 
\textsc{\textbf{A}}   & 35.42\%  & 8.33\% & 12.50\%  \\
\textsc{\textbf{Tie}}   & 8.33\%  & 4.17\%  & 4.17\%  \\
\textsc{\textbf{B}} & 10.42\%  & 6.25\% & 10.42\% \\
\bottomrule
\end{tabular}
\caption{\textbf{Agreement between the judge model and humans.} We compute the rates using prompt template from \cref{app:judge-prompt} without human references.}
\label{tab:agreement_rate}
\end{table}

\noindent We analyze the agreement rate between the judge model and humans, as well as self-preference bias on an BM subset of Primeape,\footnote{An ongoing work on multilingual human preference prediction and explanation at the moment of writing a camera-ready version of this paper; the data is provided by the dataset creators in response to our request. To be available at \href{https://github.com/Toloka/primeape}{\texttt{github.com/Toloka/primeape}}.} and explore language and position biases on NorRewrite-Instruct and NorSummarize-Instruct. The Primeape subset consists of 48 pairwise comparisons between \href{https://huggingface.co/meta-llama/Llama-3.3-70B-Instruct}{\texttt{meta-llama/Llama-3.3-70B-Instruct}}, \href{https://huggingface.co/google/gemma-2-27b-it}{\texttt{google/gemma-2-27b-it}}, \href{https://huggingface.co/mistralai/Mistral-Small-Instruct-2409}{\texttt{mistralai/Mistral-Small-Instruct-2409}}, \href{https://assets.anthropic.com/m/785e231869ea8b3b/original/claude-3-7-sonnet-system-card.pdf}{\texttt{Claude 3.7 Sonnet}}, and \texttt{GPT-4o} \cite{hurst2024gpt}, evaluated over 19 manually translated instructions from AlpacaEval \cite{dubois2023alpacafarm}. 

\paragraph{Agreement with Human Annotators} The agreement matrix is shown in \autoref{tab:agreement_rate}. The results indicate that the judge model agrees with human annotators in 50\% of the cases, compared to a random agreement rate of 33.3\%. When ties are excluded, the agreement rate increases to 66.7\%, with the corresponding random baseline of 50\%. The results and their interpretation can be affected by the limited sample size.

\paragraph{Self-preference Bias} We evaluate the self-preference bias \cite{panickssery2024llm} of \texttt{meta-llama/Llama-3.3-70B-Instruct} to analyze how often the judge model favors its own outputs. Among 32 pairwise comparisons, the judge's verdicts result in 8 wins, 8 ties, and 16 losses. These results suggest that the judge model exhibits relatively low self-preference bias and behaves fairly as an evaluator.

\begin{table}[t!]
\resizebox{\textwidth}{!}{%
\begin{tabular}{@{}l*{5}{c}c*{5}{c}@{}}
\toprule
\scriptsize
& \multicolumn{5}{c}{\textsc{\textbf{NorRewrite-Instruct}}} & & \multicolumn{5}{c}{\textsc{\textbf{NorSummarize-Instruct}}} \\
\cmidrule(lr){2-6} \cmidrule(lr){8-12}
\textbf{Model} & \textsc{\textbf{nob}} & \textsc{\textbf{nno}} & \textsc{\textbf{swe}} & \textsc{\textbf{dan}} & \textsc{\textbf{eng}} & & \textsc{\textbf{nob}} & \textsc{\textbf{nno}} & \textsc{\textbf{swe}} & \textsc{\textbf{dan}} & \textsc{\textbf{eng}} \\
\midrule
Instructions & \textbf{99.3\%} & 0.7\% & 0.0\% & 0.0\% & 0.0\% & & \textbf{96.4\%} & 3.6\% & 0.0\% & 0.0\% & 0.0\%\\[0.5em]
NorMistral-7B-warm-IT & \textbf{98.6\%} & 0.7\% & 0.0\% & 0.0\% & 0.7\% & & \textbf{99.0\%} & 0.5\% & 0.0\% & 0.0\% & 0.5\% \\
Mistral-Nemo-12B-IT & \textbf{87.5\%} & 0.7\% & 0.0\% & 1.4\% & 9.7\% & & \textbf{77.2\%} & 0.0\% & 0.0\% & 0.5\% & 21.8\% \\
Mistral-7B-IT & 29.9\% & 0.0\% & 0.0\% & 6.9\% & \textbf{63.2\%} & & 35.5\% & 0.0\% & 0.0\% & 4.6\% & \textbf{59.4\%} \\
Meta/Llama-3-8B-IT & 34.0\% & 0.0\% & 0.0\% & 0.0\% & \textbf{66.0\%} & & \textbf{49.7\%} & 0.0\% & 0.0\% & 0.5\% & 49.2\% \\
AI-Sweden/Llama3-8B-IT & 0.0\% & 0.0\% & \textbf{100.0\%} & 0.0\% & 0.0\% & & 0.0\% & 0.0\% & \textbf{100.0\%} & 0.0\% & 0.0\% \\
\bottomrule
\end{tabular}%
}
\caption{\textbf{Language distribution in model responses on NorRewrite-Instruct and NorSummarize-Instruct.} We show the percentages of instructions and responses in Norwegian Bokmål, Nynorsk, Swedish, Danish and English.}
\label{tab:language-distribution}
\end{table}

\renewcommand{\arraystretch}{1.75}

\begin{table}[t!]
\resizebox{\textwidth}{!}{%
\begin{tabular}{@{}l@{\hspace{3em}}ccccc@{\hspace{1.25em}}ccccccc@{\hspace{1.25em}}c}
\toprule
& \multicolumn{6}{@{}c@{}}{\textsc{\textbf{NorRewrite-Instruct}}} &  & \multicolumn{6}{@{}c@{}}{\textsc{\textbf{NorSummarize-Instruct}}} \\[1em]
\textbf{Model} & \rotatebox{90}{NorMistral-7B-warm-IT} & \rotatebox{90}{Mistral-Nemo-12B-IT} & \rotatebox{90}{Mistral-7B-IT} & \rotatebox{90}{Meta/Llama-3-8B-IT} & \rotatebox{90}{AI-Sweden/Llama-3-8B-IT} & \textbf{Average} & &
\rotatebox{90}{NorMistral-7B-warm-IT} & \rotatebox{90}{Mistral-Nemo-12B-IT} & \rotatebox{90}{Mistral-7B-IT} & \rotatebox{90}{Meta/Llama-3-8B-IT} & \rotatebox{90}{AI-Sweden/Llama-3-8B-IT} & \textbf{Average} \\
\midrule
NorMistral-7B-warm-IT &  --- & \cellcolor[rgb]{0.94,0.80,0.73}38.9 & \cellcolor[rgb]{0.58,0.71,1.00}79.2 & \cellcolor[rgb]{0.97,0.67,0.56}24.7 & \cellcolor[rgb]{0.48,0.62,0.98}87.5 & \cellcolor[rgb]{0.80,0.85,0.93}57.6\% & &
--- & \cellcolor[rgb]{0.96,0.76,0.67}33.8 & \cellcolor[rgb]{0.78,0.84,0.94}59.4 & \cellcolor[rgb]{0.96,0.59,0.47}18.1 & \cellcolor[rgb]{0.53,0.67,0.99}83.2 & \cellcolor[rgb]{0.88,0.86,0.85}48.6\% \\
Mistral-Nemo-12B-IT &  \cellcolor[rgb]{0.77,0.84,0.95}61.1 & --- & \cellcolor[rgb]{0.51,0.65,0.99}84.8 & \cellcolor[rgb]{0.91,0.84,0.79}43.8 & \cellcolor[rgb]{0.44,0.58,0.96}91.4 & \cellcolor[rgb]{0.67,0.78,0.99}70.3\% & &
\cellcolor[rgb]{0.72,0.81,0.98}66.2 & --- & \cellcolor[rgb]{0.60,0.73,1.00}77.3 & \cellcolor[rgb]{0.94,0.80,0.73}38.3 & \cellcolor[rgb]{0.44,0.58,0.96}91.1 & \cellcolor[rgb]{0.70,0.80,0.99}68.2\% \\
Mistral-7B-IT & \cellcolor[rgb]{0.96,0.63,0.51}20.8 & \cellcolor[rgb]{0.95,0.56,0.44}15.2 & --- & \cellcolor[rgb]{0.88,0.41,0.32}4.9 & \cellcolor[rgb]{0.77,0.84,0.95}60.7 & \cellcolor[rgb]{0.97,0.68,0.56}25.4\% & & \cellcolor[rgb]{0.93,0.82,0.75}40.6 & \cellcolor[rgb]{0.97,0.65,0.53}22.7 & --- & \cellcolor[rgb]{0.94,0.55,0.44}15.1 & \cellcolor[rgb]{0.70,0.80,0.98}67.5 & \cellcolor[rgb]{0.95,0.79,0.70}36.5\% \\
Meta/Llama-3-8B-IT & \cellcolor[rgb]{0.62,0.74,1.00}75.3 & \cellcolor[rgb]{0.81,0.85,0.92}56.2 & \cellcolor[rgb]{0.40,0.53,0.93}95.1 & --- & \cellcolor[rgb]{0.43,0.56,0.95}92.7 & \cellcolor[rgb]{0.57,0.70,1.00}79.8\%  & &
\cellcolor[rgb]{0.55,0.68,0.99}81.9 & \cellcolor[rgb]{0.76,0.83,0.96}61.7 & \cellcolor[rgb]{0.51,0.65,0.99}84.9 & --- & \cellcolor[rgb]{0.37,0.50,0.91}97.9 & \cellcolor[rgb]{0.55,0.69,0.99}81.6\% \\
AI-Sweden/Llama-3-8B-IT & \cellcolor[rgb]{0.93,0.52,0.41}12.5 & \cellcolor[rgb]{0.91,0.46,0.36}8.6 & \cellcolor[rgb]{0.94,0.81,0.74}39.3 & \cellcolor[rgb]{0.90,0.44,0.34}7.3 & --- & \cellcolor[rgb]{0.95,0.58,0.46}16.9\% & &
\cellcolor[rgb]{0.95,0.58,0.46}16.8 & \cellcolor[rgb]{0.91,0.47,0.36}8.9 & \cellcolor[rgb]{0.96,0.75,0.65}32.5 & \cellcolor[rgb]{0.86,0.36,0.29}2.1 & --- & \cellcolor[rgb]{0.94,0.55,0.44}15.0\% \\
\bottomrule
\end{tabular} %
}
\caption{\textbf{Instruction-finetuned LMs' win-rates (\%)} when evaluating for the language bias. Cold-colored cells indicate high win-rate, while warm-colored cells indicate low win-rate.}
\label{tab:judge-invariant}
\end{table}

\paragraph{Language Bias} Since we evaluate the \textit{Norwegian} capabilities of LMs responding to \textit{Norwegian} instructions, only responses written in Norwegian should be the preferred ones. We use GlotLID \citep{kargaran-etal-2023-glotlid} to analyze the language distribution in the instructions as well as in the model responses (see \Cref{tab:language-distribution}). Surprisingly, only NorMistral-7B-warm-IT consistently answers in Norwegian. Other models often switch either to English or related Scandinavian languages. To better understand the effect of requiring the responses to be in Norwegian, we modify our LLM-as-a-judge prompt template from \Cref{app:judge-prompt} by explicitly instructing the judge to be invariant to the language of the responses. This allows us to measure the Norwegian instruction following capabilities rather than the quality of producing Norwegian. \Cref{tab:judge-invariant} shows that the LM ranking has changed, with Meta/Llama-3-8B-IT becoming the most capable instruction-following LM. Conversely, NorMistral-7B-warm-IT has the expected win-rate of 58\% and 48\%, which suggests that its high rate in the main experiment is more due to its capabilities to consistently producing Norwegian.

\paragraph{Position Bias} Position bias is a common bias within the LLM-as-a-judge paradigm, where a judge model prefers a response based on its position regardless of the content \citep{wang-etal-2024-large-language-models-fair}. While we  mitigate this bias by evaluating each response pair twice with switched positions as shown in \Cref{eq:winrate}, we observe a minor preference for the second position. Our judge prefers the first response $416\times$ and the second one $538\times$ on NorRewrite-Instruct; on NorSummarize-Instruct, the bias is less apparent -- with 1\,100 and 1\,156 position preferences. In case of Primeape, we observe self-agreement rates of 45.83\% with ties and 64.52\% without ties. Overall, position bias appears to have an insignificant impact on NorRewrite-Instruct and NorSummarize-Instruct, whereas its effect is more pronounced on Primeape.

\subsection{Prompt Template for LLM-as-a-judge}
\label{app:judge-prompt}

We adapt the HREF prompt template provided in \newcite{lyu2024href} by localizing it to Norwegian and specifying that a Norwegian response should always be preferred over a non-Norwegian one.

\vspace{1em}
\noindent\footnotesize\texttt{\textbf{System prompt:}}\vspace{-0.3em}
\begin{minted}[linenos=false, breaklines=true, baselinestretch=1.0, bgcolor=bg, breakanywhere=true, fontfamily=tt, fontsize=\scriptsize, xleftmargin=1em, xrightmargin=1em]{markdown}

You are a helpful assistant that helps us rate a Norwegian AI model's responses to instructions.

\end{minted}

\noindent\footnotesize\texttt{\textbf{User prompt:}}\vspace{-0.3em}
\begin{minted}[linenos=false, breaklines=true, baselinestretch=1.0, bgcolor=bg, breakanywhere=true, fontfamily=tt, fontsize=\scriptsize, xleftmargin=1em, xrightmargin=1em]{markdown}

Decide which response from the Norwegian AI system following the instruction is better, considering the following questions:
1. Most importantly, the AI systems should always respond in Norwegian. If a response is not in Norwegian, then you should consider it incorrect --- such a response should always be rated lower than any (even incorrect) response in Norwegian.
2. Does the response precisely follow the instruction? For example, a response that includes unrelated information or does not fulfill the task is not precisely following the instruction. Compare each response with the provided human response to decide if a response faithfully follows the instruction, especially when the instruction asks for expected word count or format.
3. Is the response helpful? For example, if the instruction asks for a recipe for healthy food, and the response is a useful recipe, then you can consider it helpful.
4. Is the language of the response natural? For example, AI responses are often verbose or repetitive, which is not natural. Compare with the provided human response to decide whether a response is natural.
5. Is the response factual/accurate? AI responses often make up new information. For example, if the response claims that Jens Stoltenberg is the current prime minister of Norway, then you should consider it inaccurate. Compare with the provided human response to verify whether a response is factual and accurate, especially with numbers.
6. Based on your aesthetics, which one do you prefer? For example, you might prefer one poem over another poem.

Select the response A or B that you prefer, or select tie if the two responses are similarly good or bad. Note that the responses can be truncated (don't consider that as a mistake).

Here are three examples:

# Example 1:

## Instruction:
Omformulér følgende spørsmål: "Hva er hovedstaden i Frankrike?"

## Response A:
Hovedstaden i Frankrike er Paris.

## Response B:
Kan du fortelle meg navnet på byen som fungerer som hovedstaden i Frankrike?

## Human Response:
Hva heter Frankrikes hovedstad?

In this example, B paraphrases the question as asked by the instruction. In contrast, A does not follow instruction as it answers the question instead. Human Response also paraphrases the question, just in a slightly different way. To sum up, B is the best response because it follows the instruction.

## Which is best, A, B, or tie?
B

# Example 2:

## Instruction:
Bytt ut det første verbet med et synonym:
Jeg elsker å surfe

## Response A:
Jeg hater å surfe

## Response B:
I like to surf

## Human Response:
Jeg liker å surfe

Response A tries to follow the instruction as it changes the first verb of the sentence, but it uses an antonym instead of a synonym. The response B might be correct, but it is written in English, not Norwegian, and non-Norwegian responses should always be rated as worse. Human Response changes the first verb, "elsker" (love), into its synonym, "liker" (like), as asked by the instruction. In conclusion, A is better than B because it is written in Norwegian.

## Which is best, A, B, or tie?
A

# Example 3:

## Instruction:
Bytt ut det første verbet med et synonym:
Jeg elsker å surfe

## Response A:
Jeg hater å surfe

## Response B:
Jeg liker ikke å surfe

## Human Response:
Jeg liker å surfe

In this example, neither output is correct and the responses are very similar. Human Response changes the first verb into its synonym, as asked by the instruction. To conclude, both A and B are equally incorrect, so the answer is tie.

## Which is best, A, B, or tie?
tie

Now here is the real task, first describe the qualities of each response and then end your message by writing "## Which is best, A, B, or tie?" and selecting among: A, B, or tie.

# Task:

## Instruction:
{{instruction}}

## Response A:
{{output_1}}

## Response B:
{{output_2}}

## Human Response:
{{output_human}}
\end{minted}

\renewcommand{\arraystretch}{1.0}